\documentclass[3p,12pt]{elsarticle}

\usepackage{graphicx}
\usepackage{amsfonts,amsmath}
\usepackage{lineno,hyperref}
\usepackage{subfig}
\usepackage{float}
\usepackage{multirow}
\usepackage[section]{placeins} 

\newcommand{\bp}{{\boldsymbol{p}}}
\newcommand{\bx}{{\boldsymbol{x}}}

\makeatletter
\def\els@aparagraph[#1]#2{\elsparagraph[#1]{#2\@addpunct{.}}}
\def\els@bparagraph#1{\elsparagraph*{#1\@addpunct{.}}}
\makeatother

\newtheorem{Ex}{Example}

\journal{Neurocomputing}

\begin{document}

\begin{frontmatter}

\title{An evidential classifier based on Dempster-Shafer theory and deep learning}


\author[utc]{Zheng Tong\corref{cor1}}
 \ead{zheng.tong@hds.utc.fr}
 \cortext[cor1]{Corresponding author}
\author[utc]{Philippe Xu}
 \ead{philippe.xu@hds.utc.fr}
\author[utc,shu,iuf]{Thierry Den{\oe}ux}
 \ead{thierry.denoeux@hds.utc.fr}


\address[utc]{Universit\'e de technologie de Compi\`egne, CNRS, Heudiasyc, Compi\`egne, France}
\address[shu]{Shanghai University, UTSEUS, Shanghai, China}
\address[iuf]{Institut universitaire de France, Paris, France}

\begin{abstract}
We propose a new classifier based on Dempster-Shafer (DS) theory and a  convolutional neural network (CNN) architecture for set-valued classification. In this classifier,  called the evidential deep-learning classifier,  convolutional and pooling layers first extract high-dimensional features from input data. The features are then converted into mass functions and aggregated by Dempster's rule in a DS layer. Finally, an expected utility layer performs set-valued classification based on mass functions. We propose an end-to-end learning strategy for jointly updating the network parameters. Additionally, an approach for selecting partial multi-class acts is proposed. Experiments on image recognition, signal processing, and semantic-relationship classification tasks demonstrate that the proposed combination of deep CNN, DS layer, and expected utility layer makes it possible to improve classification accuracy and to make cautious decisions by assigning  confusing patterns to multi-class sets.
\end{abstract}


\end{frontmatter}


\section{Introduction}
\label{sec:introduction}

In machine learning,  classification refers to the task of predicting the class of a new sample based on a learning set  of labeled instances. The most common classification problem is \emph{precise classification}, in which a sample is classified into one and only one of the possible classes. Unfortunately, such a hard assignment often leads to misclassification in case of high uncertainty. For example, ambiguity occurs when the feature vector does not contain sufficient information to identify a precise class, and multiple classes have similar probabilities. Also, a classifier with only precise classification may fail to identify outliers belonging to class that is not represented in the learning set.

\emph{Set-valued classification} \cite{ha1997optimum,mortier2019efficient,mapartial} is a potential way to solve this problem; it is defined as the assignment of a new observation into a non-empty subset of  classes when the uncertainty is too high to make a precise classification. For instance, given a class set $\Omega = \{\omega_1, \omega_2, \omega_3\}$, we may not be able to reliably classify a sample $\boldsymbol{x}$  into a single class, but it may be almost sure that it does not belong to $\omega_3$.  In this case, it is more cautious to assign $\boldsymbol{x}$ to the set $\{\omega_1, \omega_2\}$. Classification with a reject option in \cite{chow70,tong2019ConvNet} can be regarded as a special case of  set-valued classification,  rejection being equivalent to assigning a sample to the entire set of possible classes.  A related problem concerns the treatment of outliers, which cannot be classified into any of the known classes, a problem referred to as ``novelty detection'' or ``distance rejection'' \cite{dubuisson93}. Depending on the method, such samples may be assigned to the empty set, or to the whole set $\Omega$, reflecting maximum uncertainty \cite{denoeux2000neural}. Set-valued classification makes it possible to better reflect classification uncertainty, increase the cautiousness of classifiers and ultimately reduce the error rate. Precise classification can be considered as a special case of set-valued classification, in which only the sets with one class  are considered.

In this study, we propose a new classifier based on Dempster-Shafer (DS) theory and deep convolutional neural networks (CNN) for set-valued classification, called the \emph{evidential deep-learning classifier}\footnote{A short preliminary version of this paper was presented at the SUM 2019 conference \cite{tong2019ConvNet}.}. In this classifier, a deep CNN is used to extract high-order features from raw data. Then, the features are imported into a distance-based DS layer \cite{denoeux2000neural} for constructing mass functions. Finally, mass functions are used to compute the utilities of acts assigning to a set of classes for set-valued classification.  The whole network is trained using  an end-to-end learning procedure. Additionally, we provide a strategy for considering only some subsets of classes instead of considering all of them. The effectiveness of the classifier and its decision strategy are demonstrated and discussed using  three types of datasets (image, signal, and semantic relationship).  The main contribution of this study is the demonstration that  CNNs  can be enhanced with set-valued classification and novelty detection capabilities thanks to the addition of an additional DS layer, while maintaining their very good performance in precise classification tasks.

\subsection*{Related work}

In recent years, with the explosive development of deep learning \cite{lecun2015deep}, several models have been developed for precise classification, such as convolutional neural networks (CNNs) \cite{kim2014convolutional,leng20163d,WANG202095}, recurrent neural networks \cite{LICIOTTI2020501,mikolov2010recurrent,5947611}, graph neural networks \cite{scarselli2008graph,1517930}, and deep autoencoders \cite{vincent2008extracting,vincent2010stacked}. Deep learning is a class of machine learning methods that uses multiple layers to progressively extract higher-level features from raw data as object representation. For example, when processing images using a CNN, lower layers may identify edges, while higher layers may identify more abstract concepts relevant to humans such as digits, letters or faces. Object representation based on deep learning is generally robust and reliable. In particular, the representation has a strong tolerance to translation and distortion of raw data. However, despite the power of the deep learning-based models in precise classification, we still face the problem of making them more cautious by allowing them to assign highly uncertain samples to sets of classes.

The Dempster-Shafer (DS) theory of belief functions \cite{dempster2008upper,shafer1976mathematical}, also referred to as \emph{evidence theory}, can be harnessed to provide a solution to the problem.  DS theory is a well-established formalism for reasoning and making decisions with uncertainty \cite{yager2008classic}. It  is based on representing independent pieces of evidence by completely monotone capacities and combining them using a generic operator called Dempster's rule \cite{shafer1976mathematical}. In the last two decades, DS theory has been increasingly applied to pattern recognition and supervised classification, following  three main directions. The first one is \emph{classifier fusion}, in which the outputs of several classifiers are transformed into belief functions and aggregated by a suitable combination rule (e.g., \cite{bi2012impact,liu2018classifier,quost11,ZHOU2016145}). Another direction is \emph{evidential calibration}: the decisions of classifiers are converted into mass functions with some frequency calibration property (e.g., \cite{MA20181282,minary2017face,minary2019,xu2016evidential,YUE201513}). The last approach is to design \emph{evidential classifiers }(e.g., \cite{denoeux2000neural,denoeux19f}), which break down the evidence of  input features into elementary mass functions and combine them by Dempster's rule. The outputs of an evidential classifier can be used for decision-making \cite{chen2018evidential,7532858}. Thanks to the generality and expressiveness of the DS formalism, the outputs of an evidential classifier provide more information than those of conventional classifiers (e.g., a neural network with a softmax layer) that convert an input feature vector into a probability distribution or any other distribution. For example, the expressiveness of an evidential classifier can be used for uncertainty quantification and ambiguity rejection \cite{denoeux96b,mapartial}. Over the years, two main principles for designing an evidential classifier have been proposed: the model-based and distance-based approaches. The former uses estimated class-conditional distributions \cite{smets1993belief}, while the latter constructs mass functions based on distances to prototypes \cite{denoeux2000neural,denoeux19f}. In practice, the performance of an evidential classifier mainly depends on two factors: the training data set and the reliability of object representation. 

In the last twenty years we have seen an increase in the size of benchmark datasets for supervised learning at an unprecedented rate from $10^2$ to $10^5$ \cite{krizhevsky2012imagenet} and even $10^9$ instances \cite{sakaguchi2014efficient}. However, little has been done to hybridize recent techniques for object representation, such as deep learning, with evidential classifiers for decision-making. 
Some studies combining DS theory and deep learning has been reported, but most of these studies address the problem of deep-learning classifier fusion, where the outputs of several deep-learning models are regarded as pieces of evidence and aggregated by Dempster's rule of combination. For example, Soua et al. \cite{soua2016big} use deep belief networks to independently predict traffic flow using streams of data and event-based data, and then update the beliefs from the networks by Dempster's conditional rule to achieve enhanced prediction. Tian et al. \cite{tian2020deep} also use Dempster's rule to fuse the beliefs from several deep-learning models with different types of data to detect anomalous network behavior patterns. Das et al. \cite{das2019combining} use CNNs to perform superpixel semantic segmentation with three levels;  DS theory is then utilized to combine the segmentation results of the three levels into reliable ones. Besides, Guo et al. \cite{guo2019ifusion} propose an ``iFusion'' framework, which uses Dempster's rule to combine different deep-learning discrimination models taking real-time  or heterogeneous data as input. Similar works using DS theory for deep-learning classifier fusion can also be found in the field of posture recognition \cite{li2020standing}, remote-sensing images processing \cite{du2020incorporating}, and emotion classification \cite{xu2020emotion}. In \cite{denoeux2019logistic}, the author shows that the operations performed in a multilayer perceptron classifier can be analyzed from the point of view of DS theory as the application of Dempster's rule; however, he does not propose a new model. Though Yuan et al. \cite{yuan2020evidential} propose a method using DS theory to measure the uncertainty of outputs from deep neural networks for decision-making, it still appears that little has been done to use features from a deep-learning model as inputs of an evidential classifier to generate informative mass-function outputs for decision-making allowing set-valued classification, a gap that we aim to fill in this work.

The rest of the paper is organized as follows. Section \ref{sec:background} starts with a brief reminder of DS theory, the DS layer for constructing mass functions, and feature representation via deep CNN. The new classifier is then introduced in Section \ref{sec:classifier}. Section \ref{sec:experimental_results} reports numerical experiments, which demonstrate the advantages of the proposed classifier. Finally, we conclude the paper in Section \ref{sec:conclusions}.

\section{Background}
\label{sec:background}

This section first recalls some necessary definitions regarding DS theory (Section \ref{sec:DS_theory}) and the evidential neural network  introduced in \cite{denoeux2000neural} (Section \ref{sec:DS_layer}). Then, a brief description of feature representation via deep CNN is provided in Section \ref{sec:feature_representation}.

\subsection{Dempster-Shafer theory}
\label{sec:DS_theory}

    The main concepts regarding DS theory are briefly presented in this section, and some basic notations are introduced. Detailed information can be found in Shafer's original work \cite{shafer1976mathematical} and in the recent review \cite{denoeux20b}.

    Let $\Omega=\{\omega_1,\ldots,\omega_M\}$ be a finite set of states, called the \emph{frame of discernment}. A \emph{mass function} on $\Omega$ is a mapping $m$ from $2^\Omega$ to [0,1] such that $m(\emptyset)=0$ and
    \begin{equation}
    \sum_{A\subseteq\Omega}m(A)=1.
    \end{equation}
    For any $A\subseteq\Omega$, each mass $m(A)$ is interpreted as a share of a unit mass of belief allocated to the hypothesis that the truth is in $A$, and which cannot be allocated to any strict subset of $A$ based on the available evidence. Set $A$ is called a \emph{focal element} of $m$ if $m(A)>0$.

    Two  mass functions $m_1$ and $m_2$ representing independent items of evidence can be combined conjunctively by Dempster's rule $\oplus$ \cite{shafer1976mathematical} as
    \begin{subequations}
    \label{con:dempster}
    \begin{equation}
     \label{con:dempster1}
    (m_1\oplus m_2)\left(A\right)=\frac{(m_1 \cap m_2)(A)}{1-(m_1 \cap m_2)(\emptyset)}
    \end{equation}
     for all $A\neq\emptyset$, with
       \begin{equation}
    (m_1\cap m_2)(A)=\sum_{B\cap C=A}m_1\left(B\right)m_2\left(C\right)
        \end{equation}
        and
          \begin{equation}
        (m_1\cap m_2)(\emptyset)=\sum_{B\cap C=\emptyset}m_1\left(B\right)m_2\left(C\right).
        \end{equation}
    \end{subequations}
    Mass functions $m_1$ and $m_2$ can be combined if and only if the denominator on the right-hand side of Eq. \eqref{con:dempster1} is strictly positive. The operator $\oplus$ is commutative and associative.

    For decision-making with belief functions, we define the \emph{lower and upper expected utilities}  \cite{denoeux2019decision} of selecting $\omega_i$ as, respectively,
    \begin{subequations}\label{con:expected_precise}
    \begin{equation}\label{con:lower_precise}
    \mathbb{\underline{E}}_m(f_{\omega_i})=\sum_{B\subseteq\Omega}^{}m(B) \min_{\omega_j\in B} u_{ij},
    \end{equation}
    and
    \begin{equation}\label{con:upper_precise}
    \mathbb{\overline{E}}_m(f_{\omega_i})=\sum_{B\subseteq\Omega}^{}m(B) \max_{\omega_j\in B} u_{ij},
    \end{equation}
    \end{subequations}
    where $u_{ij} \in [0,1]$ is the utility of selecting $\omega_i$ when the true state is $\omega_j$, and $f_{\omega_i}$ denotes the act of selecting $\omega_i$. A pessimistic decision-maker (DM) typically selects the act with the largest lower expected utility, while an optimistic DM maximizes the upper expected utility. The generalized Hurwicz decision criterion \cite{hurwicz1951generalized,jaffray89,strat90,denoeux2019decision} models the DM's attitude to ambiguity by a \emph{pessimism index} $\nu$ and defines the expected utility of act $f_{\omega_i}$ as the weighted sum
      \begin{equation}\label{con:hurwicz}
    \mathbb{E}_{m,\nu}(f_{\omega_i})= \nu \mathbb{\underline{E}}(f_{\omega_i}) + (1-\nu)\mathbb{\overline{E}}(f_{\omega_i}).
    \end{equation}
    It is clear that the pessimistic and optimistic attitudes correspond, respectively, to $\nu=1$ and $\nu=0$.

    \subsection{Evidential neural network}
    \label{sec:DS_layer}

    Based on DS theory, Den{\oe}ux \cite{denoeux2000neural} proposed a distance-based neural-network layer for constructing mass functions, also known as the \emph{evidential neural network (ENN) classifier}. In the ENN classifier, the proximity of an input vector to prototypes is considered as  evidence about its class. This evidence is converted into mass functions and combined using Dempster's rule. This section provides a short description of the ENN classifier.

    We consider a training set ${{\cal X}=\left\{\boldsymbol{x}_1,\ \boldsymbol{x}_2,\ldots,\boldsymbol{x}_N\right\}\subset\mathbb{R}^P}$  of $N$ examples represented with $P$-dimensional feature vectors, and an ENN classifier composed of $n$ prototypes $\{\boldsymbol p^1,\ldots,\boldsymbol p^n\}$  in $\mathbb{R}^P$. For a test sample $\boldsymbol x$, the ENN classifier constructs mass functions that quantify the uncertainty about its class in  $\Omega=\{\omega_1,\ldots,\omega_M\}$, using a three-step procedure. This procedure can  be implemented in a neural-network layer, which will be plugged into a deep CNN in Section \ref{sec:network_architecture}. The three-step procedure is defined as follows.

\begin{description}
    \item{Step 1:} The distance-based support between $\boldsymbol x$ and each reference pattern $\boldsymbol p^i$ is computed as
        \begin{equation}\label{con:si}
        s^i=\alpha^i\exp(-\left(\eta^id^i\right)^2) \quad i=1,\ldots,n,
        \end{equation}
        where $d^i={\left\|\boldsymbol x-\boldsymbol p^i\right\|}$ is the Euclidean distance between $\bx$ and prototype $\bp^i$, and $\alpha^i \in (0,1)$ and $\eta^i \in \mathbb{R}$ are parameters associated with  prototype $\boldsymbol p^i$. Prototype vectors $\boldsymbol p^1,\ldots,\boldsymbol p^n$ can be considered as vectors of  connection weights between the input layer and a hidden layer of $n$  Radial basis Function (RBF) units.

    \item{Step 2:} The mass function $m^i$ associated to reference pattern $\boldsymbol p^i$ is computed as
        \begin{subequations}
        \label{con:m^i}
        \begin{align}
        m^{i}(\{\omega_j\})& = h_j^i s^{i}, \quad j=1,\ldots,M\\
        m^{i}(\Omega)& =1-s^{i},
        \end{align}
        \end{subequations}
        where $h_j^i$ is the degree of membership of prototype $\bp^i$ to class $\omega_j$ with $\sum_{j=1}^Mh_j^i=1$. We denote the vector of masses induced by prototype $\bp^i$ as
      \[
        \boldsymbol m^{i}=(m^{i}(\{\omega_1\}), \ldots,m^{ i}(\{{\omega}_{M}\}),m^{\mathit i}(\Omega))^T.
        \]
        Eq. \eqref{con:m^i} can be regarded as computing the activation of units in a second hidden layer of the ENN classifier, composed of $n$ modules of $M+1$ units each. The result of module $i$ corresponds to the belief masses assigned by $m^i$.

    \item{Step 3:} The $n$ mass functions $\boldsymbol m^i$, $i=1,\ldots,n$, are aggregated by Dempster's rule \eqref{con:dempster}. The combined mass function is computed iteratively as $\mu^1=m^1$ and $\mu^i=\mu^{i-1}\cap m^i$ for $i=2,\ldots,n$. We have
        \begin{subequations}
        \label {con:BFoutput}
        \begin{equation}
        \mu^i(\{\omega_j\})=\mu^{i-1}(\{\omega_j\})m^{ i}(\{\omega_j\})+\mu^{i-1}(\{\omega_j\})m^{ i}(\{\Omega\})+\mu^{i-1}(\Omega)m^{ i}(\{\omega_j\})        \label{con:mu^ij}
        \end{equation}
        for  $i=2,\ldots,n$ and $j=1,\ldots,M$, and
        \begin{equation}
        \mu^i(\Omega)=\mu^{i-1}(\Omega)m^{ i}(\Omega) \quad i=2,\ldots,n.
        \label{con:mu^iM}
        \end{equation}
        \end{subequations}
    The vector of outputs from the ENN classifier $\boldsymbol m=(m(\{\omega_1\}), \ldots,m(\{{\omega}_{ M}\}),m(\Omega))^T$ is finally obtained as
    \[
    m(\{\omega_j\})=\frac{\mu^n(\{\omega_j\})}{\sum_{j'=1}^M \mu^n(\{\omega_{j'}\})+\mu^n(\Omega)}
    \]
    and
        \[
    m(\Omega)=\frac{\mu^n(\Omega)}{\sum_{j'=1}^M \mu^n(\{\omega_{j'}\})+\mu^n(\Omega)}.
    \]
\end{description}

    \subsection{Feature representation via deep CNN}
    \label{sec:feature_representation}

    In practice, the effectiveness of an ENN classifier heavily depends on the information contained in its input features. Feature representation, an essential part of the machine learning workflow, consists in discovering the predictors needed for classification from raw data. In recent years, deep learning models \cite{lecun2015deep} have become very popular because of their ability to construct rich deep feature representations, allowing them to achieve exceptional performance in such tasks as   pattern recognition and segmentation \cite{eitel2015multimodal,long2015fully,Zeng2014relation}, signal processing \cite{piczak2015environmental,7829341}, and even material discovery \cite{moosavi2019capturing,stokes2020deep}.

    Deep CNNs, one of the most widely used deep learning architectures,  are a special type of multi-layered neural network and the main focus of this paper. The most common  CNNs consist of convolutional layers, pooling layers, and fully connected layers. Convolutional and pooling layers are defined as stages. A stage converts its input data into an intermediate representation, working as a feature extractor. In general, a deep CNN is composed of several stacked stages that process raw data and repeatedly converts them into higher-level feature maps. Then,  fully connected layers, serving as a decision maker, assign the  input to one of the classes based on the feature maps. Therefore, the final output of the stacked stages in a deep CNN can be considered as a feature representation of the input data. In the study, these high-level features  are used as  input to a DS layer capable of set-valued classification, as will be shown in Section \ref{sec:network_architecture}.

    To understand the feature representation of deep CNNs, we briefly recall the processes of convolutional and pooling layers. Consider a stage with input $\boldsymbol z=(z^1,\dots,z^D)$ consisting of $D$ input maps or \emph{input channels} $z^i$ ($i=1,\dots,D$) with  size  $H \times W$. A convolutional layer consists of several convolution kernels that extract feature maps from  $\boldsymbol z$. A convolution kernel is a small matrix used to apply a convolution operation to each input map by sliding over the map, performing an element-wise multiplication with the part of the input map where the kernel is currently on, summing up the multiplied results into a single value, and then adding the bias of the kernel to the summed value. Thus, the processes in a convolutional layer, consisting of $e$ convolution kernels with  size  $a \times b$, are expressed as
    \begin{equation}\label{con:convolution}
      c^j=f(b^j+\sum_iw^{i,j}\ast z^i),
    \end{equation}
    where $w^{i,j}$ is the convolution kernel between the $i$-th input map and the $j$-th output map; $b^j$ is the bias of kernel $w^{i,j}$; $\ast$ denotes the convolution operation; $z^i$ is the $i$-th input map with  size  $H \times W$, $i=1,\dots,D$; $c^j$ is the $j$-th output feature map, with size  $\frac{H-a+1}{r} \times \frac{W-b+1}{r}$, $j=1,\dots,e$; $r$ is the stride with which the kernel slides over input map $z^i$; $f$ is the activation function, such as the rectified linear unit $\textsf{ReLU}(x)=\max(0,x)$ \cite{5459250}. A pooling operation with an $s \times s$ non-overlapping local region is formulated as
    \begin{equation}\label{con:pooling}
      po_{a,b}^k=\left(\beta^{1},\dots,\beta^{s \times s}\right)^T \cdot \textsf{Or}(c^k_{as,bs},\dots,c^k_{as+s,bs},\dots,c^k_{as+s,bs+s}),
    \end{equation}
    where $po_{a,b}^k$ is the element $(a,b)$ from the $k$-th output map, which is in the $a$-th row and the $b$-th column; $\textsf{Or}$ is a sort function from maximum to minimum; $\cdot$ denotes dot product; $\left(\beta^{1},\dots,\beta^{s \times s}\right)$ is the pooling weight vector, such as max pooling $\left(\beta^{1},\beta^{2},\dots,\beta^{s \times s}\right)=(1,0,\dots,0)$ and mean pooling
    \[
    \left(\beta^{1},\dots,\beta^{s \times s}\right)=\left(\frac{1}{s \times s},\dots,\frac{1}{s \times s}\right).
    \]

\section{Proposed classifier}
\label{sec:classifier}

In this section, we describe the proposed classifier. Section \ref{sec:network_architecture} presents the overall architecture composed of several stages from a deep CNN for feature representation, a DS layer to construct mass functions, and an expected utility layer for decision-making. The details of the expected utility layer are described in  Section \ref{sec:EU_layer}, and the learning strategy for the proposed classifier is exposed in  Section \ref{sec:Learning}. Finally, an approach for selecting partial multi-class acts is introduced in Section \ref{sec:act_selection}.

    \subsection{Network architecture}
    \label{sec:network_architecture}

   The main idea of this work is to hybridize the ENN classifier presented in Section \ref{sec:DS_layer} and the  CNN architecture recalled in Section \ref{sec:feature_representation} by ``plugging'' a  DS layer followed by a utility layer at the output of a CNN. The architecture of the proposed method, called the \emph{evidential deep-learning classifier}, is illustrated in Figure \ref{fig:evidential_DL_classifier}. An evidential deep-learning classifier has the ability to perform set-valued classification and quantify the uncertainty about the class of the sample on  $\Omega=\{\omega_1,\ldots,\omega_M\}$ by a belief function. Propagation of information through this network can be described by  the following three-step procedure:

\begin{figure}
\centering
\includegraphics[width=\linewidth]{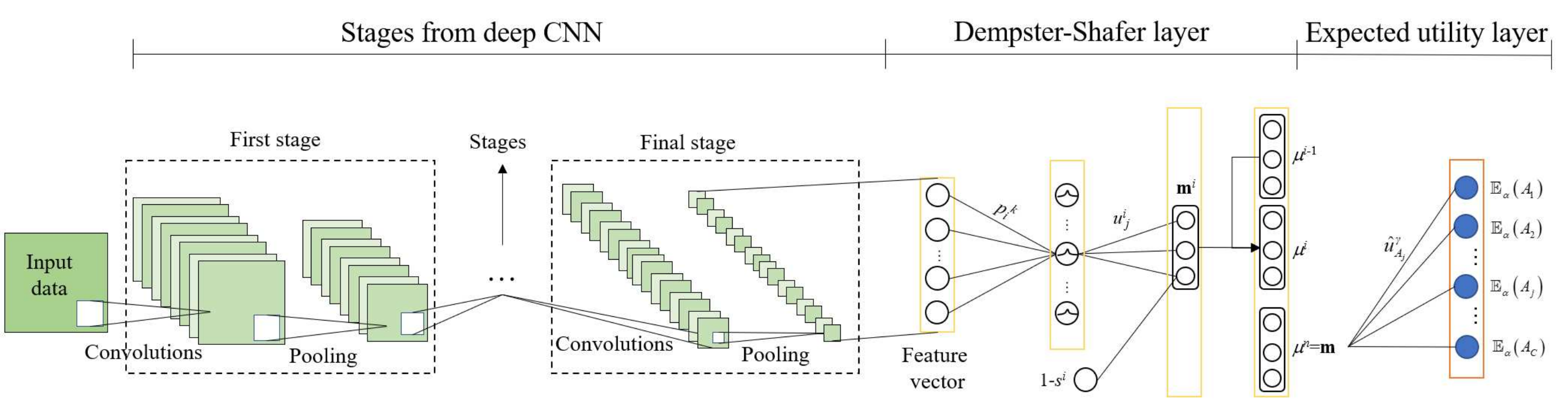}\\
\caption{Architecture of an evidential deep-learning classifier.}\label{fig:evidential_DL_classifier}
\end{figure}

\begin{description}
    \item{Step 1:} An input sample is propagated into several stages of a CNN architecture to extract latent features relevant for classification, as done in a probabilistic CNN. In the final stage, the $P$-dimensional output vector is a feature representation of the sample, ready to be fed as input to the DS layer. This architecture provides a robust and reliable representation of the input sample. Thanks to this representation, the evidential deep-learning classifier yields  similar or even better performance for precise classification than does a probabilistic classifier with the same stages. This superiority will be demonstrated by  performance comparisons between the evidential and probabilistic deep-learning classifiers in precise classification tasks (Section \ref{sec:experimental_results}).
    \item{Step 2:}  The feature vector computed in Step 1 is fed into the DS layer, in which it is  converted into mass functions  aggregated by Dempster's rule, as explained in Section \ref{sec:DS_layer}. The output of the DS layer is an  $(M+1)$ mass vector
    \[
    \boldsymbol m=(m(\{\omega_1\}), \ldots,m(\{{\omega}_{ M}\}),m(\Omega))^T,
    \] 
    which characterizes the classifier's belief about the probability of the sample class and quantifies the uncertainty in the object representation. The mass $m(\{\omega_i\})$ is a degree of belief that the sample belongs to class $\omega_i$. 
    The DS layer tends to allocate masses uniformly across classes when the feature representation contains confusing and conflicting information. The additional degree of freedom $m(\Omega)$ makes it possible to quantify the lack of evidence and verify whether the model is well trained \cite{tong2019ConvNet}. The advantages of the DS layer will be verified in the performance evaluation of set-valued classification using evidential deep-learning classifiers reported in Section \ref{sec:experimental_results}.
    \item{Step 3:}  The output mass vector $\boldsymbol m$ is fed into an expected utility layer for decision-making, where it is used to compute the expected utilities of acts. Each act is defined as the assignment of the sample to a non-empty subset $A$ of $\Omega$. Thus, the output of the expected utility layer is an expected-utility vector of size at most equal to $2^{M}-1$ if all of the possible acts are considered. The expected utility layer allows the proposed classifier to perform set-valued classification. This capability will be demonstrated by the performance comparison between the evidential and probabilistic deep-learning classifiers in set-valued classification and novelty detection tasks reported in Section \ref{sec:experimental_results}. The details of the expected utility layer for set-valued classification are introduced in the next section.
\end{description}

    \subsection{Expected utility layer}
    \label{sec:EU_layer}

    Let $\Omega=\{\omega_1,\dots,\omega_M\}$ be the set of classes. For classification problems with only precise prediction, an act is defined as the assignment of an example to one and only one of the $M$ classes. The set of acts is $\mathcal{F}=\{f_{\omega_1},\dots,f_{\omega_M}\}$, where $f_{\omega_i}$ denotes assignment to class $\omega_i$. To make decisions, we define a utility matrix $\mathbb{U}_{M \times M}$, whose general term $u_{ij} \in [0,1]$ is the utility of assigning an example to class $\omega_i$ when the true class is $\omega_j$. Here, $\mathbb{U}_{M \times M}$ is called  the \emph{original utility matrix}. For  decision-making with belief functions, each act $f_{\omega_i}$ induces expected utilities, such as the lower and upper expected utilities defined  by \eqref{con:expected_precise}.

    For classification problems with imprecise prediction, Ma and Den{\oe}ux \cite{mapartial} proposed an approach to conduct set-valued classification under uncertainty by generalizing the set of acts as partially assigning a sample to a non-empty subset $A$ of $\Omega$. Thus, the set of acts becomes $\mathcal{F}=\{f_A,A \in 2^\Omega \backslash {\emptyset}\}$, in which $2^\Omega$ is the power set of $\Omega$ and $f_A$ denotes the assignment to a subset $A$. In this study, subset $A$ is defined as a \emph{multi-class set} if $|A| \geq 2$. For decision-making with $\mathcal{F}$, the original utility matrix $\mathbb{U}_{M \times M}$ is extended to $\mathbb{U}_{(2^\Omega-1) \times M}$, where each element $\widehat{u}_{A,j}$ denotes the utility of assigning a sample to set $A$ of classes when the true label is $\omega_j$.

    When the true class is $\omega_j$, the utility of assigning a sample to set $A$ is defined as an Ordered Weighted Average (OWA) aggregation \cite{yager1988ordered} of the utilities of each precise assignment in $A$ as
    \begin{equation}\label{con:owa}
    {\widehat u}_{A,j}=\sum_{k=1}^{\left|A\right|}g_k \cdot u_{(k)j}^A,
    \end{equation}
    where $u_{(k)j}^A$ is the $k$-th largest element in the set $\{u^A_{ij},\omega_i \in A\}$ made up of the elements in the original utility matrix $\mathbb{U}_{M \times M}$, and weights $\boldsymbol g=(g_1,\dots,g_{|A|})$ represent the preference  to choose $u_{(k)j}^A$ when a classifier has to make a precise decision among a set of possible choices. The elements in weight vector $\boldsymbol g$ represent the DM's  \emph{tolerance  to imprecision}. For example, full tolerance to imprecision is achieved when the assignment act $f_A$ has utility 1 once set $A$ contains the true label, no matter how imprecise $A$ is. In the case, only the maximum utility of elements in set $\{u^A_{ij},\omega_i \in A\}$ is considered: $(g_1,g_2,\dots,g_{|A|})=(1,0,\dots,0)$. At the other extreme,  a DM attaching no value to imprecision would consider the act $f_A$ as equivalent to selecting one class uniformly at random from $A$: this is achieved when
    \[
    (g_1,g_2,\dots,g_{|A|})=\left(\frac1{|A|},\frac1{|A|},\dots,\frac1{|A|}\right).
    \]
    In this study, following \cite{mapartial}, we determine the weight vector $\boldsymbol g$ of the OWA operators by adapting O'Hagan's method \cite{ohagan88}. We define the \emph{imprecision tolerance degree} as
    \begin{equation}\label{TDI}
      TDI(\boldsymbol g)=\sum_{k=1}^{\left|A\right|}\frac{\left|A\right|-k}{\left|A\right|-1}g_k=\gamma,
    \end{equation}
    which equals to 1 for the maximum, 0 for the minimum, and 0.5 for the average. In practice, we only need to consider values of $\gamma$ between 0.5 and 1 as a precise assignment is preferable to an imprecise one when $\gamma \textless 0.5$ \cite{mapartial}. Given a value of $\gamma$, we can compute the weights of the OWA operator by maximizing the entropy
    \begin{equation}\label{ENT}
      ENT(\boldsymbol {g})=-\sum_{k=1}^{\left|A\right|} g_k  \log g_k
    \end{equation}
    subject to the constraints $TDI(\boldsymbol {g})=\gamma$, $\sum_{k=1}^{\left|A\right|}g_k=1$, and $g_k \geq 0$.

\begin{Ex} 
Table \ref{tab:example_utility} shows an example of the extended utility matrix generated by an OWA operator with $\gamma=0.8$ for a classification problem. The first three rows constitute the original utility matrix, indicating that the utility equals 1 when assigning a sample to its true class, otherwise it equals 0. The remaining rows are the matrix of the aggregated utilities. For example, we get a utility of 0.8  when assigning a sample to set $\{\omega_1,\omega_2\}$ if the true label is $\omega_1$.

\begin{table}
\center
\caption{Utility matrix extended by an OWA operator with $\gamma=0.8$.}\label{tab:example_utility}
\begin{tabular*}{\hsize}{@{}@{\extracolsep{\fill}}cccc@{}}
\hline
\multirow{2}{*}{}           & \multicolumn{3}{c}{Classes} \\ \cline{2-4}
                            & $\omega_1$ & $\omega_2$ & $\omega_3$ \\ \hline
$f_{\{\omega_1\}}$          & 1          & 0          & 0          \\
$f_{\{\omega_2\}}$          & 0          & 1          & 0          \\
$f_{\{\omega_3\}}$          & 0          & 0          & 1          \\
$f_{\{\omega_1,\omega_2\}}$ & 0.8        & 0.8        & 0          \\
$f_{\{\omega_1,\omega_3\}}$ & 0.8        & 0          & 0.8         \\
$f_{\{\omega_2,\omega_3\}}$ & 0          & 0.8        & 0.8        \\
$f_{\{\Omega\}}$            & 0.6819     & 0.6819     & 0.6819     \\ \hline
\end{tabular*}
\end{table}
\end{Ex}
    Based on an extended utility matrix $\mathbb{U}_{(2^\Omega-1) \times M}$ and the outputs of a DS layer $\boldsymbol m$, we can compute the expected utility of an act assigning a sample to set $A$ using the generalized Hurwicz criterion \eqref{con:hurwicz} as
    \begin{subequations}
    \begin{equation}\label{con:Hurwicz}
    \mathbb{E}_{m,\nu}(f_A)=\nu\mathbb{\underline{E}}_m(f_A)+(1-\nu)\mathbb{\overline{E}}_m(f_A),
    \end{equation}
    where $\mathbb{\underline{E}}_m(f_A)$ and $\mathbb{\overline{E}}_m(f_A)$ are, respectively, the lower and upper expected utilities
    \begin{equation}\label{con:lower}
    \mathbb{\underline{E}}_m(f_A)=\sum_{B\subseteq\Omega}^{}m(B) \min_{\omega_k\in B} {\widehat u}_{A,j},
    \end{equation}
    \begin{equation}\label{con:upper}
    \mathbb{\overline{E}}_m(f_A)=\sum_{B\subseteq\Omega}^{}m(B) \max_{\omega_k\in B}{\widehat u}_{A,j},
    \end{equation}
    \end{subequations}
    and $\nu$ is the pessimism index, which is considered as a hyperparameter of the proposed classifier. The sample is finally assigned to set $A$ such that 
    \begin{equation}
    \label{eq:decision}
    A=\arg \max_{\emptyset\neq B\subseteq \Omega} \mathbb{E}_{m,\nu}(f_{B}).
    \end{equation}

\begin{figure}
\centering
\includegraphics[width=0.5\linewidth]{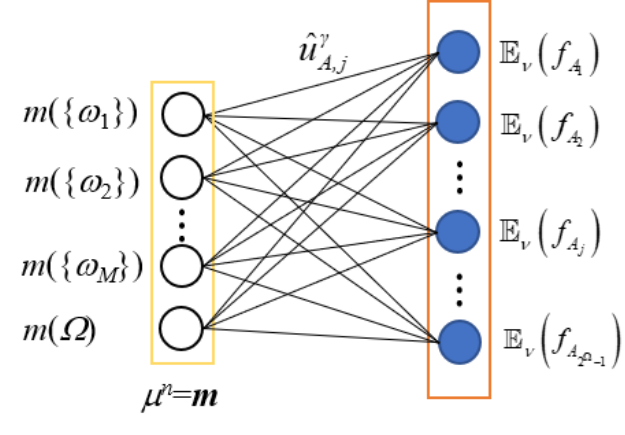}\\
\caption{Architecture of the expected utility layer.}\label{fig:eu_layer}
\end{figure}

    Similar to the DS layer, the procedure of assigning a sample to a set in $\mathcal{F}$ using utility theory can be summarized as a  layer of the neural network, called an \emph{expected utility layer}, as shown in Figure \ref{fig:eu_layer}. In this layer, the inputs and outputs are, respectively, the mass vector $\boldsymbol m$ from the preceding DS layer and the expected utilities of all acts in $\mathcal{F}$. The connection weight between unit $j$ of the DS layer and output unit $A\subseteq \Omega$ corresponding to the assignment to set $A$ is the utility value ${\widehat u}_{A,j}$. As coefficient $\gamma$ describing the imprecision tolerance degree is fixed, the connection weights of the expected utility layer do not need to be updated during training.

    \subsection{Learning}
    \label{sec:Learning}

    The evidential deep-learning classifier can be trained by a stochastic gradient descent algorithm. 
    Given a sample $\boldsymbol x$ with class label $\omega_\ast$, we define the prediction loss as

    \begin{subequations}\label{con:loss_define}
    \begin{equation}\label{con:loss}
    \begin{aligned}
    \mathcal{L}_\nu\left(\boldsymbol x\right)=&-\sum_{k=1}^M y_k \log\mathbb{E}_\nu(f_{\omega_k})+(1-y_k)\log (1-\mathbb{E}_\nu(f_{\omega_k}))
    \end{aligned}
    \end{equation}
    with
    \begin{equation}
    y_k=\left\{\begin{array}{l}1 \quad if \; \omega_k = \omega_\ast\\0 \quad if\;\omega_k \neq \omega_\ast\end{array}\right..
    \end{equation}
    \end{subequations}
    The loss $\mathcal{L}_\nu\left(\boldsymbol x\right)$ is minimized when  $\mathbb{E}_\nu(f_{\omega_k})=1$  for $\omega_k = \omega_\ast$ and $\mathbb{E}_\nu(f_{\omega_l})=0$ for $\omega_l \neq \omega_\ast$.

\begin{Ex} 
Table \ref{tab:example_DS_outputs} shows several examples, whose utilities of single-valued assignments and losses are shown in Table \ref{tab:example_utilities}. The extended utility matrix is shown in Table \ref{tab:example_utility}, and $\nu$ equals  1. We assume that $\Omega=\{\omega_1,\omega_2,\omega_3\}$ and $\omega_*=\omega_1$. Eq. \eqref{con:loss_define} yields different losses  given a set of DS layer outputs.

    \begin{table}[]
    \center
    \caption{Examples of DS layer outputs}\label{tab:example_DS_outputs}
    \begin{tabular}{ccccc}
    \hline
    Examples & \multicolumn{4}{c}{Outputs of a DS layer}         \\ \cline{2-5}
        & $m(\{\omega_1\})$ & $m(\{\omega_2\})$ & $m(\{\omega_3\})$ & $m(\Omega)$ \\ \hline
    $\#1$ & 0.70 & 0.10 & 0.10 & 0.10        \\
    $\#2$ & 0.97 & 0.01 & 0.01 & 0.01        \\
    $\#3$ & 0.50 & 0.50 & 0 & 0        \\
    $\#4$ & 0.40 & 0.40 & 0 & 0.2        \\ \hline
    \end{tabular}
    \end{table}

    \begin{table}[]
    \center
    \caption{Example of utilities and losses}\label{tab:example_utilities}
    \begin{tabular}{ccccc}
    \hline
    \multirow{2}{*}{Examples} & \multicolumn{3}{c}{Expected utility}                                                       & \multirow{2}{*}{Loss ($\omega_\ast=\omega_1$)} \\ \cline{2-4}
                          & $\mathbb{E}_1(\{\omega_1\})$ & $\mathbb{E}_1(\{\omega_2\})$ & $\mathbb{E}_1(\{\omega_3\})$ &                                                 \\ \hline
    $\#1$                     & 0.70                         & 0.10                         & 0.10                         & 0.303                                           \\
    $\#2$                     & 0.97                         & 0.01                         & 0.01                         & 0.026                                           \\
    $\#3$                     & 0.50                         & 0.50                         & 0                            & 0.602                                           \\
    $\#4$                     & 0.40                         & 0.40                         & 0                            & 0.796                                           \\ \hline
    \end{tabular}
    \end{table}
\end{Ex}
   The derivatives of $\mathcal{L}_\nu\left(\boldsymbol x\right)$ of the error w.r.t $\boldsymbol m$ in the expected utility layer are
        \begin{subequations}\label{con:derivative_eulayer}
        \begin{equation}
          \frac{\partial \mathcal{L}_\nu\left(\boldsymbol x\right)}{\partial m(\{\omega_k\})}= -\frac{y_k}{\mathbb{E}_\nu(f_{\omega_k})} \left[{\widehat u}_{\{\omega_k\},k}+(1-\nu) \max_{i=1,\dots,M}{\widehat u}_{\{\omega_k\},i}\right],
        \end{equation}
        \begin{equation}
        \begin{aligned}
          \frac{\partial \mathcal{L}_\nu\left(\boldsymbol x\right)}{\partial m(\Omega)}=&-\sum_{k=1}^M \frac{y_k}{\mathbb{E}_\nu(\{f_{\omega_k}\})} (1-\nu) \max_{i=1,\dots,M}{\widehat u}_{\{\omega_k\},i}.
        \end{aligned}
        \end{equation}
        \end{subequations}

    The derivatives of $\mathcal{L}_\nu\left(\boldsymbol x\right)$ w.r.t $p^i_k$, $\eta^i$, and $\xi^i$ in a DS layer are the same as the original work of Den{\oe}ux \cite{denoeux2000neural}:
        \begin{equation}
          \frac{\partial \mathcal{L}_\nu\left(\boldsymbol x\right)}{\partial p^i_k}=\frac{\partial \mathcal{L}_\nu\left(\boldsymbol x\right)}{\partial s^i} 2 (\eta^i)^2 s^i (x_k-p^i_k),\quad k=1,\dots,P,\ i=1,\dots,n,
        \end{equation}
        \begin{equation}
          \frac{\partial \mathcal{L}_\nu\left(\boldsymbol x\right)}{\partial \eta^i}=\frac{\mathcal{L}_\nu\left(\boldsymbol x\right)}{\partial s^i} (-2\eta^i(d^i)^2s^i),\quad i=1,\dots,n,
        \end{equation}
        and
        \begin{equation}
          \frac{\partial \mathcal{L}_\nu\left(\boldsymbol x\right)}{\partial \xi^i}=\frac{\mathcal{L}_\nu\left(\boldsymbol x\right)}{\partial s^i} \exp(-(\eta^i d^i)^2)(1-\alpha^i)\alpha^i,\quad i=1,\dots,n,
        \end{equation}
        where $P$ is the dimension of the reference patterns and the input feature vector and $n$ is the number of prototypes.

        In the proposed classifier, the DS layer is connected to the pooling layer of the last convolutional stage, as shown in Figure \ref{fig:evidential_DL_classifier}. Thus, we can compute the derivatives of the error w.r.t. $x_k$ and $po^k$ as
        \begin{equation}\label{con:derivative xe}
        \frac{\partial\mathcal{L}_\nu\left(\boldsymbol x\right)}{\partial x_k}=\frac{\mathcal{L}_\nu\left(\boldsymbol x\right)}{\partial po^k}=-2 \frac{\displaystyle\mathcal{L}_\nu\left(\boldsymbol x\right)}{\displaystyle\partial s^i} (\eta^i)^2s^i \sum_{i=1}^n(x_k-p_k^i),\quad k=1,\dots,P,
        \end{equation}
        where $po^k$ is the $k$-th output map in the final pooling layer, which is a $1 \times 1$ tensor. Error propagation in the remaining stages is performed as  in a probabilistic CNN.

    \subsection{Act selection}
    \label{sec:act_selection}
    
    As explained in Section \ref{sec:EU_layer}, the set of acts when considering multi-class assignment is $\mathcal{F}=\{f_A,A \in 2^\Omega \backslash {\emptyset}\}$, as instances can be assigned to any non-empty subset $A$ of $\Omega$. However, the cardinality of $\mathcal{F}$ increases exponentially with the number of classes, which could preclude the application of this approach when the number $M$ of classes is large.
    
    In \cite{tong2019ConvNet}, we showed that a neural network with convolutional layers and a DS layer  tends to assign samples to multi-class sets when candidate classes are similar, such as, e.g., ``cat'' and ``dog'', or ``horse'' and ``deer''. Thus, it may be advantageous to only consider partial multi-class acts assigning samples to  subsets containing two or more similar classes.
    
    In this study, we propose a strategy to determine similar classes in the frame of discernment and select partial multi-class acts from $\mathcal{F}$ based on class similarity. Using the selected partial multi-class acts, rather than all acts in $\mathcal{F}$, we can reduce the compute cost in set-valued assignments. This strategy can be described as follows.
    
    \begin{description}
     \item{Step 1:} A confusion matrix with only precise assignments is generated by a trained evidential deep-learning classifier using the training set. In the confusion matrix, each column represents the predicted sample distribution in one class.
     
     \item{Step 2:} Each column in the confusion matrix is normalized using its total number. Each normalized column is regarded as the feature of its corresponding class.
     
     \item{Step 3:} The Euclidean distance between every two features is computed, and a dendrogram is generated by a hierarchical agglomerative clustering (HAC) algorithm \cite{10.1093/comjnl/20.4.364,sibson1973slink}. The distance between every two features represents the similarity of the two classes. The distance is close to 0 if two classes are similar.
     
     \item{Step 4:} The distance can be drawn versus the number of clusters based on the dendrogram, as shown in Figure \ref{fig:ex_curve}. A point of inflection in the curve can then be used to determine the threshold for cutting the dendrogram. In this study, we used the Calinski-Harabasz index (CHI) \cite{calinski1974dendrite}  to determine this point. The point of inflection is the one in the curve with the maximum CHI, as illustrated in Figure \ref{fig:ex_curve} of Example \ref{ex:act_selection}. The right of the point has a small number of highly similar classes. This can be explained by the nature of the HAC algorithm \cite{10.1093/comjnl/20.4.364}. Very similar classes are consolidated first as the algorithm proceeds. Toward the end of the HAC run, we reach a stage where dissimilar classes are left to be merged but the distance between them is large; these classes are not similar and do not need to be clustered in the act-selection strategy. 
     
     \item{Step 5:} The distance corresponding to the inflection point is used as the threshold to cut the dendrogram. Similar patterns are the classes in the clustered groups with the distance lower than the threshold. Finally, we select the multi-class acts corresponding to similar classes.
    \end{description}
    
    \begin{Ex} \label{ex:act_selection}
     Figure \ref{fig:example_act} shows an example of act selection, in which a HAC algorithm with Ward linkage is used to generate a dendrogram. Figure \ref{fig:ex_curve} display a point of inflection whose $CHI$ is 1.91 and corresponding distance is 0.927 . The distance is used as the threshold of the Euclidean distance to cut the dendrogram. There are two pairs of similar patterns: $\{\omega_1,\omega_2\}$ and $\{\omega_3,\omega_4\}$. Thus, the selected partial multi-class acts are $f_{\{\omega_1,\omega_2\}}$ and $f_{\{\omega_3,\omega_4\}}$.
     
     \begin{figure}[htbp]
      \centering
      \subfloat[\label{tab:ex_confusion_matrix}]{
       \begin{minipage}[t]{0.5\textwidth}
        \centering
        \begin{tabular}{cccccc}
         \hline
         \multicolumn{2}{c}{\multirow{2}{*}{}} & \multicolumn{4}{c}{Labels}                        \\ \cline{3-6}
         \multicolumn{2}{c}{}                  & $\omega_1$ & $\omega_2$ & $\omega_3$ & $\omega_4$ \\ \hline
         \multirow{4}{*}{Acts}   & $f{\omega_1}$  & 557        & 115        & 24         & 13         \\
         & $f{\omega_2}$  & 107        & 679        & 32         & 14         \\
         & $f{\omega_3}$  & 13         & 16         & 663        & 128        \\
         & $f{\omega_4}$  & 25         & 32         & 145        & 627        \\ \hline
        \end{tabular}
       \end{minipage}
      }
      \subfloat[\label{tab:ex_normailized_confusion_matrix}]{
       \begin{minipage}[t]{0.5\textwidth}
        \centering
        \begin{tabular}{cccccc}
         \hline
         \multicolumn{2}{c}{\multirow{2}{*}{}} & \multicolumn{4}{c}{Labels}                        \\ \cline{3-6}
         \multicolumn{2}{c}{}                  & $\omega_1$ & $\omega_2$ & $\omega_3$ & $\omega_4$ \\ \hline
         \multirow{4}{*}{Acts}   & $f{\omega_1}$  & 0.793      & 0.136      & 0.027      & 0.017      \\
         & $f{\omega_2}$  & 0.152      & 0.806      & 0.037      & 0.018      \\
         & $f{\omega_3}$  & 0.018      & 0.019      & 0.767      & 0.167      \\
         & $f{\omega_4}$  & 0.035      & 0.038      & 0.168      & 0.802      \\ \hline
        \end{tabular}
       \end{minipage}
      }\\
      \subfloat[]{
       \begin{minipage}[t]{0.5\textwidth}
        \centering
        \includegraphics[width=0.9\textwidth]{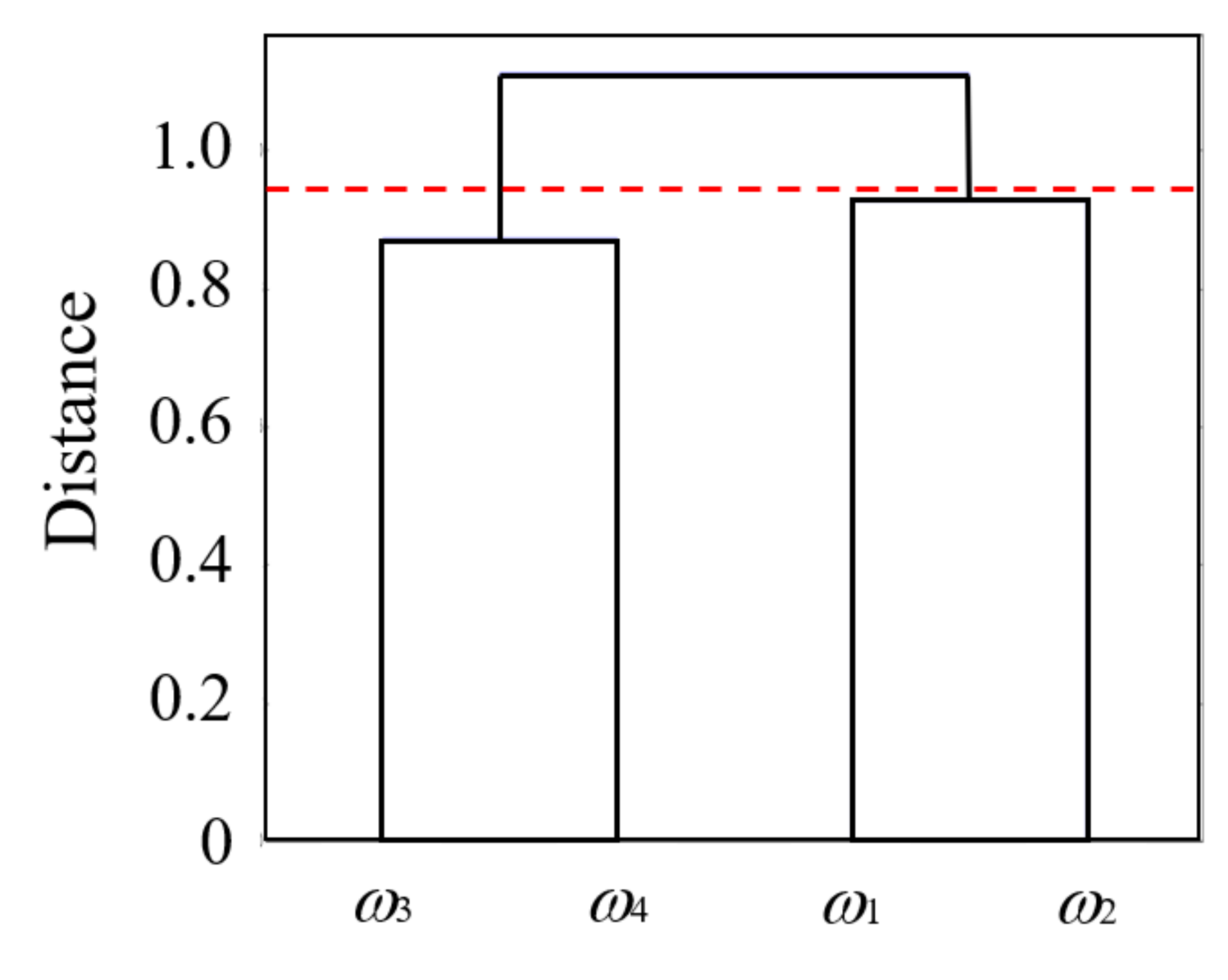}
       \end{minipage}
      }
      \subfloat[\label{fig:ex_curve}]{
       \begin{minipage}[t]{0.5\textwidth}
        \centering
        \includegraphics[width=\textwidth]{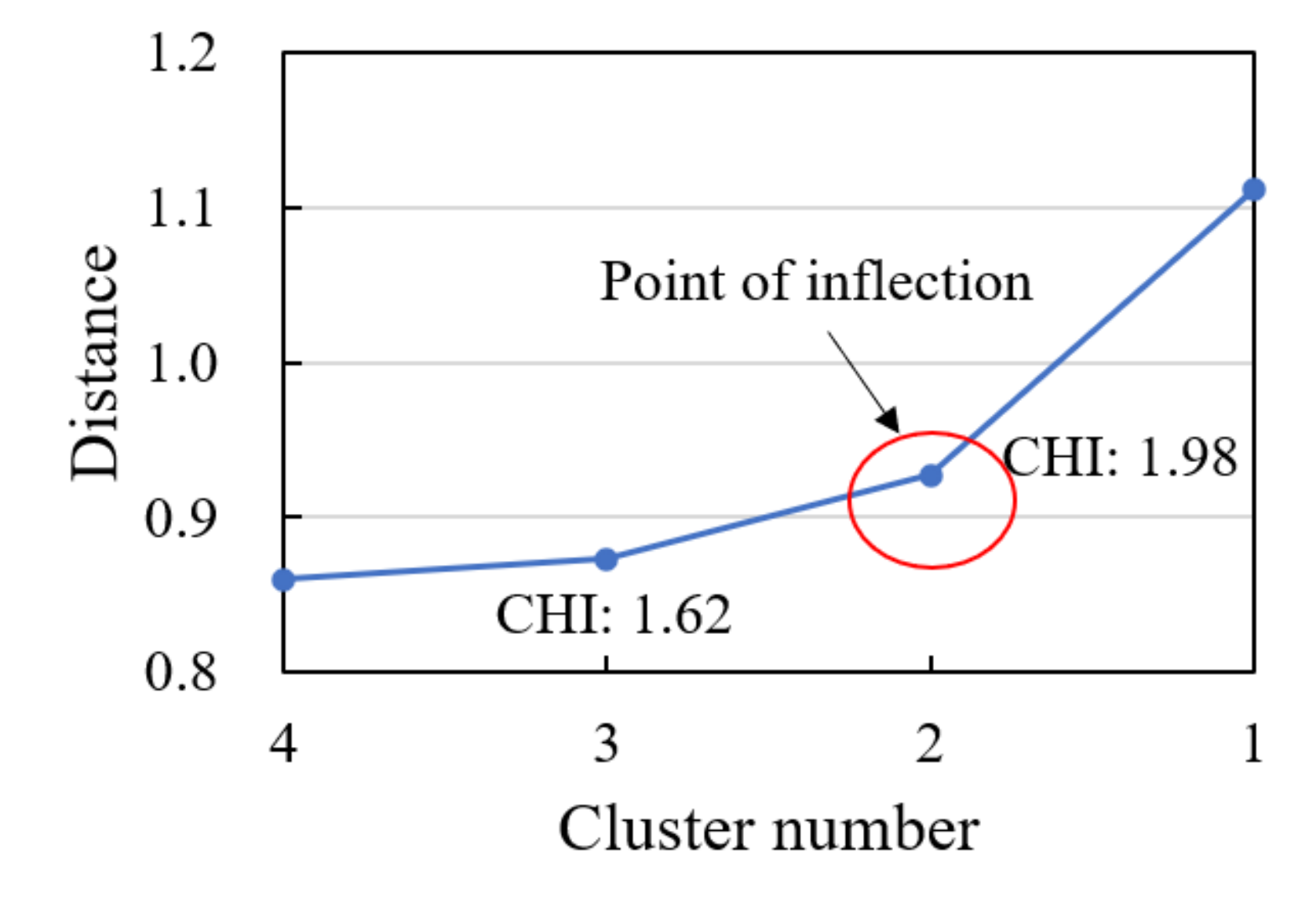}
       \end{minipage}
      }
      \caption{An example of act selection: confusion matrix (a), normalized confusion matrix (b), dendrogram (c), and a curve of distance vs. cluster number (d).}\label{fig:example_act}
     \end{figure}
    \end{Ex}

\section{Experiments}
\label{sec:experimental_results}

In this section, we present  numerical experiments demonstrating the advantages of the proposed classifier. In section \ref{sec:evaluation_metric}, we provide three metrics for performance evaluation. Experimental results on image recognition, signal processing and semantic-relationship classification tasks are then reported and discussed, respectively, in  Sections \ref{sec:image_classification_experiment}, \ref{sec:signal_classification_experiment} and \ref{sec:semantic_classification_experiment}.

    \subsection{Evaluation of set-valued classification}
    \label{sec:evaluation_metric}

    In the applications of evidential deep-learning classifiers, we use the extended utility matrix $\mathbb{U}_{(2^\Omega-1) \times M}$ for performance evaluation. For a dataset $T$, the classification performance is evaluated by the \emph{averaged utility} as
    \begin{equation}\label{con:averaged_utility}
      AU(T)=\frac1{|T|}\sum_{i=1}^{|T|}{\widehat u}_{A(i),y_i},
    \end{equation}
    where $y_i$ is the true class of learning example $i$, $A(i)$ is the selected subset for example $i$ using \eqref{eq:decision}    and, using the notation introduced in Section \ref{sec:EU_layer}, ${\widehat u}_{A,y_i}$ is the utility of assigning sample $i$ to subset $A\subseteq \Omega$ when its true class is $y_i$. When only considering precise acts, the $AU$ criterion defined by \eqref{con:averaged_utility} boils down to classification accuracy. \emph{The averaged cardinality} is also computed as
    \begin{equation}\label{con:averaged_cardinality}
      AC(T)=\frac1{|T|}\sum_{i=1}^{|T|}|A(i)|.
    \end{equation}
    
    Additionally, we also consider the case where a dataset  $T'=\{T'_O,T'_I\}$ is composed of a subset $T'_O$ of outliers whose class does not belong to the frame of discernment $\Omega$, and a subset $T'_I$ of inliers whose class belongs to $\Omega$. We compare the rate of $f_\Omega$ in $T'_I$ and $T'_O$ to evaluate the capacity of a classifier to reject outliers together with ambiguous samples.  This capacity is  called \emph{novelty detection} in \cite{denoeux2000neural}. Generally, a well-trained classifier is expected to have a low rate of $f_\Omega$ in $T'_I$ but a high rate in $T'_O$.

    In this study, we compare the proposed classifiers with probabilistic CNNs. To ensure a fair comparison, we adopt the following strategy for probability-based set-valued classification in CNNs: $f_{A}\succeq_\ast f_{A'}$ if and only if $\mathbb{E}(f_{A})\leq \mathbb{E}(f_{A'})$, with $\mathbb{E}(f_{A})=\sum_{\omega_k \in A} p(\omega_k)\cdot{\widehat u}_{A,k}$.

    \subsection{Image classification experiment}
    \label{sec:image_classification_experiment}

    We used the CIFAR-10 dataset to evaluate the performance of the proposed classifier in image classification. The CIFAR-10 dataset \cite{krizhevsky2009learning} consists of 60,000 RGB images of size $32\times 32$ partitioned in 10 classes. There are 50,000 training examples and 10,000 testing examples. During training, we randomly selected 10,000 images as validation data. We used two datasets (CIFAR-100 \cite{krizhevsky2009learning} and MNIST \cite{726791}) for novelty detection. The CIFAR-100 dataset is just like the CIFAR-10 except it has 100 classes containing 600 images each, while the MNIST dataset of handwritten digits has 70,000 examples. All examples in the two datasets are used for novelty detection except some images whose classes are included in the CIFAR-10 dataset.

 \paragraph{Precise classification}   In this experiment, the convolutional stages of three probabilistic CNNs were combined with the DS and expected utility layers, as shown in Table \ref{tab:stage_CIFAR-10}. The three probabilistic CNNs have the same number of output feature maps but  different convolutional and pooling layers. As shown in Table \ref{tab:test_average_utilities_CIFAR-10}, the proposed classifiers slightly outperform the probabilistic ones in precise classification, except with ViT-L/16 feature extraction. McNemar's test results indicate a small but statistically significant effect of the proposed combination on the image classification task with $p$-values below 5\%. These results suggest that the utility of an evidential classifier is larger than that of a probabilistic CNN classifier with the same stage as the evidential one. They also demonstrate that the use of the convolutional and pooling layers in Step 1 of Section \ref{sec:network_architecture} allows for good precise-classification performance of the evidential deep-learning classifier.
    
    \begin{table}[]
     \centering
     \caption{The three baseline stages used on CIFAR-10 data.}\label{tab:stage_CIFAR-10}
     \resizebox{\textwidth}{!}{
     \begin{tabular}{ccc}
      \hline
      \multicolumn{1}{c|}{NIN \cite{lin2013network}}                       & \multicolumn{1}{c|}{FitNet-4  \cite{mishkin2015all}}                                & ViT-L/16 \cite{dosovitskiy2020image}                                                               \\ \hline
      & Input: 32 $\times$ 32 $\times$ 3                             &                                                                        \\ \hline
      \multicolumn{1}{c|}{}                                         & \multicolumn{1}{c|}{}                                        & 16 $\times$ 16 $\times$ 3  $\times$ 4 patches with positional encoding \\ \hline
      \multicolumn{1}{c|}{5 $\times$ 5 Conv. NIN 64 $ReLU$}         & \multicolumn{1}{c|}{3 $\times$ 3 Conv. 32 $ReLU$}            & 3 $\times$ 3 Conv. 32 $ReLU$                                           \\
      \multicolumn{1}{c|}{}                                         & \multicolumn{1}{c|}{3 $\times$ 3 Conv. 32 $ReLU$}            & 3 $\times$ 3 Conv. 32 $ReLU$                                           \\
      \multicolumn{1}{c|}{}                                         & \multicolumn{1}{c|}{3 $\times$ 3 Conv. 32 $ReLU$}            & 3 $\times$ 3 Conv. 32 $ReLU$                                           \\
      \multicolumn{1}{c|}{}                                         & \multicolumn{1}{c|}{3 $\times$ 3 Conv. 48 $ReLU$}            & 3 $\times$ 3 Conv. 48 $ReLU$                                           \\
      \multicolumn{1}{c|}{}                                         & \multicolumn{1}{c|}{3 $\times$ 3 Conv. 48 $ReLU$}            & 3 $\times$ 3 Conv. 48 $ReLU$                                           \\ \hline
      & 2 $\times$ 2 max-pooling with 2 strides                      &                                                                        \\ \hline
      \multicolumn{1}{c|}{5 $\times$ 5 Conv. NIN 64 $ReLU$}         & \multicolumn{1}{c|}{3 $\times$ 3 Conv. 80 $ReLU$}            & 3 $\times$ 3 Conv. 80 $ReLU$                                           \\
      \multicolumn{1}{c|}{2 $\times$ 2 mean-pooling with 2 strides} & \multicolumn{1}{c|}{3 $\times$ 3 Conv. 80 $ReLU$}            & 3 $\times$ 3 Conv. 80 $ReLU$                                           \\
      \multicolumn{1}{c|}{}                                         & \multicolumn{1}{c|}{3 $\times$ 3 Conv. 80 $ReLU$}            & 3 $\times$ 3 Conv. 80 $ReLU$                                           \\
      \multicolumn{1}{c|}{}                                         & \multicolumn{1}{c|}{3 $\times$ 3 Conv. 80 $ReLU$}            & 3 $\times$ 3 Conv. 80 $ReLU$                                           \\
      \multicolumn{1}{c|}{}                                         & \multicolumn{1}{c|}{3 $\times$ 3 Conv. 80 $ReLU$}            & 3 $\times$ 3 Conv. 80 $ReLU$                                           \\ \hline
      & 2 $\times$ 2 max-pooling with 2 strides                      &                                                                        \\ \hline
      \multicolumn{1}{c|}{5 $\times$ 5 Conv. NIN 128 $ReLU$}        & \multicolumn{1}{c|}{3 $\times$ 3 Conv. 128 $ReLU$}           & 3 $\times$ 3 Conv. 128 $ReLU$                                          \\
      \multicolumn{1}{c|}{2 $\times$ 2 mean-pooling with 2 strides} & \multicolumn{1}{c|}{3 $\times$ 3 Conv. 128 $ReLU$}           & 3 $\times$ 3 Conv. 128 $ReLU$                                          \\
      \multicolumn{1}{c|}{}                                         & \multicolumn{1}{c|}{3 $\times$ 3 Conv. 128 $ReLU$}           & 3 $\times$ 3 Conv. 128 $ReLU$                                          \\
      \multicolumn{1}{c|}{}                                         & \multicolumn{1}{c|}{3 $\times$ 3 Conv. 128 $ReLU$}           & 3 $\times$ 3 Conv. 128 $ReLU$                                          \\
      \multicolumn{1}{c|}{}                                         & \multicolumn{1}{c|}{3 $\times$ 3 Conv. 128 $ReLU$}           & 3 $\times$ 3 Conv. 128 $ReLU$                                          \\
      \multicolumn{1}{c|}{}                                         & \multicolumn{1}{c|}{8 $\times$ 8 max-pooling with 2 strides} & 4 $\times$ 4 max-pooling with 2 strides+positional encoding            \\ \hline
      \multicolumn{1}{c|}{Average global pooling}                   & \multicolumn{1}{c|}{}                                        & Transformer decoder                                                    \\ \hline
     \end{tabular}}
    \end{table}

    \begin{table}[]
     \centering
     \caption{Test average utilities in precise classification on CIFAR-10 data.}\label{tab:test_average_utilities_CIFAR-10}
     \resizebox{\textwidth}{!}{
      \begin{tabular}{ccccccc}
       \hline
       \multirow{2}{*}{Models}                                                                    & \multicolumn{2}{c}{NIN \cite{lin2013network}} & \multicolumn{2}{c}{FitNet-4  \cite{mishkin2015all}} & \multicolumn{2}{c}{ViT-L/16  \cite{dosovitskiy2020image}} \\ \cline{2-7} 
       & Probabilistic classifier                & Evidential classifier               & Probabilistic classifier          & Evidential classifier         & Probabilistic classifier         & Evidential classifier        \\ \hline
       Utility                                                                                    & 0.8959                        & 0.8978                              & 0.9353                  & 0.9361                        & 0.9921                 & 0.9908                       \\
       \begin{tabular}[c]{@{}c@{}}\emph{p}-value\\ (McNemar's test)\end{tabular} & \multicolumn{2}{c}{0.0489}                                          & \multicolumn{2}{c}{0.0492}                              & \multicolumn{2}{c}{0.0452}                            \\ \hline
     \end{tabular}}
    \end{table}

 \begin{table}[]
  \centering
  \caption{Test average utilities for precise classification of the CIFAR-100 data after transfer learning.}\label{tab:transfer_learning_CIFAR-10}
  \resizebox{\textwidth}{!}{
   \begin{tabular}{ccccccc}
    \hline
    \multirow{2}{*}{Models}                                                                    & \multicolumn{2}{c}{NIN \cite{lin2013network}} & \multicolumn{2}{c}{FitNet-4  \cite{mishkin2015all}} & \multicolumn{2}{c}{ViT-L/16  \cite{dosovitskiy2020image}} \\ \cline{2-7} 
    & CNN classifier                & Evidential classifier               & Probabilistic classifier          & Evidential classifier         & Probabilistic classifier         & Evidential classifier        \\ \hline
    Utility                                                                                    & 0.3442                        & 0.3461                              & 0.6688                  & 0.6714                        & 0.8251                 & 0.8217                       \\\hline
  \end{tabular}}
 \end{table}
 
\paragraph{Transfer learning} The feasibility of transfer learning on the proposed classifier was also verified in this study. The three evidential deep-learning classifiers trained on the CIFAR-10 classification task, as well as the three probabilistic CNNs, were fine-tuned using the training set of the CIFAR-100 dataset as a new task. Table \ref{tab:transfer_learning_CIFAR-10} shows the testing utilities of fine-tuned classifiers on the CIFAR-100 dataset. The evidential and probabilistic classifiers achieve  close results for precise classification after fine-tuning. Besides, the average utilities of the evidential deep-learning classifiers are close to those already reported in \cite{lin2013network,mishkin2015all,dosovitskiy2020image}. This demonstrates the feasibility of  transfer learning with the proposed classifiers.

\begin{figure}
\centering
\subfloat[\label{fig:cifar_nu_utility_model1}]{\includegraphics[width=0.5\textwidth]{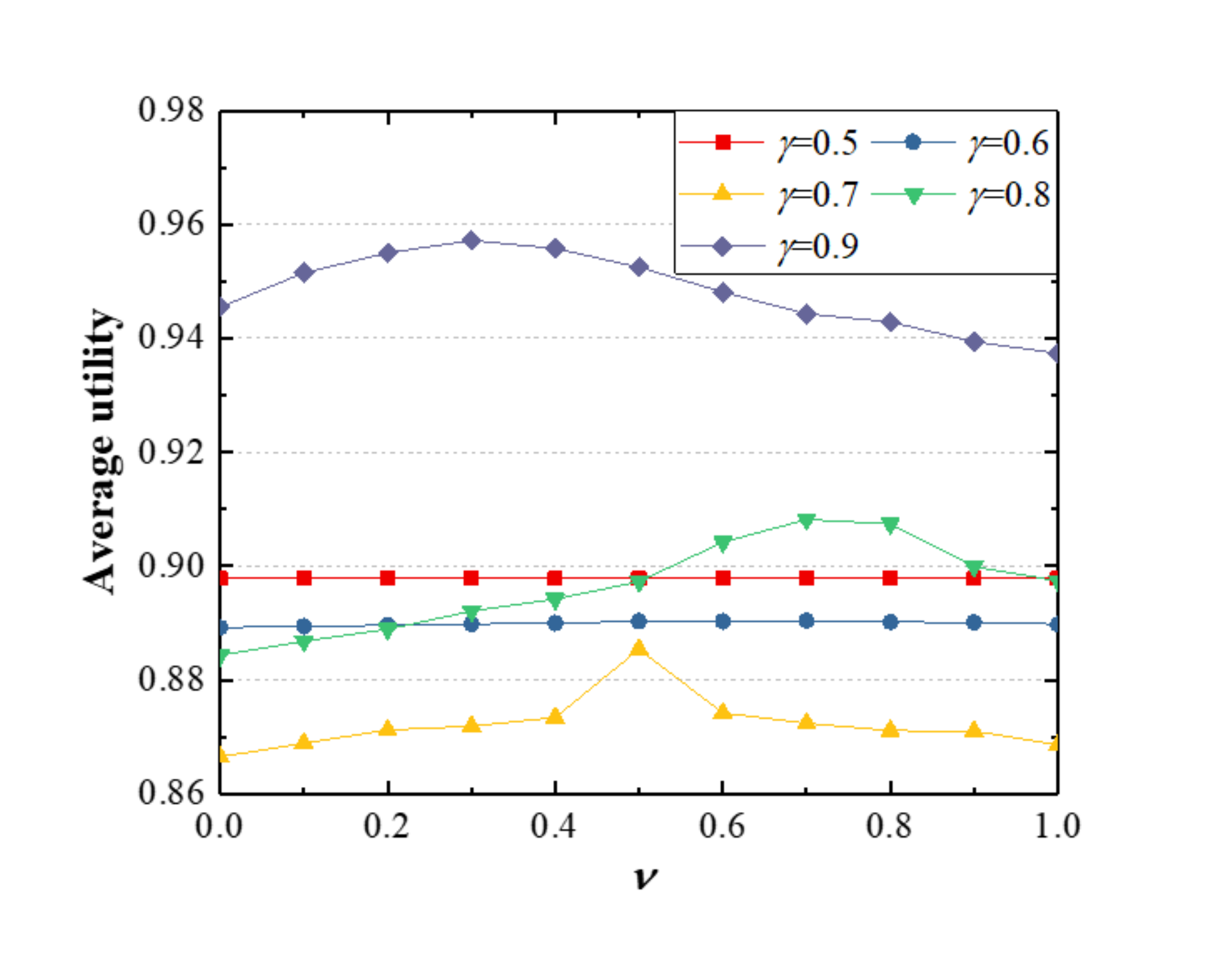}}
\subfloat[\label{fig:cifar_nu_utility_model2}]{\includegraphics[width=0.5\textwidth]{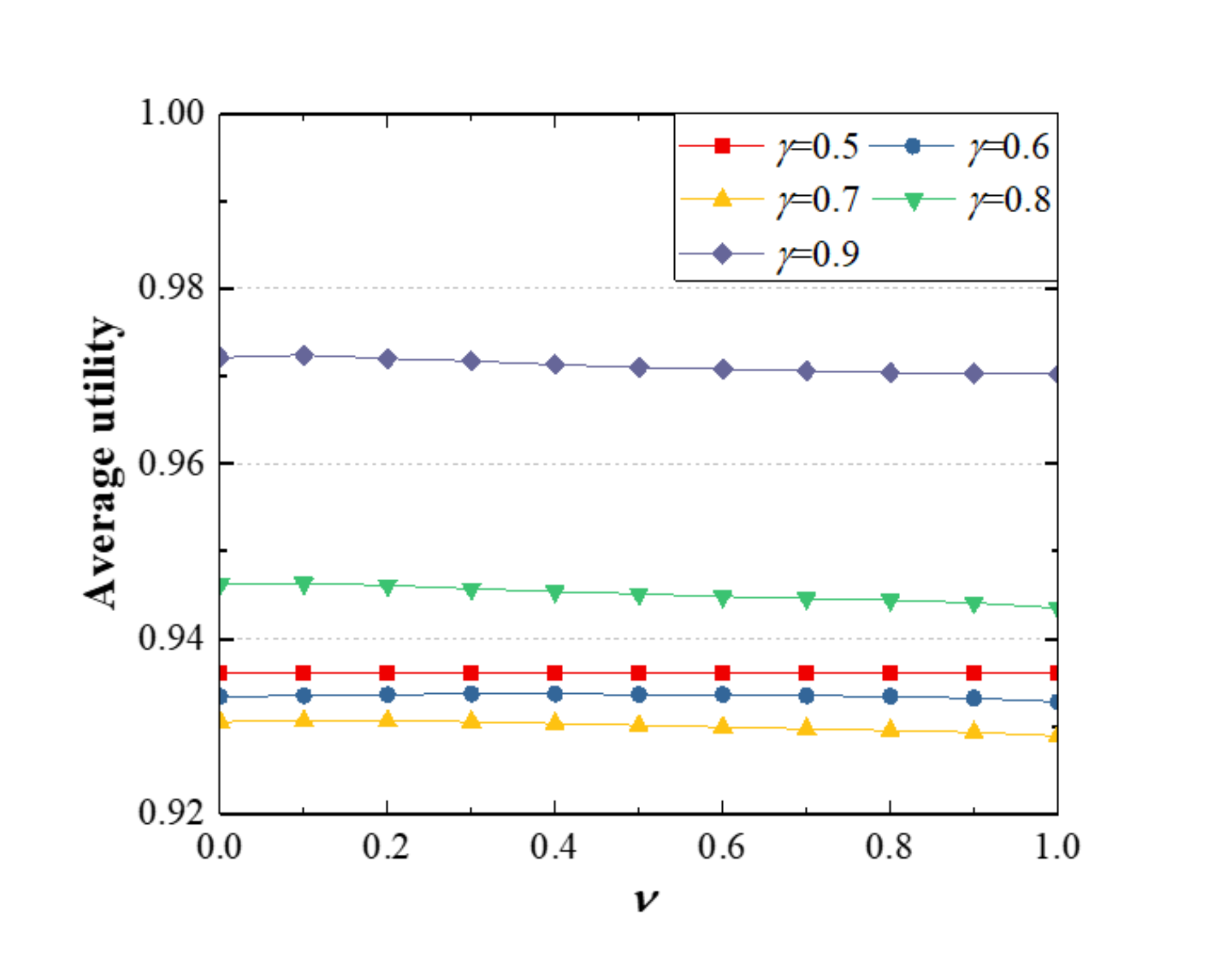}}\\
\subfloat[\label{fig:cifar_nu_utility_model3}]{\includegraphics[width=0.5\textwidth]{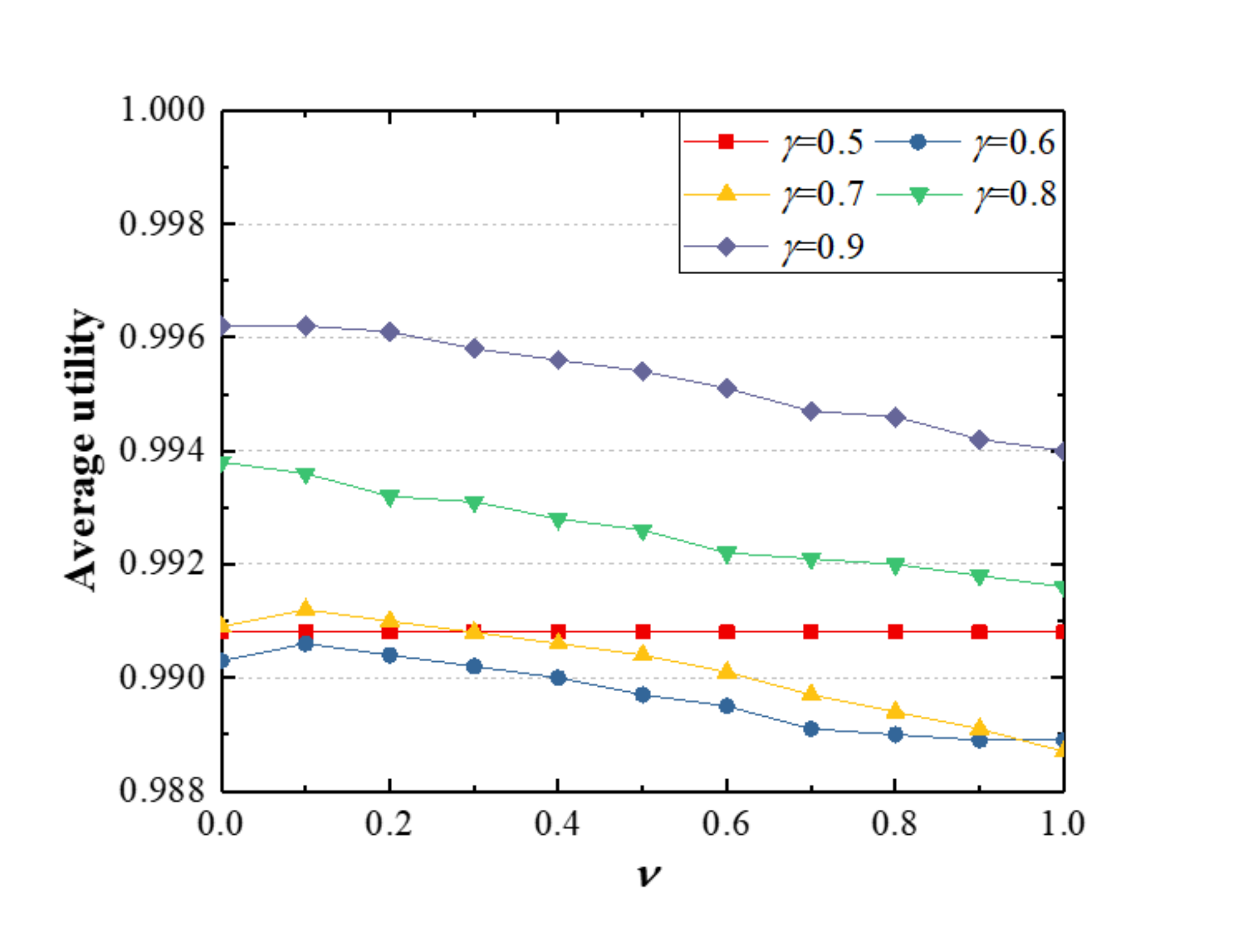}}
\caption{Average utility vs.  $\nu$  for the proposed classifiers on the CIFAR-10 dataset: NIN (a), FitNet-4 (b), and ViT-L/16 (c).}\label{fig:cifar_alpha_utility}
\end{figure}

 \paragraph{Set-valued classification}   Before evaluating the performance of the proposed classifiers in set-valued assignments, we need to determine the optimal pessimism index $\nu$ in Eq. \eqref{con:Hurwicz} once given a value of imprecision tolerance degree $\gamma$. Based on the $\nu$-utility curves on the training set (Figure \ref{fig:cifar_alpha_utility}), we can determine the optimal $\nu$ for any given  $\gamma$. As we consider all of the $2^{|\Omega|}$ acts, the three proposed classifiers always achieve average utilities of 1 when $\gamma$ equals 1. The value of $\nu$ has an apparent effect on the average utilities when $\gamma$ is higher than 0.7. These curves show  that parameter $\nu$ should be carefully tuned to ensure optimal performance of the proposed classifier in set-valued assignments.

\begin{figure}
\centering
\subfloat[\label{fig:cifar_gamma_utility_model1}]{\includegraphics[width=0.50\textwidth]{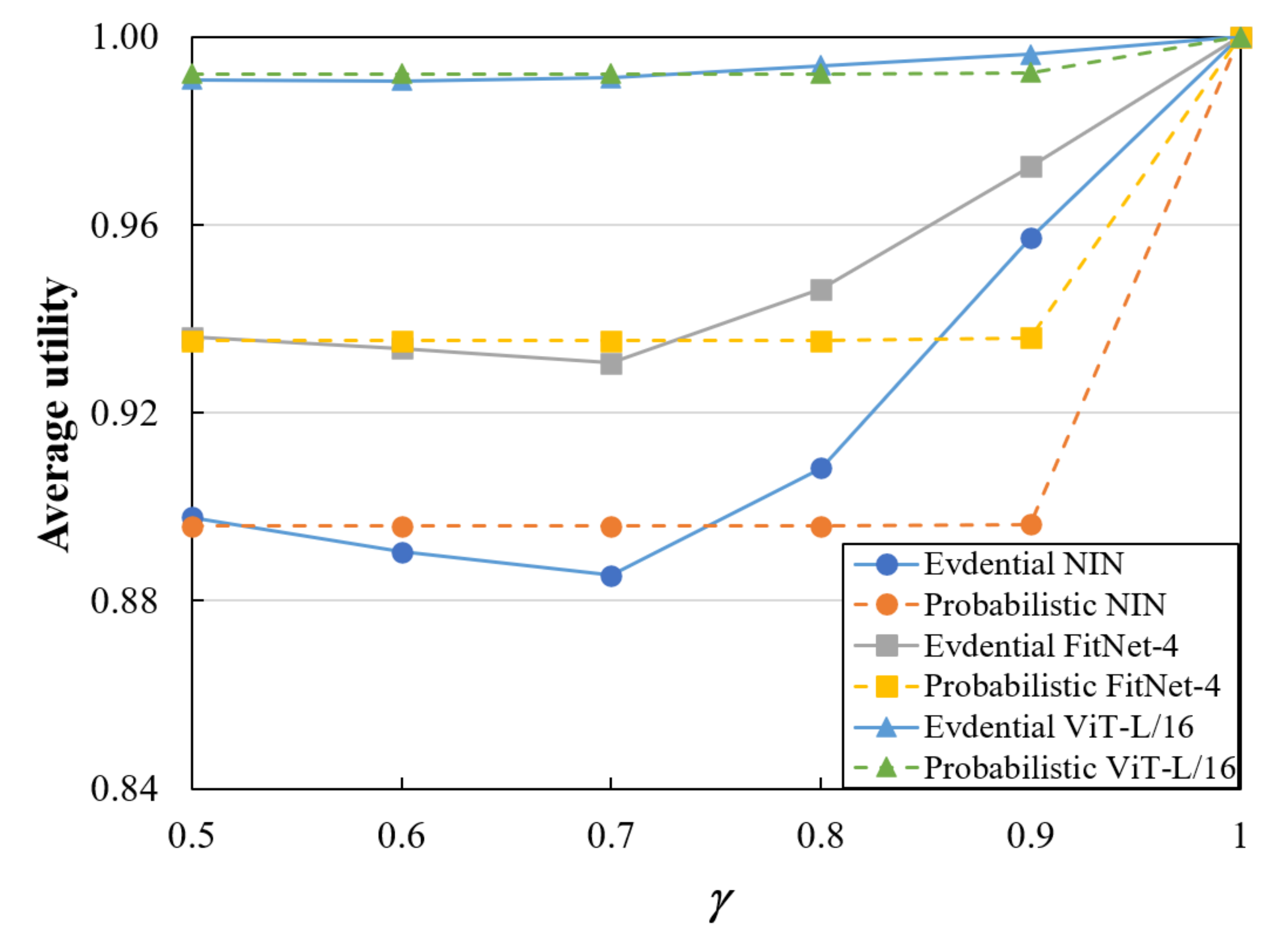}}
\subfloat[\label{fig:cifar_gamma_utility_model2}]{\includegraphics[width=0.50\textwidth]{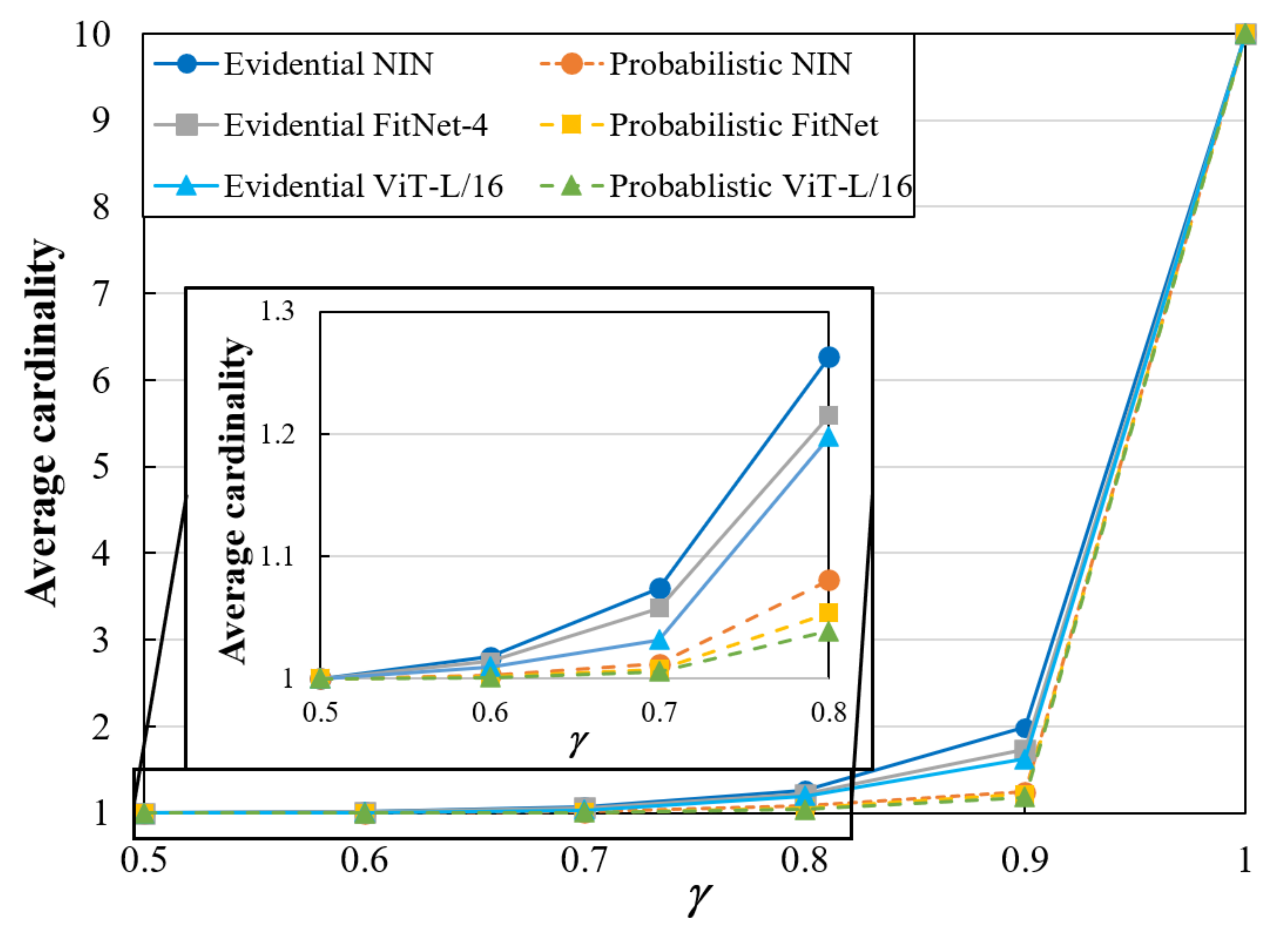}}\\
\caption{Average utility (a) and average cardinality (b) vs.  $\gamma$ for the evidential and probabilistic deep-learning classifiers on the CIFAR-10 dataset.}\label{fig:cifar_gamma_utility}
\end{figure}

    Figure \ref{fig:cifar_gamma_utility} shows the test average utilities and cardinalities of the evidential deep-learning classifiers as functions of $\gamma$ with the optimal $\nu$. When the imprecision tolerance degree increases, the average cardinalities increase. This indicates that the proposed classifiers tend to perform set-valued assignments when given a large tolerance degree of imprecision. The test average utilities decrease slightly and then increase when $\gamma$ increases. To explain this behavior, Table \ref{tab:label_classification} provides four examples with their assignments and corresponding utilities. For the first example, the utility increases from 0 to 1 as $\gamma$ becomes larger. However, for examples correctly classified when $\gamma=0.5$ (\#2 and \#3), their utilities first decrease and then increase back to 1. The majority of examples in the CIFAR-10 testing set fall in the latter category. Therefore, the test average utilities decrease slightly and then increase when $\gamma$ increases from 0.5 to 1.

    The use of the DS and expected utility layers has an effect when there is a lack of evidence in the feature-extraction part. In Figure \ref{fig:cifar_gamma_utility}, when $\gamma$ is increased from 0.5 to 0.9, the largest gains in  average utility are obtained by the evidential classifier with the NIN stages \cite{lin2013network}, whose feature extraction was found to be the worst among the three proposed classifiers since it achieved the minimum utility in the precise assignments (Table \ref{tab:test_average_utilities_CIFAR-10}). Thus, the classifier with the NIN stages is more affected by the DS and expected utility layers than the other two classifiers. Therefore, we can conclude that the effects of DS and expected utility layers are more significant if there is a lack of evidence in the feature extraction part.

    As shown in Figure \ref{fig:cifar_gamma_utility}, the proposed model with a DS layer and an expected layer outperforms probabilistic CNN classifiers for set-valued classification. The average utilities of the proposed classifiers increase significantly when $\gamma$ increases from 0.5 to 0.9. In contrast, the average utilities of the probabilistic CNN classifiers only increase sharply when $\gamma$ increases from 0.9 to 1.0. This is evidence that the proposed classifiers make well-distributed set-valued classification based on the user's tolerance degree of imprecision, while the probabilistic CNN classifiers only assign samples to the multi-class sets when the tolerance is large. This phenomenon is caused by the use of DS and expected utility layers in the proposed classifiers.  The DS layer tends to generate uniformly distributed masses when the features are not informative. As a result, the expected utility of a set-valued classification is the maximum among all acts, rather than the utility of a precise classification. This effect explains the superiority of the proposed approach for set-valued classification. However, the average utilities of the evidential classifiers are less than those of the probabilistic CNN classifiers for $\gamma=0.7$. The reason is that the probabilistic CNN classifiers make few set-valued assignments for $\gamma=0.7$, and the evidential classifiers are so cautious that they perform set-valued assignments for some instances that are correctly classified when $\gamma$ is less than 0.7, such as \#2 and \#3 in Table \ref{tab:label_classification}.

 
\begin{table}[]
\caption{Label classification/utilities with different $\gamma$.}
\label{tab:label_classification}
\resizebox{\textwidth}{!}{
\begin{tabular}{ccccc}
\hline
             & \#1($\omega^\ast=$cat) & \#2($\omega^\ast$=dog) & \#3($\omega^\ast=$deer) &\#4($\omega^\ast=$automobile) \\\hline
$\gamma$=0.5 & \{dog\}/0                  & \{dog\}/1                &\{deer\}/1                &\{airplane\}/0 \\
$\gamma$=0.6 & \{cat,dog\}/0.6             & \{cat,dog\}/0.6             &\{deer\}/1                  &\{airplane\}/0 \\
$\gamma$=0.7 & \{cat,dog\}/0.7   & \{cat,dog\}/0.7   &\{deer,horse\}/0.7                   &\{airplane\}/0        \\
$\gamma$=0.8 & \{cat,dog\}/0.8   &  \{cat,dog\}/0.8  &\{deer,horse\}/0.8                   &\{airplane\}/0       \\
$\gamma$=0.9 & \{cat,dog\}/0.9   & \{cat,dog\}/0.9   &\{cat,deer,dog,horse\}/0.7104                   &\{cat,deer,dog,horse\}/0        \\
$\gamma$=1.0 & $\Omega$/1.0      & $\Omega$/1.0      &$\Omega$/1.0       &$\Omega$/1.0 \\\hline
             & \includegraphics{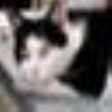} & \includegraphics{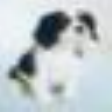} & \includegraphics{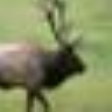} & \includegraphics{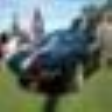} \\\hline
\end{tabular}}
\end{table}

\begin{figure}
 \centering
 \subfloat[]{\includegraphics[width=0.5\textwidth]{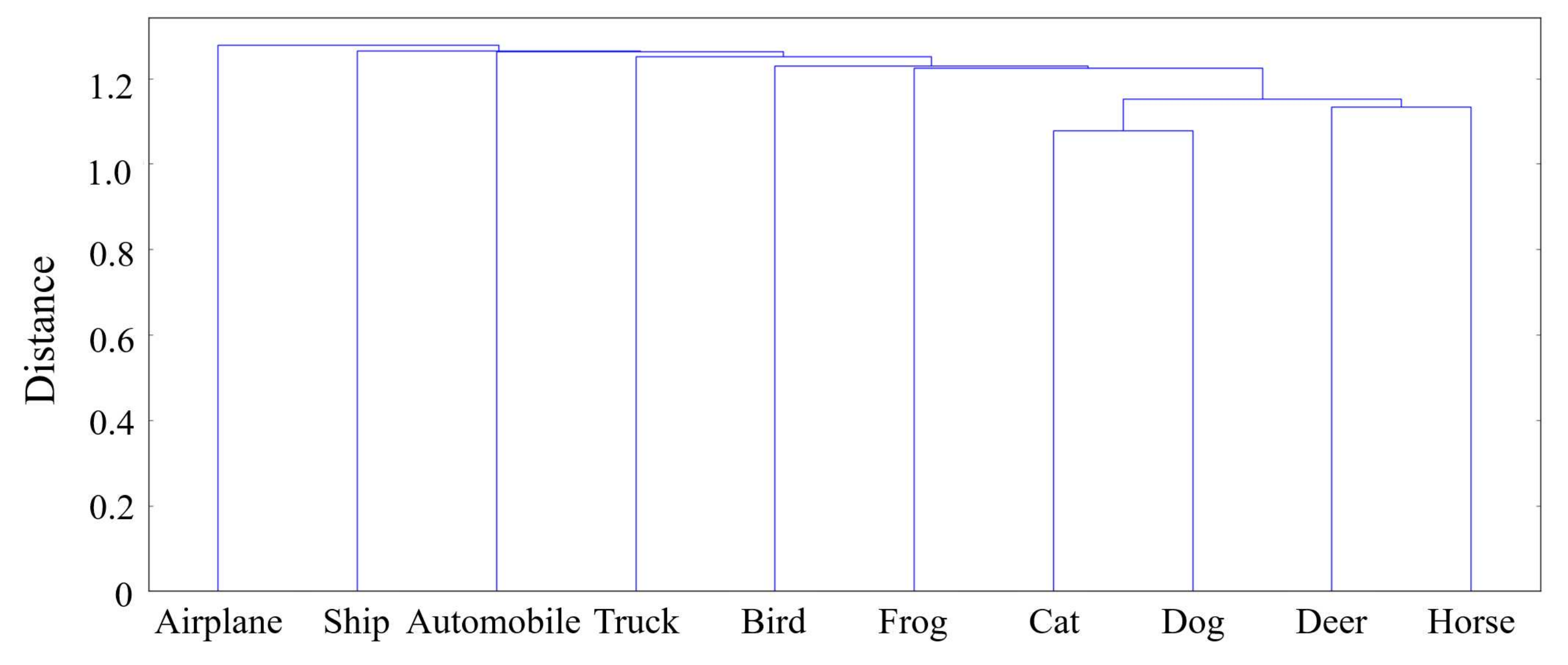}}
 \subfloat[]{\includegraphics[width=0.5\textwidth]{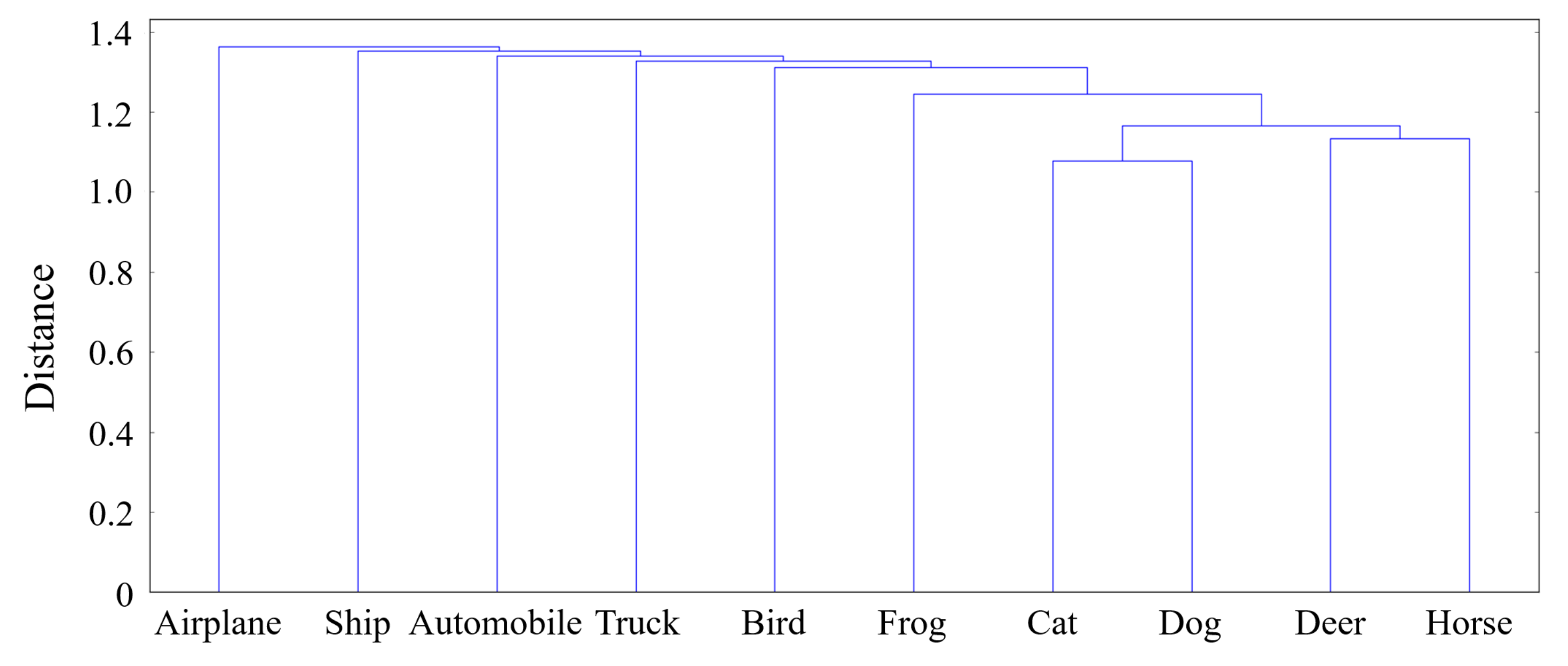}}\\
 \subfloat[]{\includegraphics[width=0.5\textwidth]{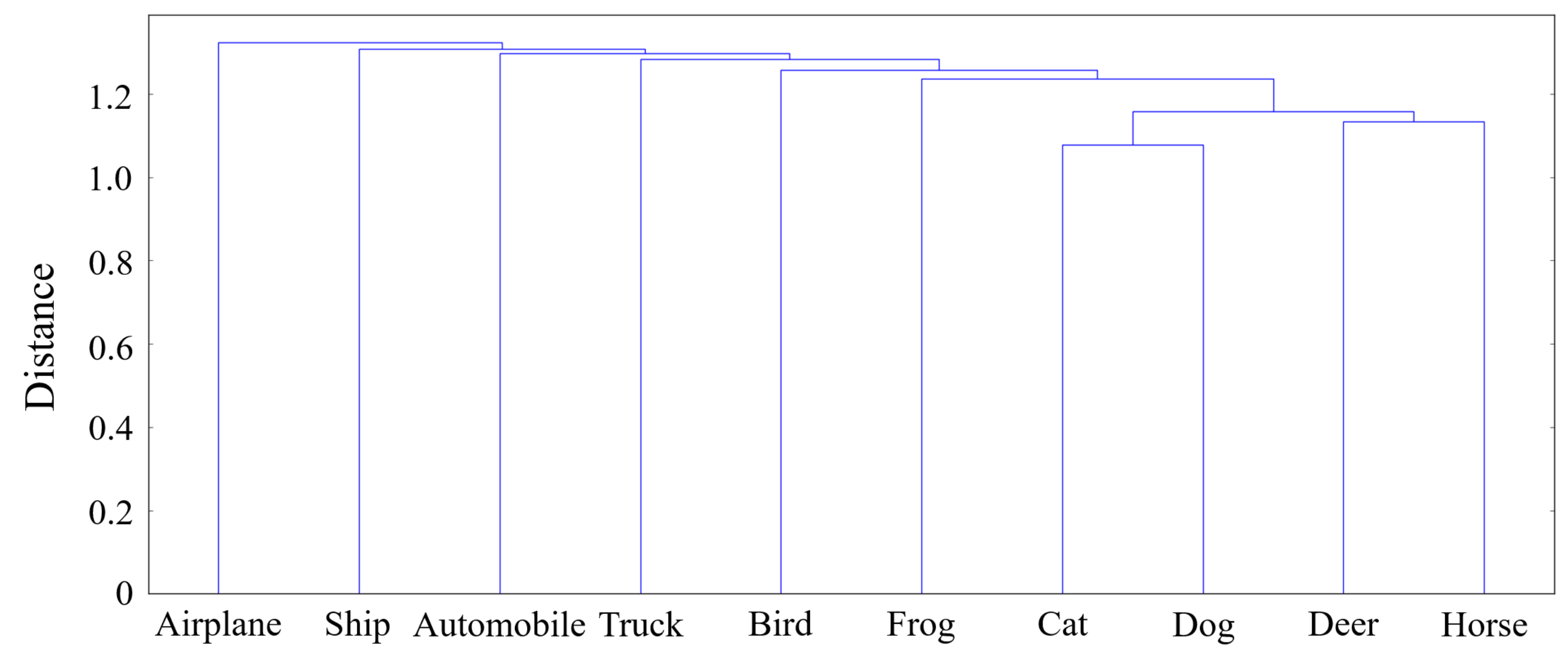}}
 \subfloat[]{\includegraphics[width=0.5\textwidth]{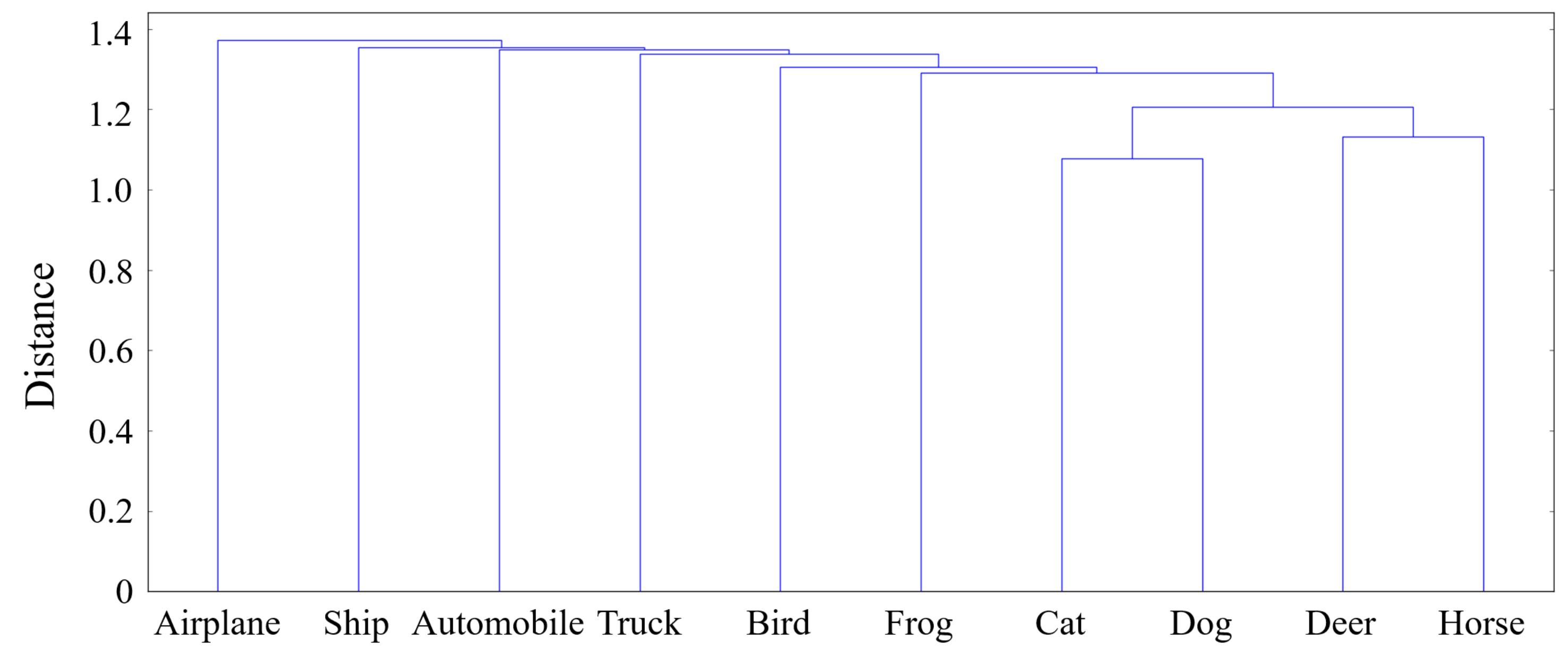}}
 \caption{Dendrograms for the CIFAR-10 dataset: single linkage (a), complete linkage (b), average linkage (c) , and Ward linkage (d).}\label{fig:dendrograms_cifar}
\end{figure}

In \cite{tong2019ConvNet}, we found that some ambiguous patterns always led the incorrect classification. Thus, we do not need to consider all of the $2^\Omega$ acts, as mentioned in Section \ref{sec:act_selection}. In this experiment, the performances of the classifiers with partial acts are compared to those with all $2^\Omega$ acts. Taking the evidential classifier with a network as in \cite{dosovitskiy2020image} as an example, we used the strategy introduced in Section \ref{sec:act_selection} to generate the dendrograms, as shown in Figure \ref{fig:dendrograms_cifar}. When using Ward linkage \cite{ward1963hierarchical}, we get an inflection point to cut the dendrogram, with the CHI equal to 1.286 and the corresponding distance equal to 1.238.  The selected multi-class sets consist of $\{cat,dog\}$, $\{deer,horse\}$, $\{cat,dog,deer,horse\}$, and $\{cat,dog,deer,horse,frog\}$ in the comparison study. Table \ref{tab:rates_set_valued_assignments} reports the testing rates of  set-valued classification using the selected and $2^\Omega$ acts. The rates of the classifiers with the selected and $2^\Omega$ acts are close when $\gamma$ is less than 0.9. Besides, the rates of the samples assigned correctly using $2^\Omega$ acts but incorrectly using the selected acts are small when $\gamma$ is less than 0.9, as shown in Table \ref{tab:rates_of_incorrect}. A set-valued assignment is regarded as correct if the multi-class set contains the true label. Thus, the proposed strategy is useful once an evidential classifier has a value of $\gamma$ in the range of 0.5-0.9.

\begin{table}[]
 \centering
 \caption{Set-valued assignment rates using the selected and $2^\Omega$ acts (unit:\%).}\label{tab:rates_set_valued_assignments}
 \begin{tabular}{cccccccc}
  \hline
  \multicolumn{2}{c}{$\gamma$}                           & 0.5 & 0.6  & 0.7  & 0.8   & 0.9   & 1     \\ \hline
  \multirow{2}{*}{CIFAR-10}            & Selected acts   & 0   & 0.52 & 1.74 & 13.24 & 19.62 & 52.04 \\
  & $2^\Omega$ acts & 0   & 0.52 & 1.76 & 14.21 & 22.67 & 100   \\ \hline
  \multirow{2}{*}{UrbanSound 8K}       & Selected acts   & 0   & 2.47 & 9.10 & 23.96 & 49.91 & 64.43\\
  & $2^\Omega$ acts & 0   & 2.47 & 9.71 & 28.74 & 55.62 & 100   \\ \hline
  \multirow{2}{*}{SemEval-2010 Task 8} & Selected acts   & 0   & 1.69 & 8.11 & 17.62 & 43.11 & 66.62\\
  & $2^\Omega$ acts & 0   & 1.69 & 8.57 & 27.71 & 52.77 & 100   \\ \hline
 \end{tabular}
\end{table}

\begin{table}[]
 \centering
 \caption{Proportions of samples correctly assigned to acts in $2^\Omega$  and incorrectly assigned to selected acts, for different values of $\gamma$.}\label{tab:rates_of_incorrect}
 \begin{tabular}{ccccccc}
  \hline
  $\gamma$            & 0.5 & 0.6 & 0.7  & 0.8  & 0.9  & 1    \\ \hline
  CIFAR-10            & 0   & 0   & 0    & 0.18 & 0.47 & 2.87 \\
  UrbanSound 8K       & 0   & 0   & 0    & 0.42 & 0.95 & 6.62 \\
  SemEval-2010 Task 8 & 0   & 0   & 0.11 & 0.48 & 0.74 & 4.43 \\ \hline
 \end{tabular}
\end{table}

\begin{figure}
\centering
\subfloat[]{\includegraphics[width=0.5\textwidth]{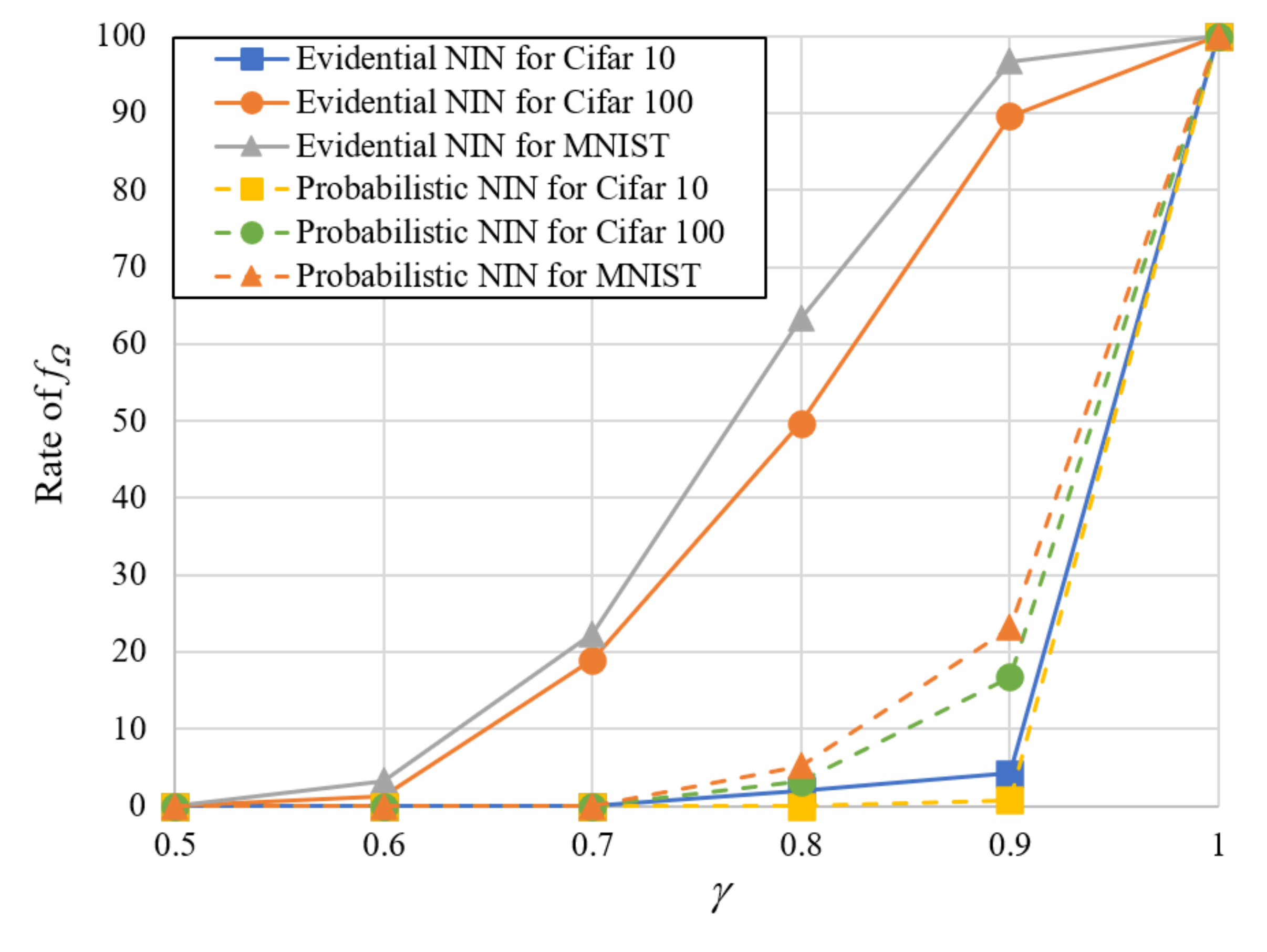}}
\subfloat[]{\includegraphics[width=0.5\textwidth]{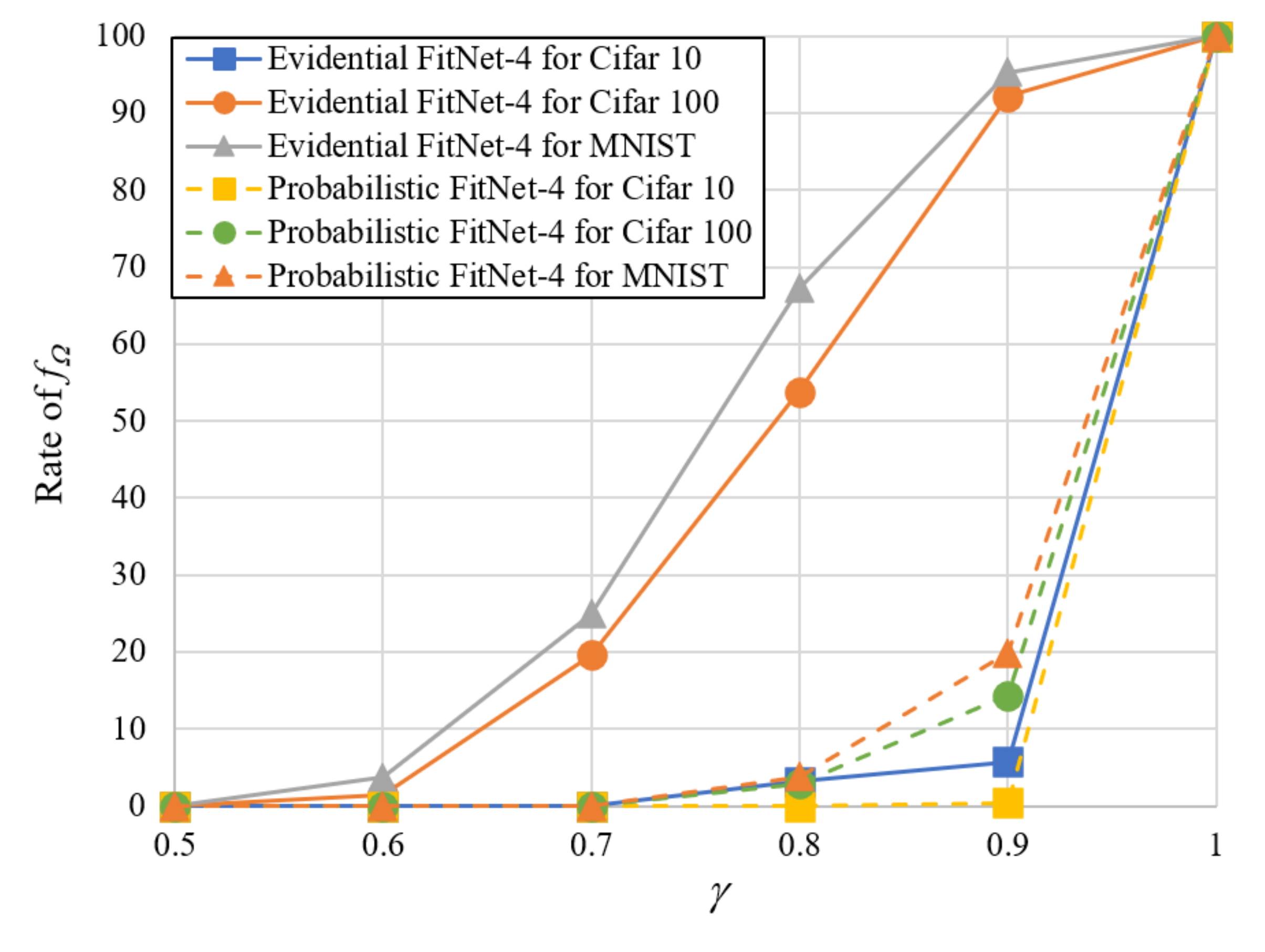}}\\
\subfloat[]{\includegraphics[width=0.5\textwidth]{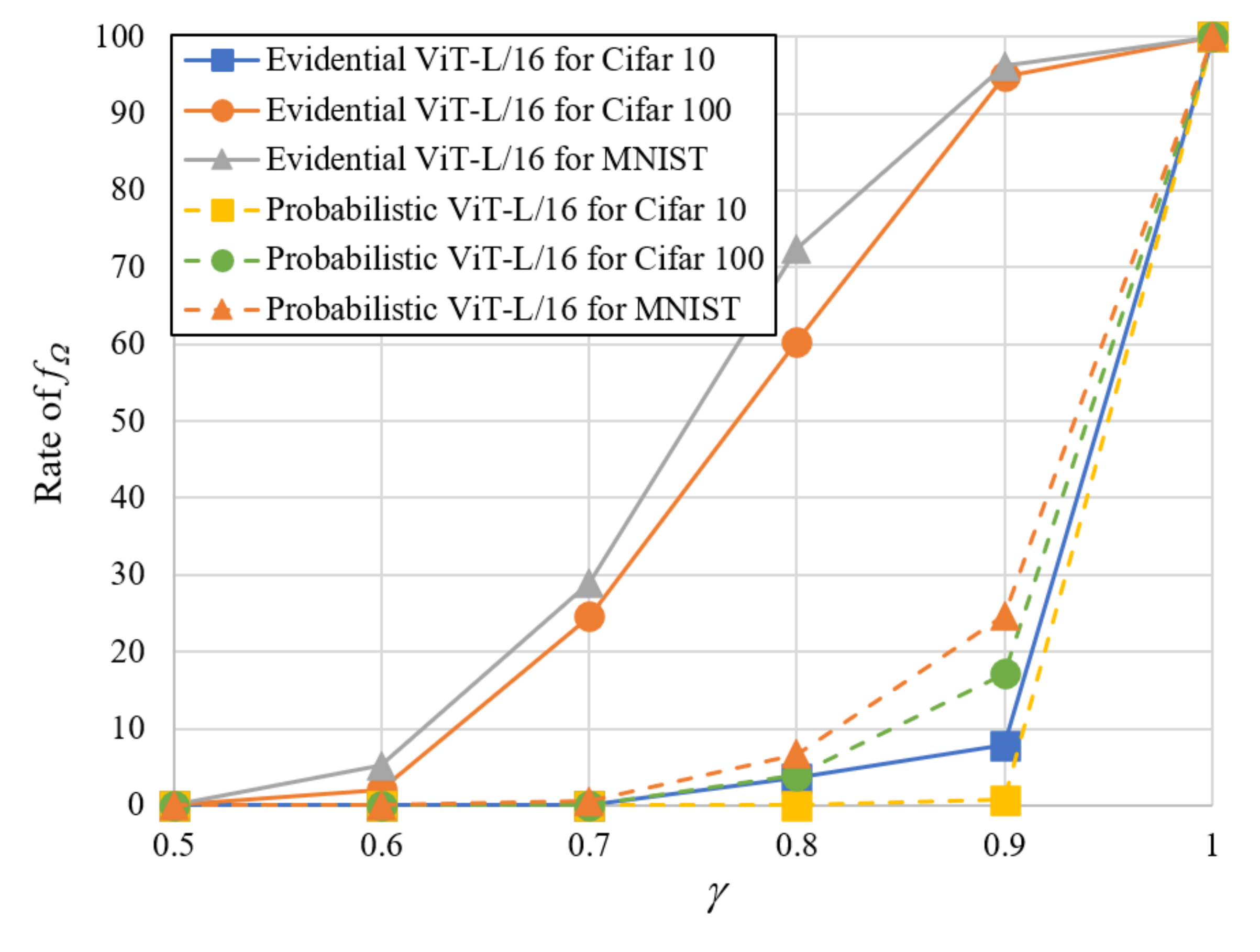}}
\caption{Rate of $f_\Omega$ vs. $\gamma$ for novelty detection in the image-classification experiment:  NIN (a), FitNet-4 (b), and ViT-L/16 (c).}\label{fig:novelty_detection_cifar}
\end{figure}

 \paragraph{Novelty detection}   Figure \ref{fig:novelty_detection_cifar} displays the results of novelty detection using evidential deep-learning and probabilistic classifiers. The evidential deep-learning classifiers can assign outliers and a few of the known-class examples to set $\Omega$ when values of $\gamma$ are between 0.7 and 0.9, while the probabilistic CNN classifiers cannot, which demonstrates that the proposed models outperform the probabilistic CNN classifiers for rejecting outliers together with ambiguous samples. This is due to the fact that, when the feature vector fed into the DS layer is far from all prototypes, the activations of the RBF units in the DS layer become close to zero, as shown by Eq. \eqref{con:si}. As a consequence, all the mass functions $m_i$ computed by Eq. \eqref{con:m^i} assign a large mass to set $\Omega$, and so does their orthogonal sum $m$. The output of the DS layer thus reflects ignorance about the class of the input sample (whereas the probabilistic output of the softmax layer does not), leading to the assignment of the sample to set $\Omega$. 
 
We also applied McNemar's test with  the CIFAR-100 and MNIST datasets, where outliers assigned to $\Omega$ are regarded as positive samples, and the others are negative ones. The results indicate the use of the DS and expected utility layers has a distinct effect on novelty detection since all \emph{p}-values are smaller than 0.001. However, none of classifiers performs well when $\gamma$ is less than 0.7 since these classifiers favor precise decisions. The classifiers tend to reject outliers whose features are different from the known classes. For example, the proposed classifiers reject more samples in the MNIST dataset than in the CIFAR-100 dataset since the hand-written digits are very different from the patterns in the CIFAR-10 dataset.

    \subsection{Signal classification experiment}
    \label{sec:signal_classification_experiment}

    In the application of the proposed classifier on signal processing, we used the UrbanSound 8K dataset \cite{10.1145/2647868.2655045} composed of 8732 short (less than 4 seconds) excerpts of various urban sound sources (air conditioner (\emph{AI}), car horn (\emph{CA}), playing children (\emph{CH}), dog bark (\emph{DO}), drilling (\emph{DR}), engine idling (\emph{EN}), gun shot (\emph{GU}), jackhammer (\emph{JA}), siren (\emph{SI}), street music (\emph{ST})) prearranged into 10 classes. The ratio between the training and testing set is about 3:1. We randomly selected 25\% of the training samples as validation data. Free Spoken Digit Dataset (FSDD) \cite{8969121}, was used to evaluate the capacity of novelty detection in the signal classification experiment. FSDD is an audio/speech dataset with 2,000 recordings (50 of each digit per speaker) in English pronunciations.

    The baseline stages in this experiment are shown in Table \ref{tab:stages_urban}. The DS and expected utility layers show a significant difference in the precise classification as $0.01 \textless p \textless 0.05$ according to McNemar's test (Table \ref{tab:average_utilities_urban}). Similarly to CIFAR-10, this demonstrates that the performance of the proposed classifiers is better than those of probabilistic CNN classifiers for precise classification.

\begin{table}[]
\centering
\caption{The three baseline stages used for UrbanSound 8K.}\label{tab:stages_urban}
\begin{tabular}{clclc}
\hline
\multicolumn{2}{c|}{Stage 1 \cite{piczak2015environmental}}         & \multicolumn{2}{c|}{Stage 2}                                  & Stage 3                                  \\ \hline
\multicolumn{5}{c}{Pre-processing: clip, data augmentation, and segmentation}                                                                                                             \\ \hline
\multicolumn{5}{c}{Input: 60 $\times$ 41 $\times$ 2}                                                                                                            \\ \hline
\multicolumn{2}{c|}{\multirow{2}{*}{57 $\times$ 6 Conv. 80 $ReLU$}} & \multicolumn{2}{c|}{57 $\times$ 6 Conv. 80 $ReLU$} & 29 $\times$ 3 Conv. 80 $ReLU$\\
\multicolumn{2}{c|}{}                                                          & \multicolumn{2}{c|}{1 $\times$ 1 Conv. 80 $ReLU$}  & 29 $\times$ 3 Conv. 80 $ReLU$ \\ \hline
\multicolumn{5}{c}{4 $\times$ 3 max-pooling stride 1 $\times$ 3 with 50\% dropout}                                                                              \\ \hline
\multicolumn{2}{c|}{\multirow{2}{*}{1 $\times$ 3 Conv. 80 $ReLU$}}  & \multicolumn{2}{c|}{1 $\times$ 3 Conv. 80 $ReLU$}  & 1 $\times$ 2 Conv. 80 $ReLU$  \\
\multicolumn{2}{c|}{}                                                          & \multicolumn{2}{c|}{1 $\times$ 1 Conv. 80 $ReLU$}  & 1 $\times$ 2 Conv. 80 $ReLU$  \\ \hline
\multicolumn{5}{c}{1 $\times$ 3 max-pooling stride 1$\times$ 3 without dropout}                                                                                                 \\ \hline
\end{tabular}
\end{table}

\begin{table}[]
 \centering
 \caption{Test average utilities in precise classification on UrbanSound 8K.}\label{tab:average_utilities_urban}
 \resizebox{\textwidth}{!}{
  \begin{tabular}{ccccccc}
   \hline
   \multirow{2}{*}{Models}                                                                    & \multicolumn{2}{c}{Stage 1 \cite{piczak2015environmental}} & \multicolumn{2}{c}{Stage 2}            & \multicolumn{2}{c}{Stage 3}            \\ \cline{2-7} 
   & Probabilistic classifier        & Evidential classifier       & Probabilistic classifier & Evidential classifier & Probabilistic classifier & Evidential classifier \\ \hline
   Utility                                                                                    & 0.7132                & 0.7261                      & 0.7164         & 0.7284                & 0.7210         & 0.7302                \\
   \begin{tabular}[c]{@{}c@{}}\emph{p}-value\\ (McNemar's test)\end{tabular} & \multicolumn{2}{c}{0.0234}                          & \multicolumn{2}{c}{0.0319}             & \multicolumn{2}{c}{0.0365}             \\ \hline
  \end{tabular}}
\end{table}

\begin{figure}
\centering
\subfloat[\label{fig:urban_nu_utility_model1}]{\includegraphics[width=0.5\textwidth]{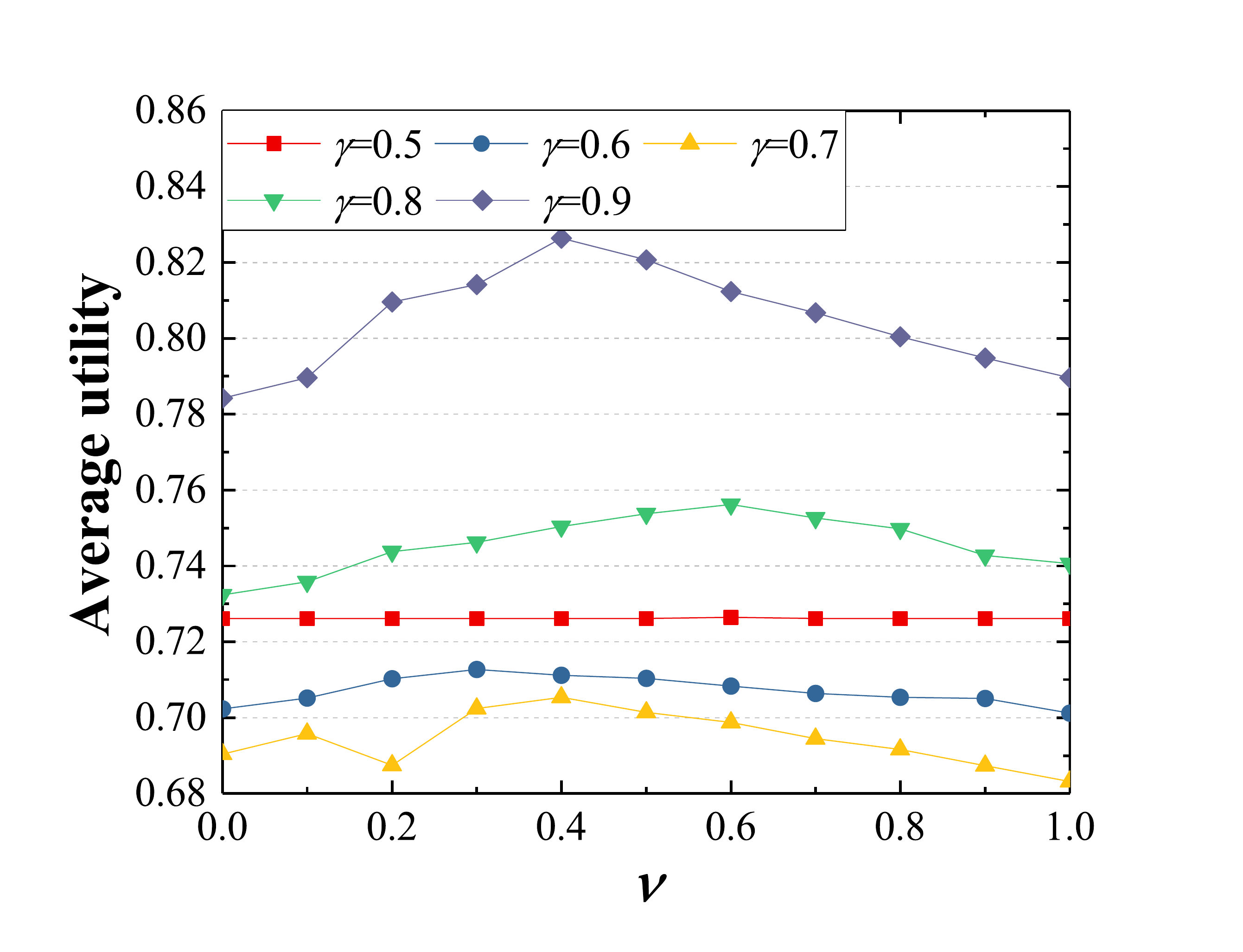}}
\subfloat[\label{fig:urban_nu_utility_model2}]{\includegraphics[width=0.5\textwidth]{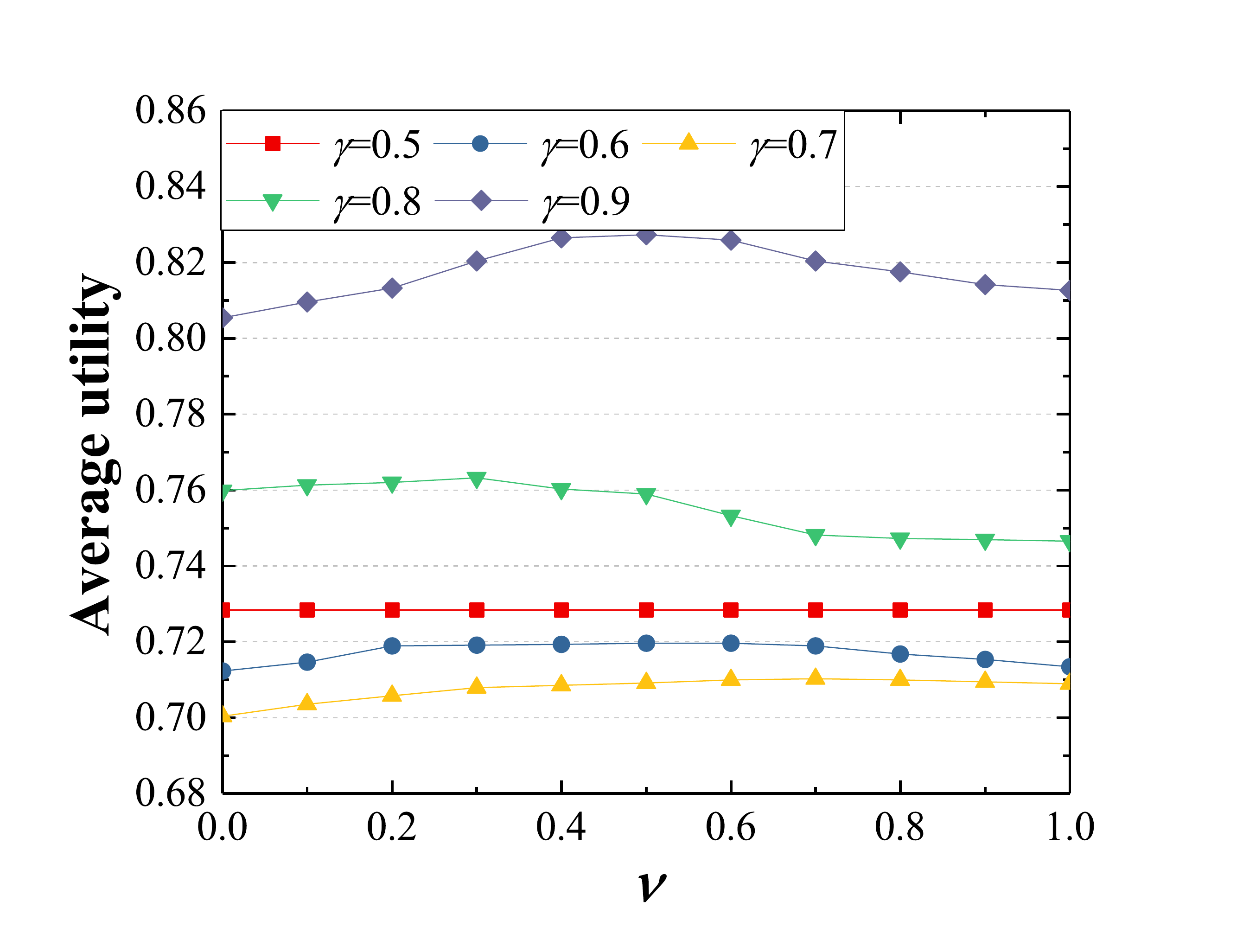}}\\
\subfloat[\label{fig:urban_nu_utility_model3}]{\includegraphics[width=0.5\textwidth]{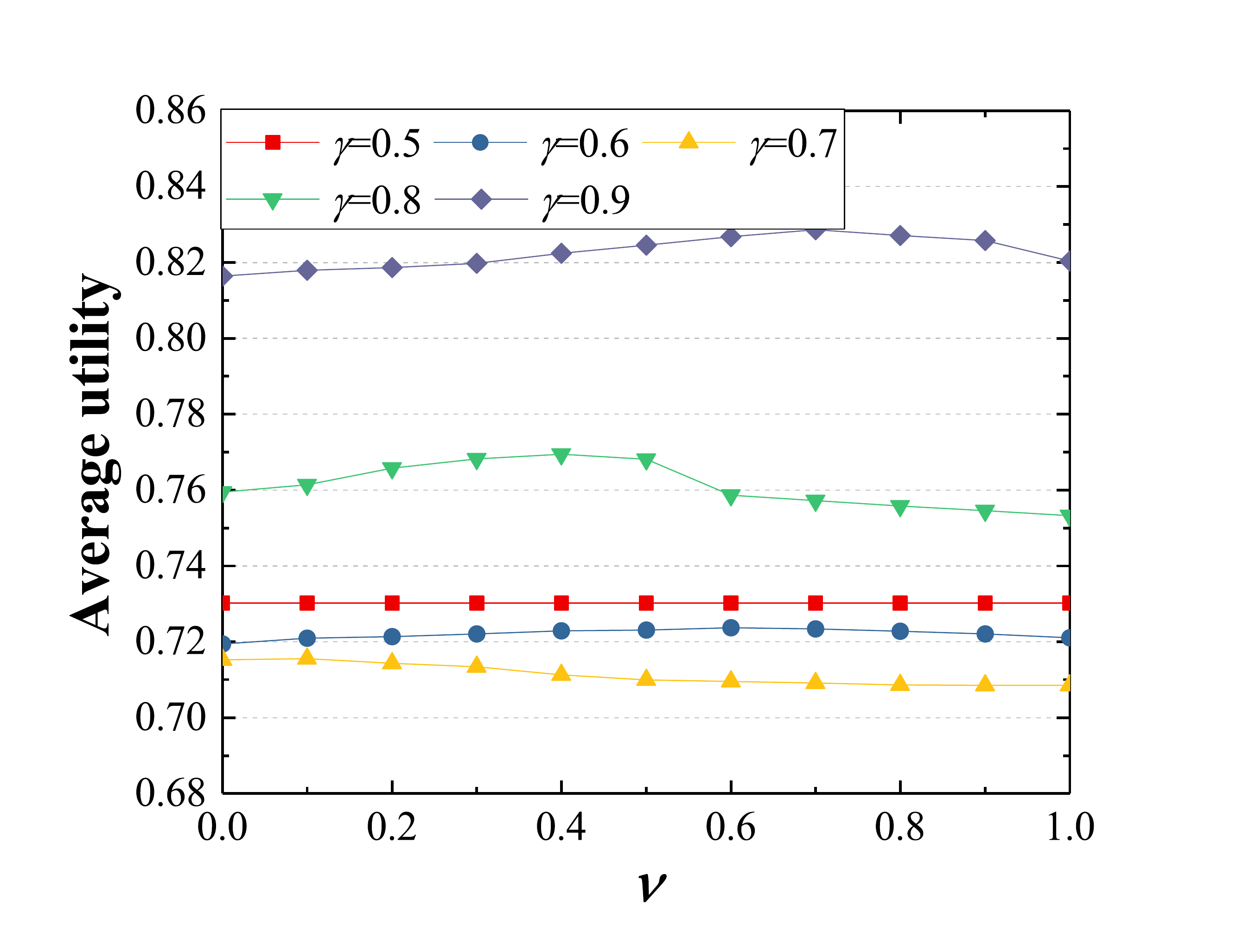}}
\caption{Average utility vs. in $\nu$ for the proposed classifiers on the UrbanSound 8K dataset: Stage 1 (a), Stage 2 (b), and Stage 3 (c).}\label{fig:urban_nu_utility}
\end{figure}

\begin{figure}
\centering
\subfloat[\label{fig:urban_gamma_utility_model1}]{\includegraphics[width=0.50\textwidth]{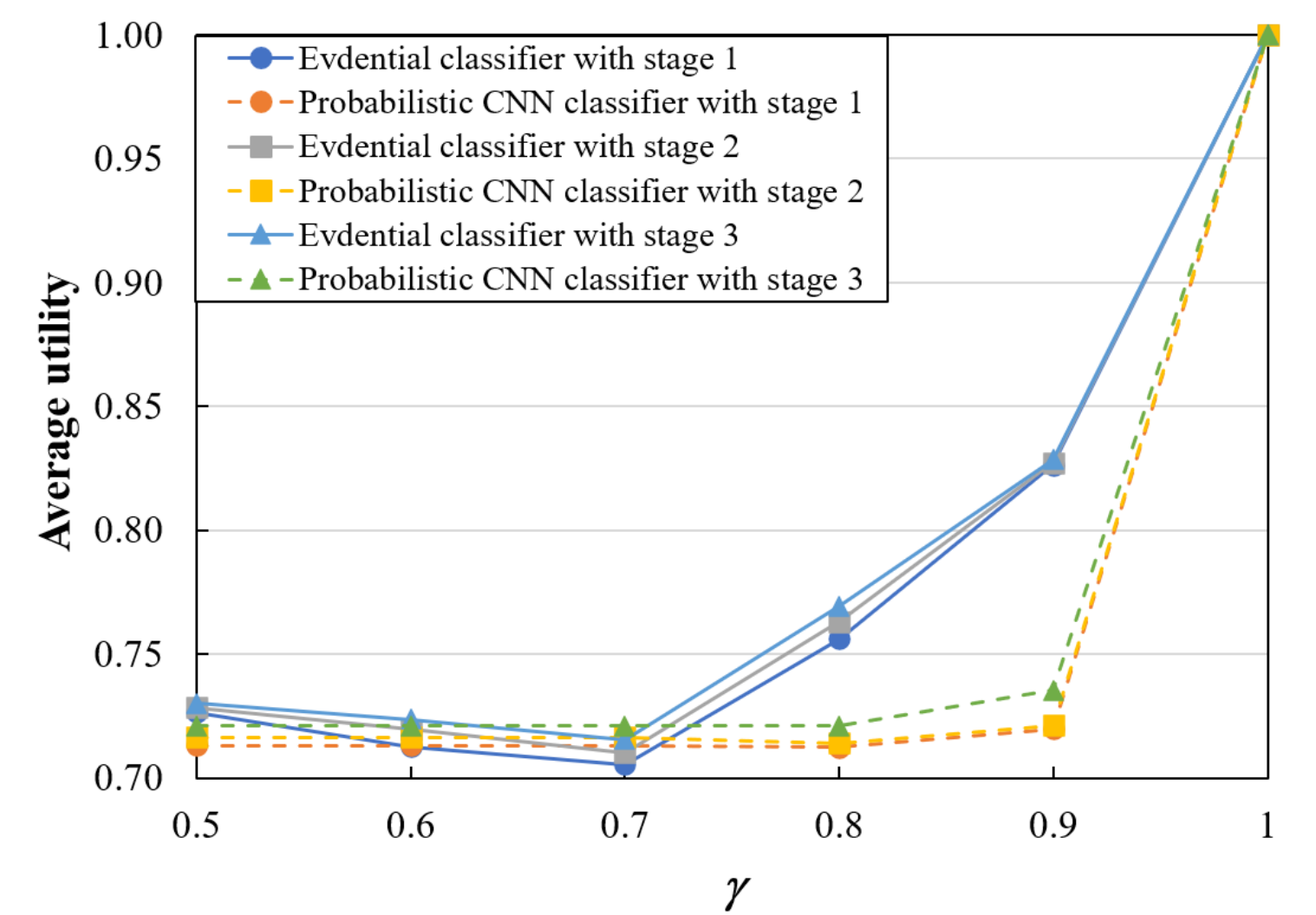}}
\subfloat[\label{fig:urban_gamma_utility_model2}]{\includegraphics[width=0.50\textwidth]{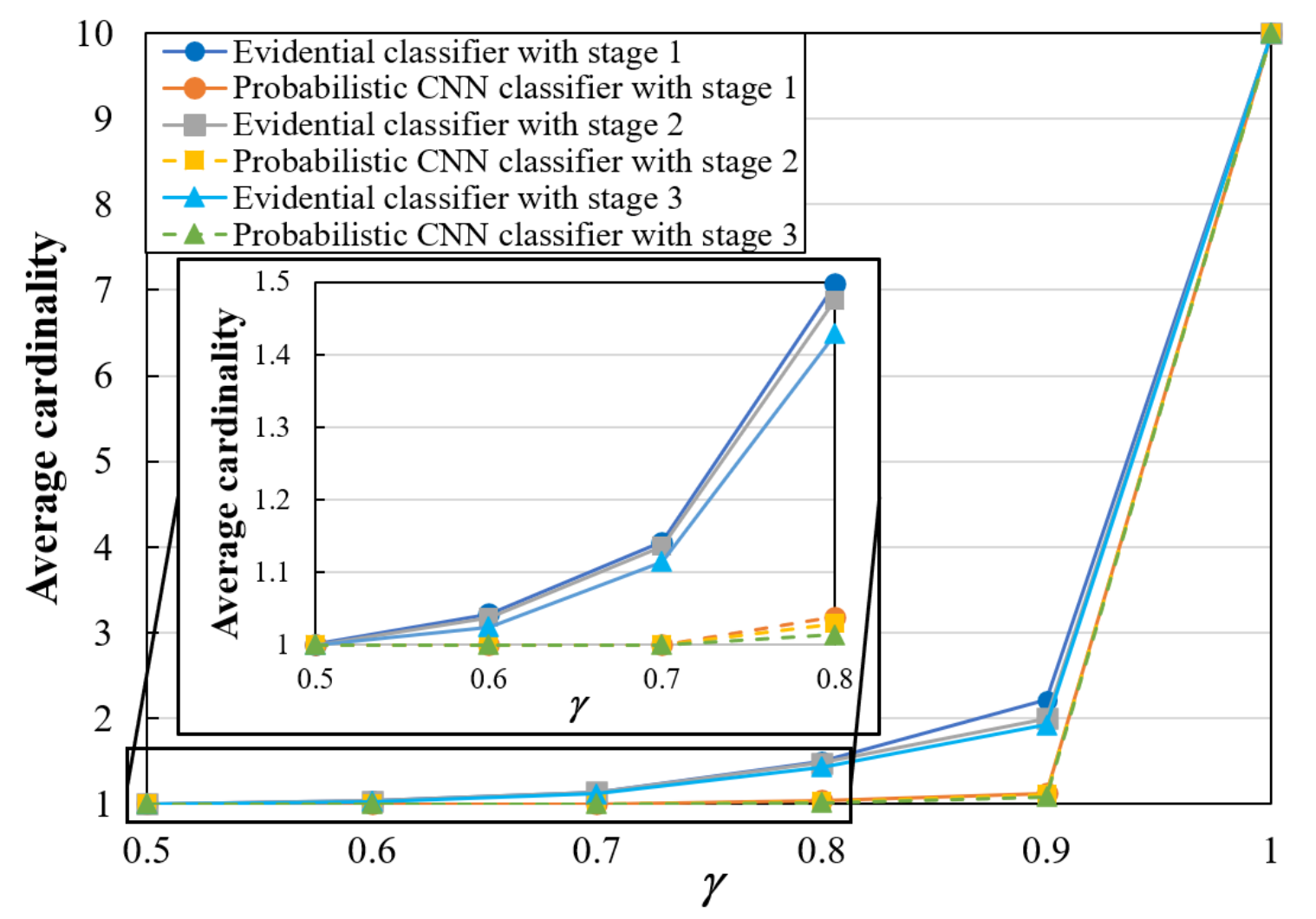}}\\
\caption{Average utility (a) and average cardinality (b) vs.  $\gamma$  for the proposed classifiers and the probabilistic CNN classifiers on the UrbanSound 8K dataset.}\label{fig:urban_gamma_utility}
\end{figure}

\begin{figure}
\centering
\subfloat[]{\includegraphics[width=0.5\textwidth]{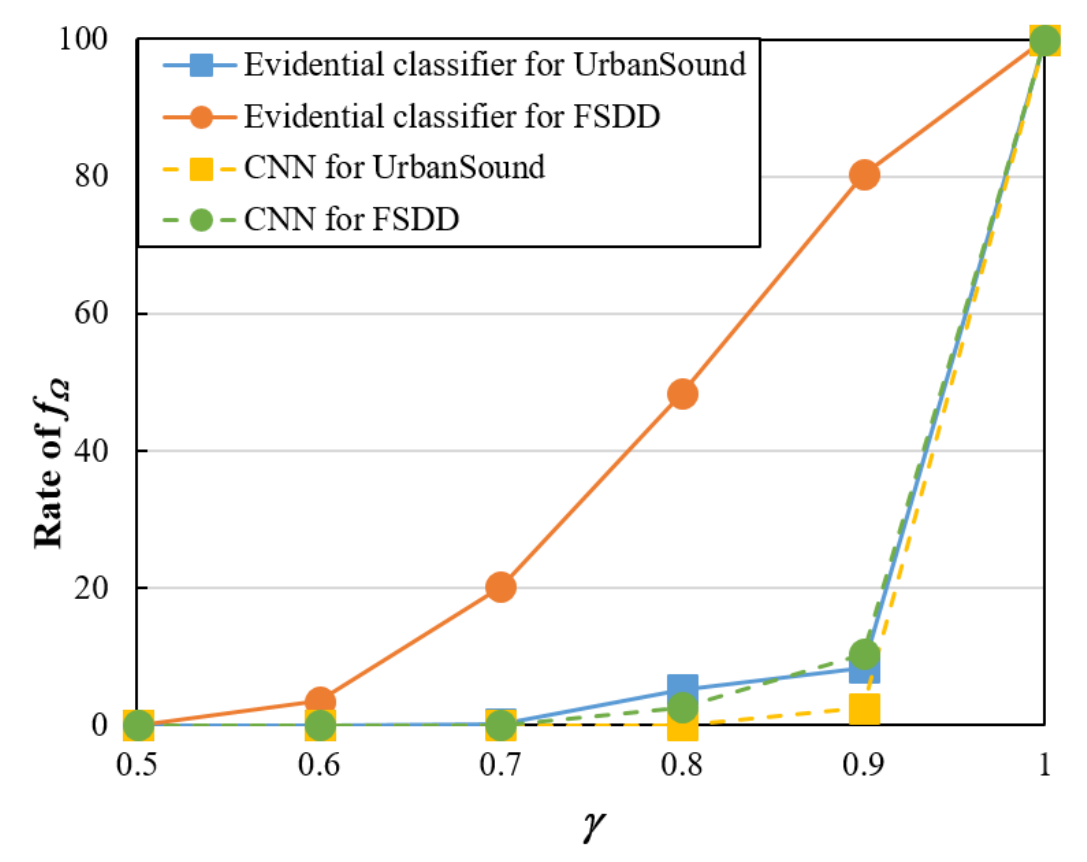}}
\subfloat[]{\includegraphics[width=0.5\textwidth]{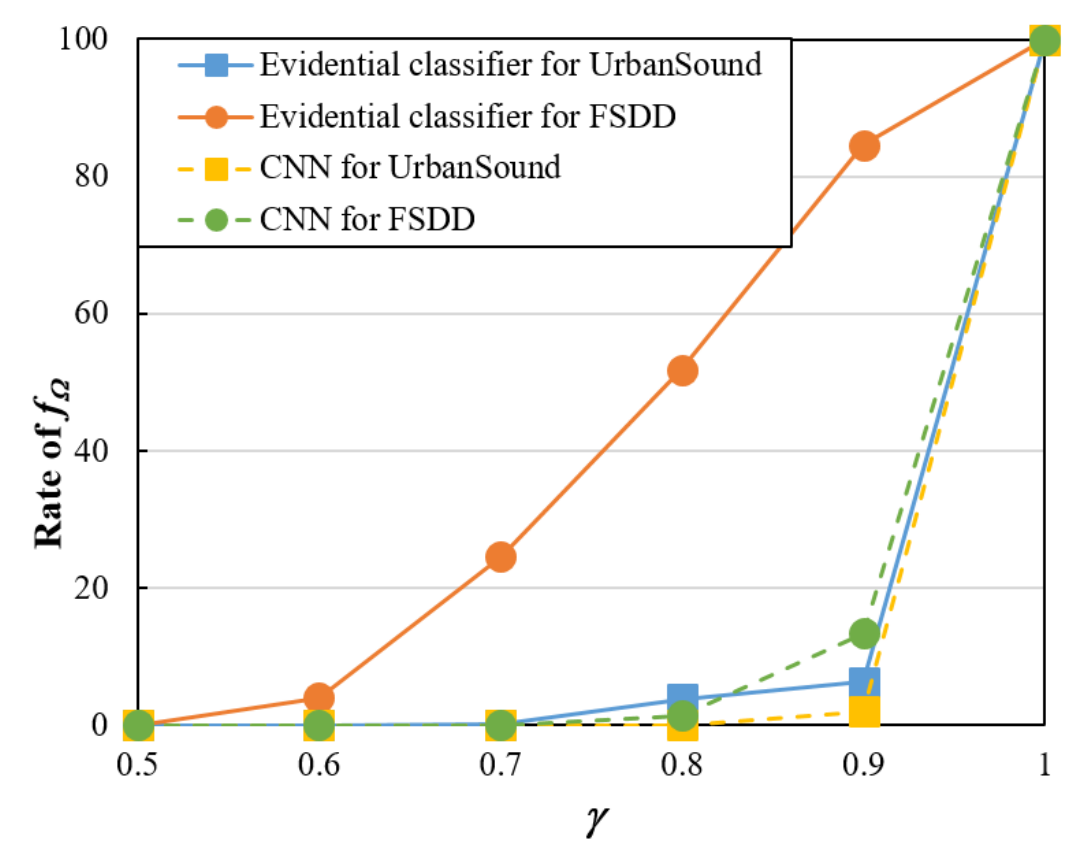}}\\
\subfloat[]{\includegraphics[width=0.5\textwidth]{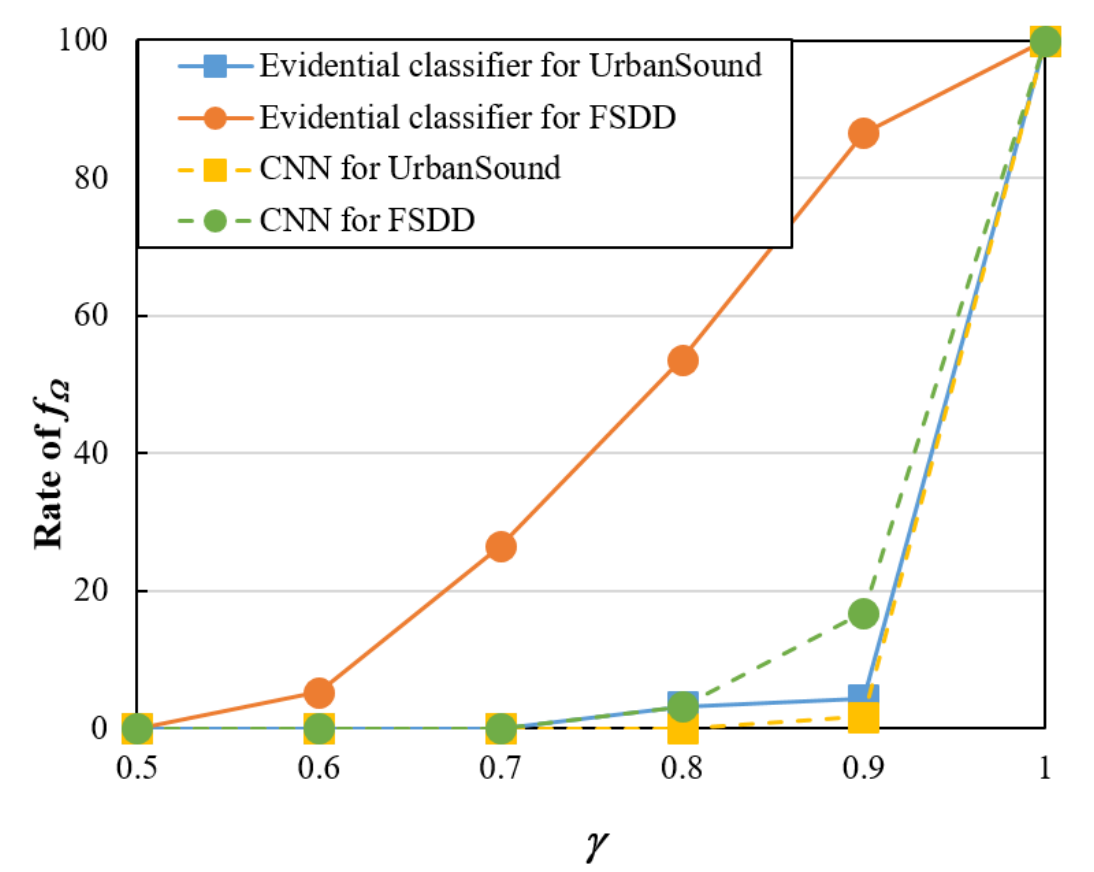}}
\caption{Rate of $f_\Omega$ vs. $\gamma$ for novelty detection in the signal-classification experiment:  Stage 1 (a), Stage 2 (b), and Stage 3 (c).}\label{fig:novelty_detection_urban}
\end{figure}

    After determining the optimal $\nu$ for each value of $\gamma$ based on the $\nu$-utility curves (Figure \ref{fig:urban_nu_utility}), we can compute the test average utilities and cardinalities of the evidential deep-learning and CNN classifiers, as shown in Figure \ref{fig:urban_gamma_utility}. The proposed classifiers outperform the CNN models for the set-valued classification in the signal processing task. The proposed classifiers make more cautious decisions than do the probabilistic CNNs since it assigns ambiguous samples to multi-class sets. Additionally, the performance of the proposed classifiers exceeds those of the CNN classifiers in novelty detection (Figure \ref{fig:novelty_detection_urban}). The use of the DS and expected utility layers has significant effects on novelty detection as the results of \emph{p}-value are close 0 according to McNemar's test.

\begin{figure}
 \centering
 \subfloat[]{\includegraphics[width=0.5\textwidth]{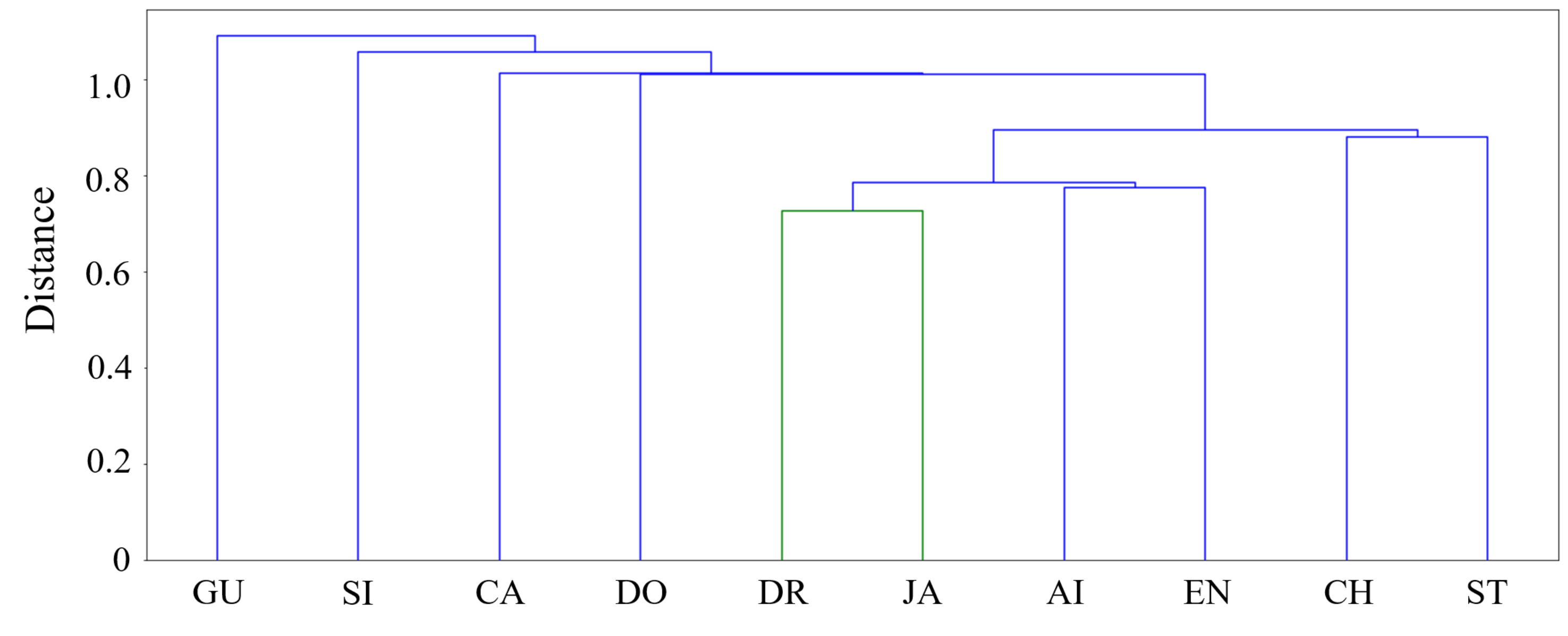}}
 \subfloat[]{\includegraphics[width=0.5\textwidth]{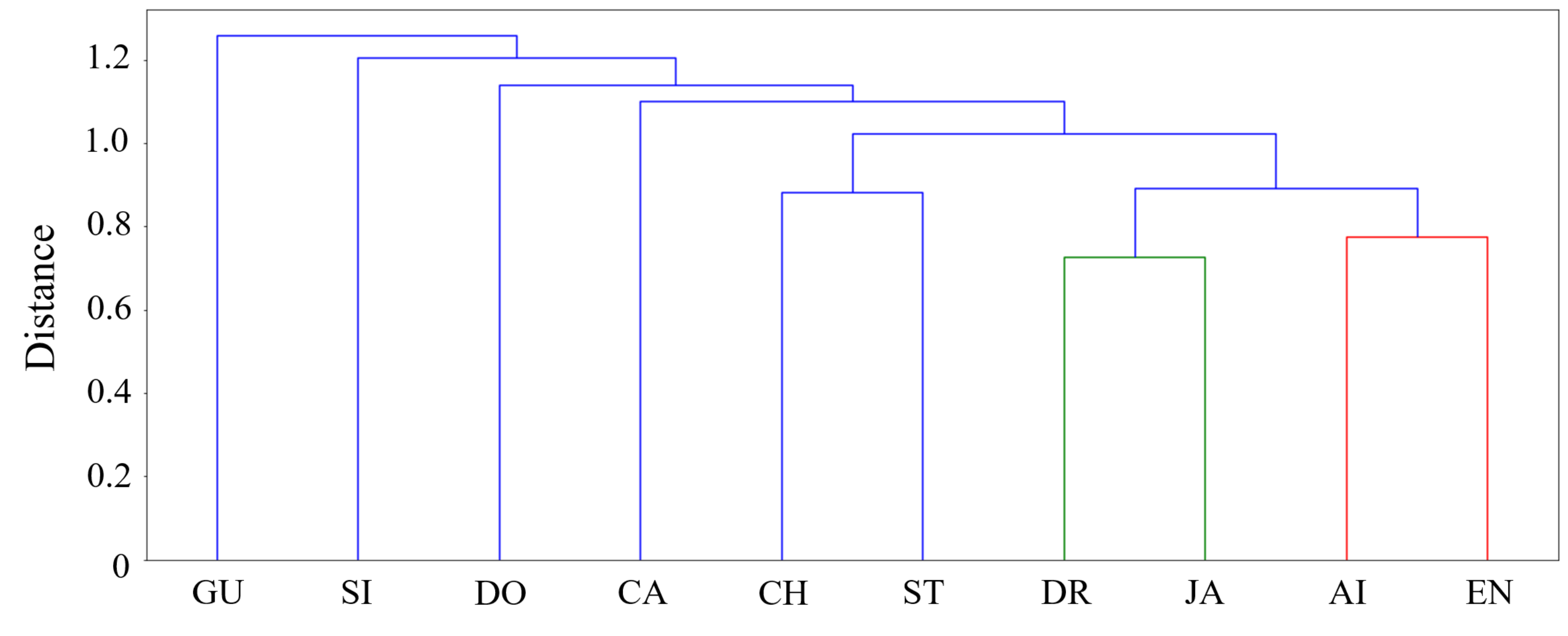}}\\
 \subfloat[]{\includegraphics[width=0.5\textwidth]{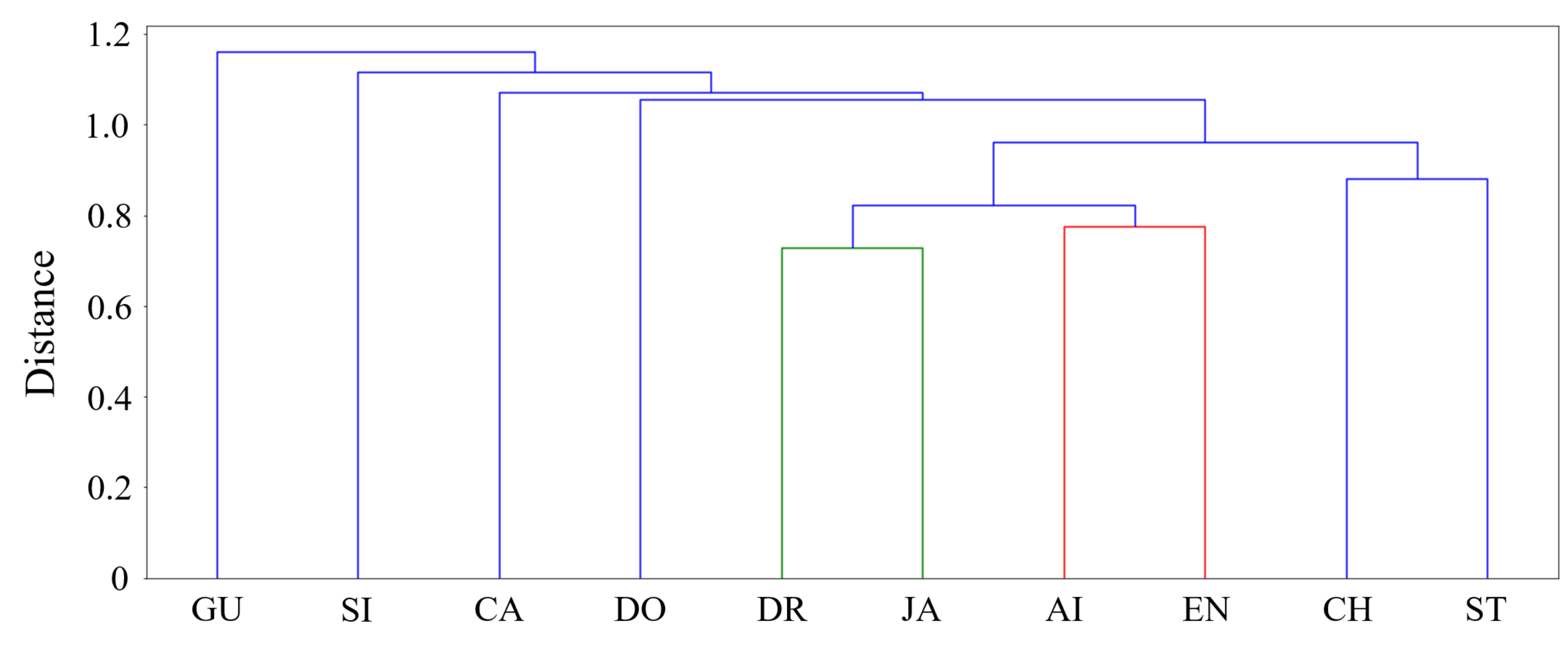}}
 \subfloat[]{\includegraphics[width=0.5\textwidth]{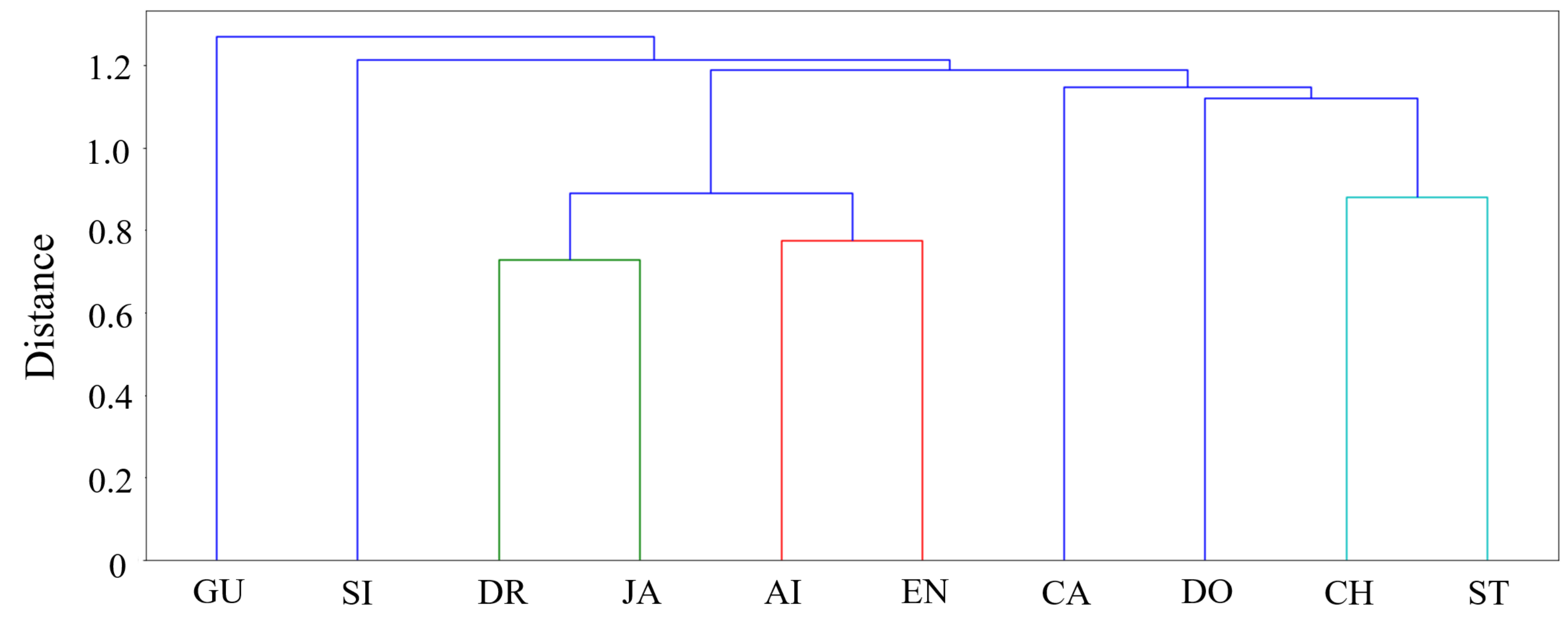}}
 \caption{Dendrograms for the UrbanSound 8K dataset: single linkage (a), complete linkage (b), average linkage (c) , and Ward linkage (d).}\label{fig:dendrograms_urban}
\end{figure}

 For the testing of act-selection strategy, an inflection point was used to cut off the complete-linkage dendrogram \cite{10.1093/comjnl/20.4.364} in Figure \ref{fig:dendrograms_urban}, in which CHI is 2.198 and corresponding distance is 1.036. Thus, we selected partial multi-class sets including $\{DR,JA\}$, $\{AI,EN\}$, $\{CH,ST\}$, $\{DR,JA,AI,EN\}$, and $\{DR,JA,AI,EN,CH,ST\}$. From Tables \ref{tab:rates_set_valued_assignments} and  \ref{tab:rates_of_incorrect}, we can see that the strategy works well if $\gamma$ is less than 0.9. This demonstrates that the proposed strategy is acceptable when the classifier has a reasonable $\gamma$.

    \subsection{Semantic-relationship classification experiment}
    \label{sec:semantic_classification_experiment}

    For the semantic-relationship classification task, we used the SemEval-2010 Task 8 dataset \cite{hendrickx-etal-2010-semeval}. It contains 10,717 annotated examples, including 8,000 training instances and 2,717 test instances. There are 10 semantic relationships in the dataset as cause-effect (\emph{CE}), instrument-agency (\emph{IA}), product-producer (\emph{PP}), content-container (\emph{CC}), entity-origin (\emph{EO}), entity-destination (\emph{ED}), component-whole (\emph{CW}), member-collection (\emph{MC}), message-topic (\emph{MT}), and other (\emph{O}). The approach to generate the validation set in this experiment is the same as those used in the experiments on the CIFAR-10 and UrbanSound 8K datasets. The FewRel dataset \cite{han2018fewrel} with 100 semantic-relationship classes and 70,000 examples was used in novelty detection, in which the known-class examples were excluded in the experiment.

    We referred to the stages shown in Table \ref{tab:stages_semantic} to design the evidential deep-learning classifiers. In the precise classification, the use of DS and expected utility layers improves the test average utilities of the deep-learning models, as shown in Table \ref{tab:average_utilities_semantic}. Thus, a DS layer and an expected utility layer instead of a softmax layer introduce a positive effect on the networks in the semantic-relationship classification.

\begin{figure}
\centering
\subfloat[\label{fig:semantic_gamma_utility_model1}]{\includegraphics[width=0.50\textwidth]{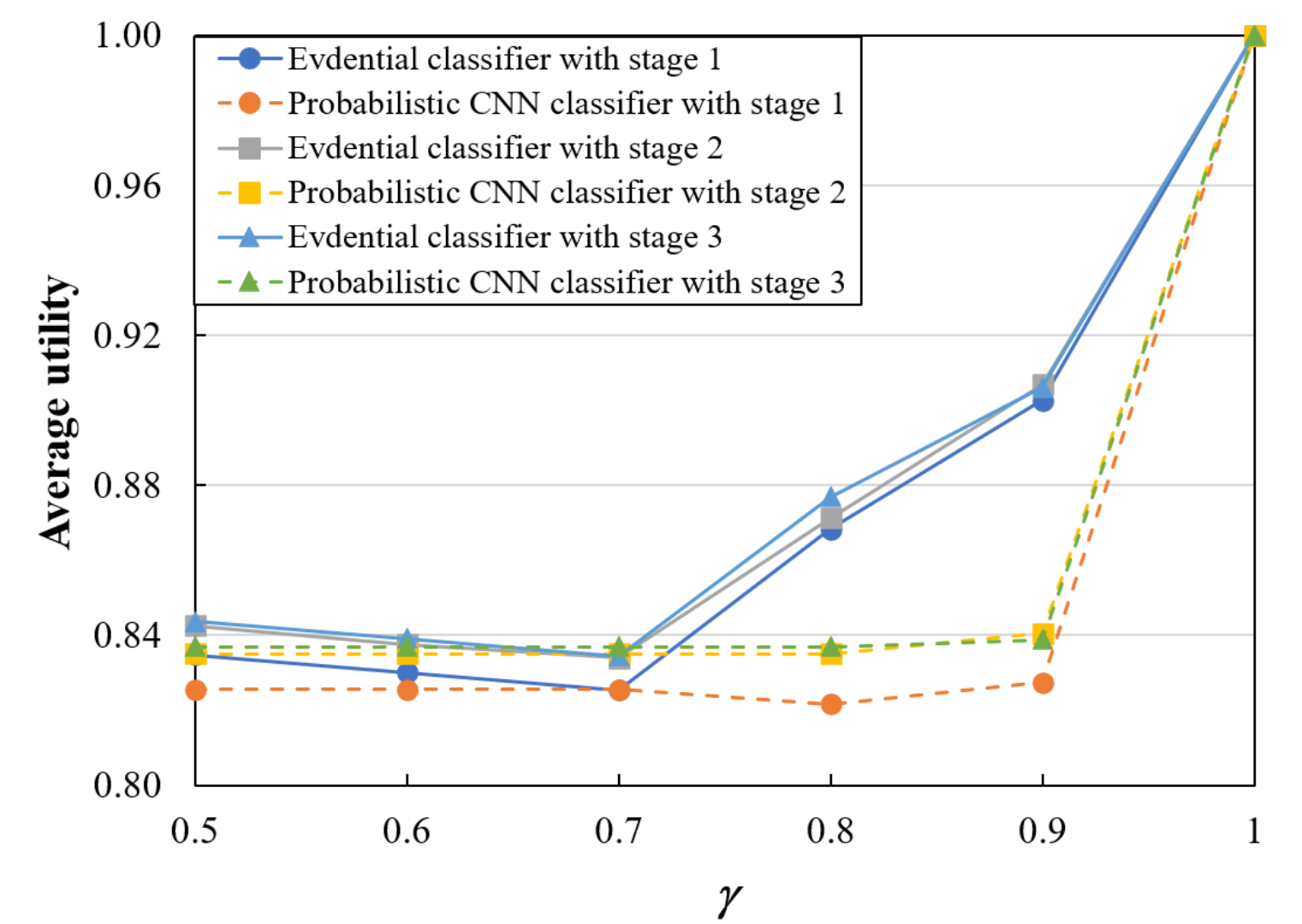}}
\subfloat[\label{fig:semantic_gamma_utility_model2}]{\includegraphics[width=0.50\textwidth]{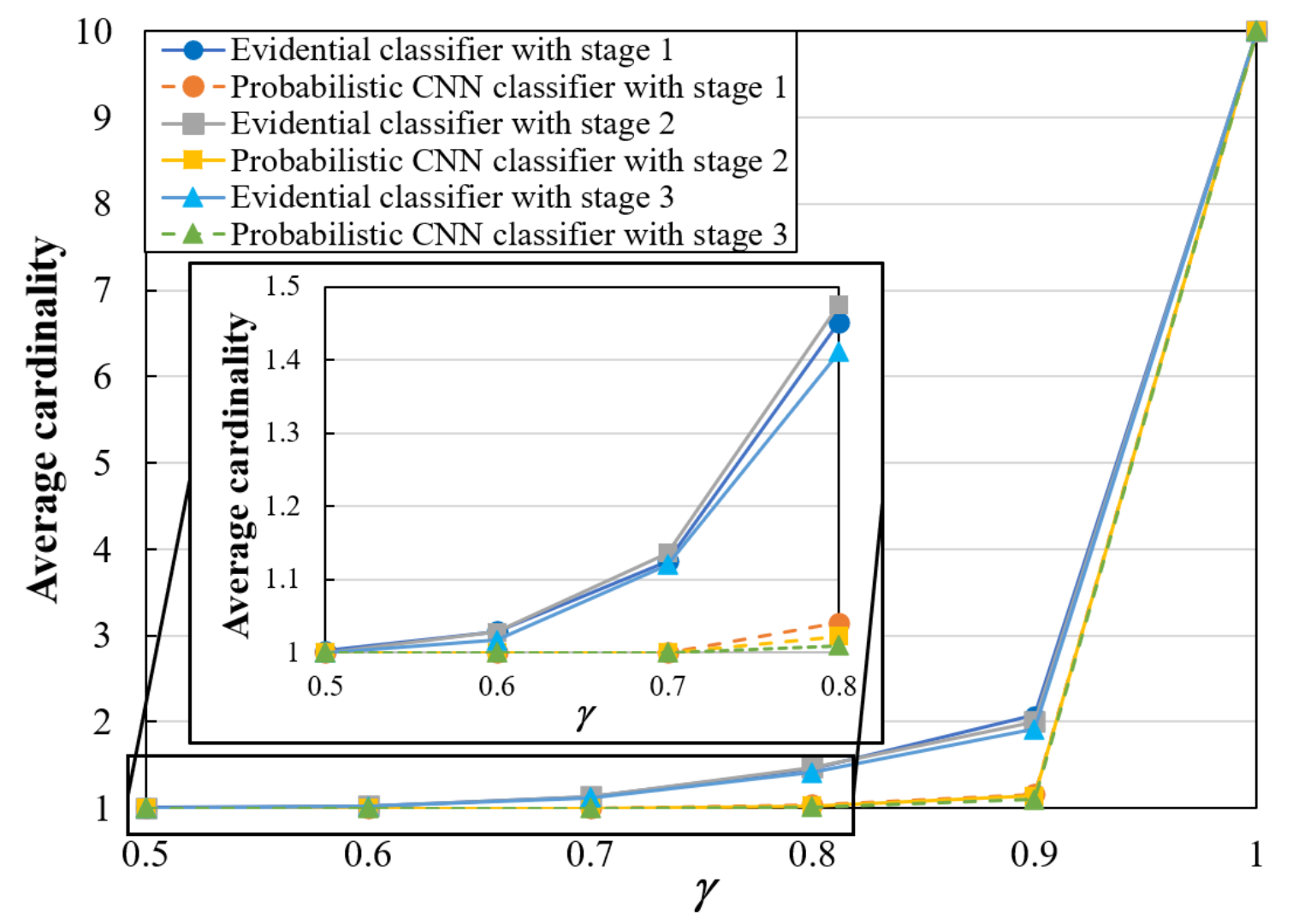}}\\
\caption{Average utility (a) and average cardinality (b) vs.  $\gamma$  for the proposed classifiers and the probabilistic CNN classifiers on the SemEval-2010 Task 8 dataset.}\label{fig:semantic_gamma_utility}
\end{figure}

\begin{figure}
\centering
\subfloat[]{\includegraphics[width=0.5\textwidth]{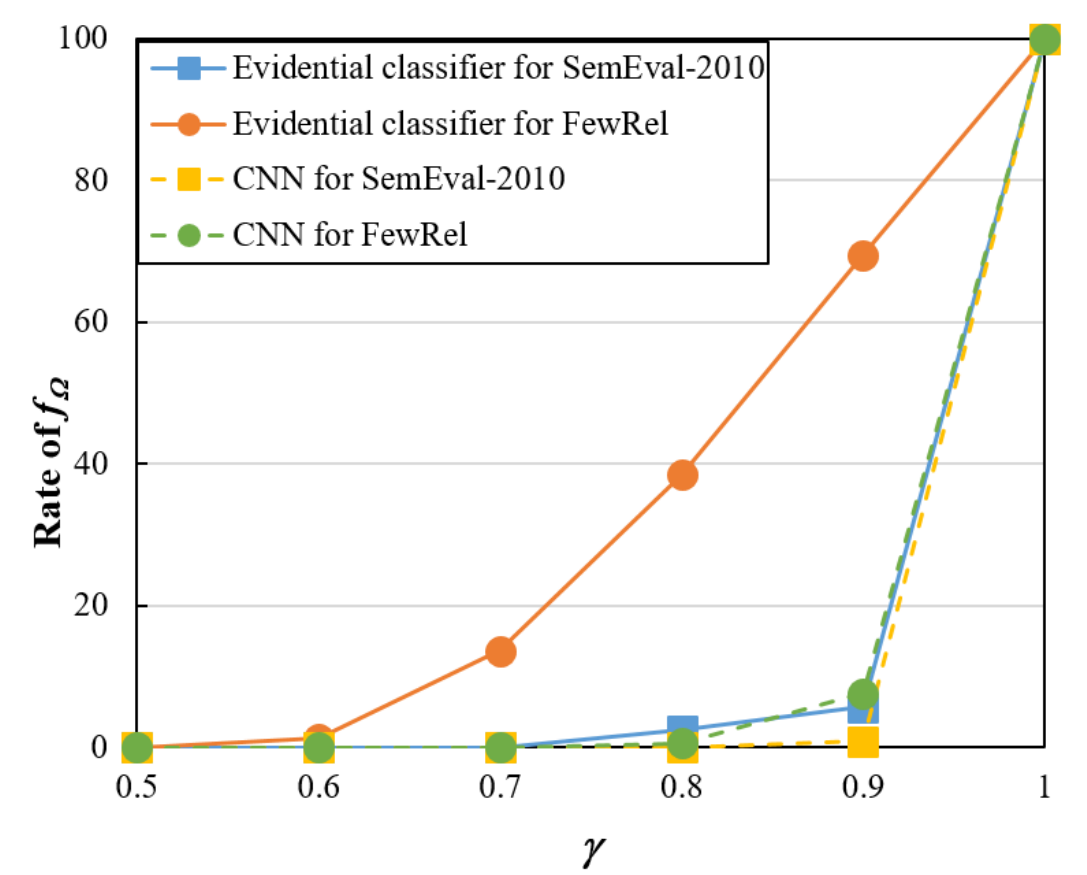}}
\subfloat[]{\includegraphics[width=0.5\textwidth]{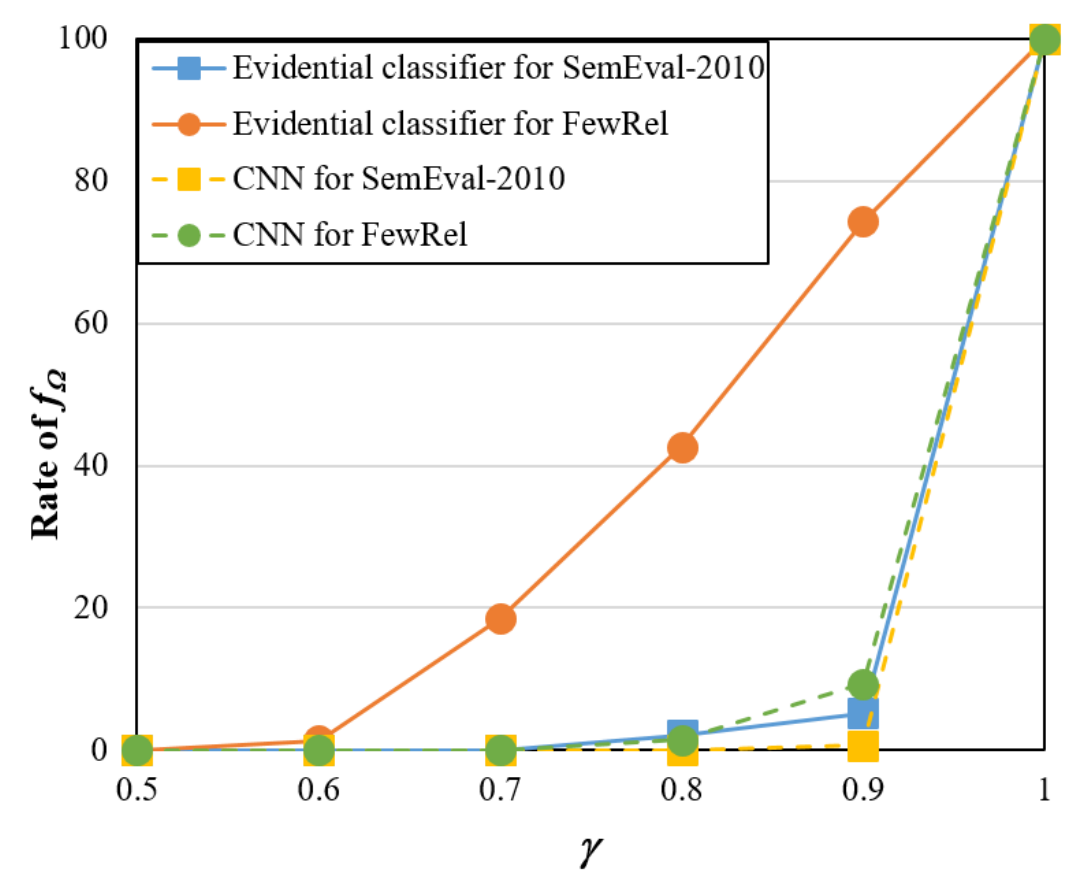}}\\
\subfloat[]{\includegraphics[width=0.5\textwidth]{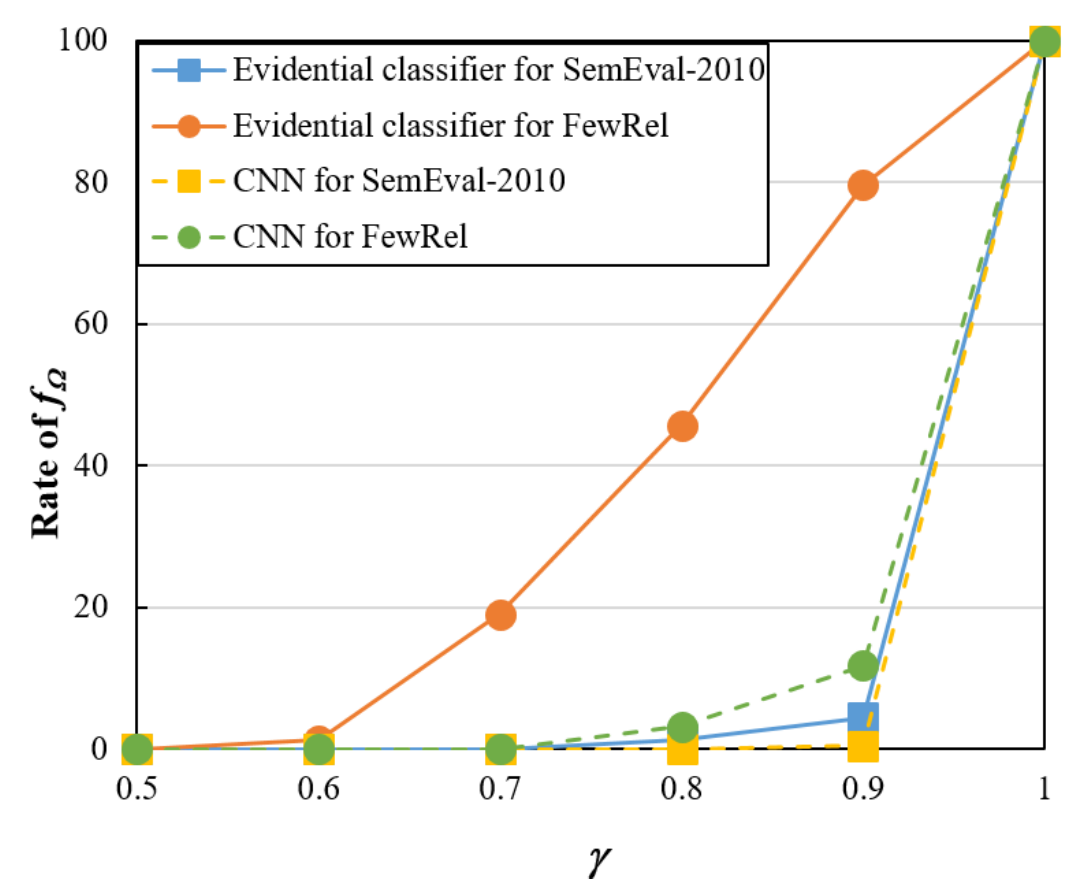}}
\caption{Rate of $f_\Omega$ vs. $\gamma$ for novelty detection in the semantic-relationship-classification experiment:  Stage 1 (a), Stage 2 (b), and Stage 3 (c).}\label{fig:novelty_detection_semantic}
\end{figure}

\begin{figure}
 \centering
 \subfloat[]{\includegraphics[width=0.5\textwidth]{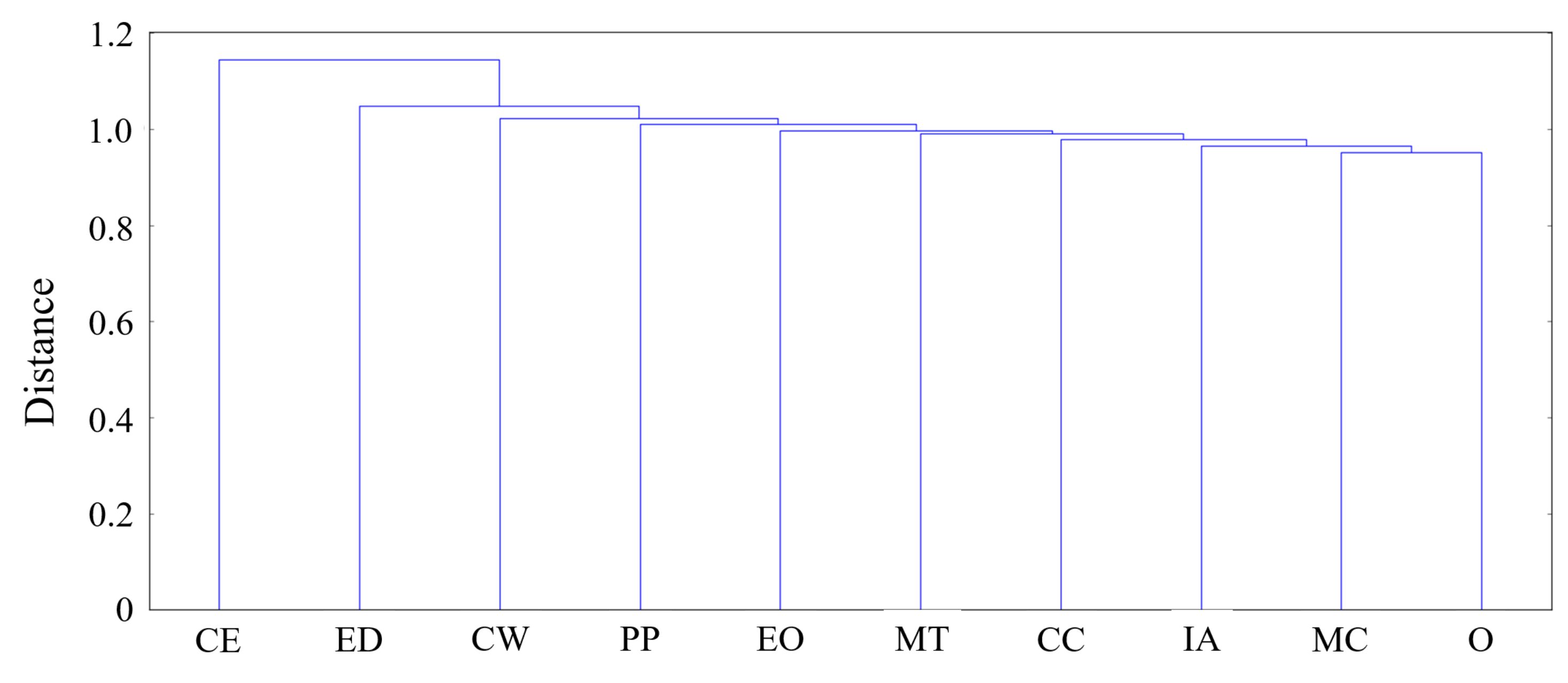}}
 \subfloat[]{\includegraphics[width=0.5\textwidth]{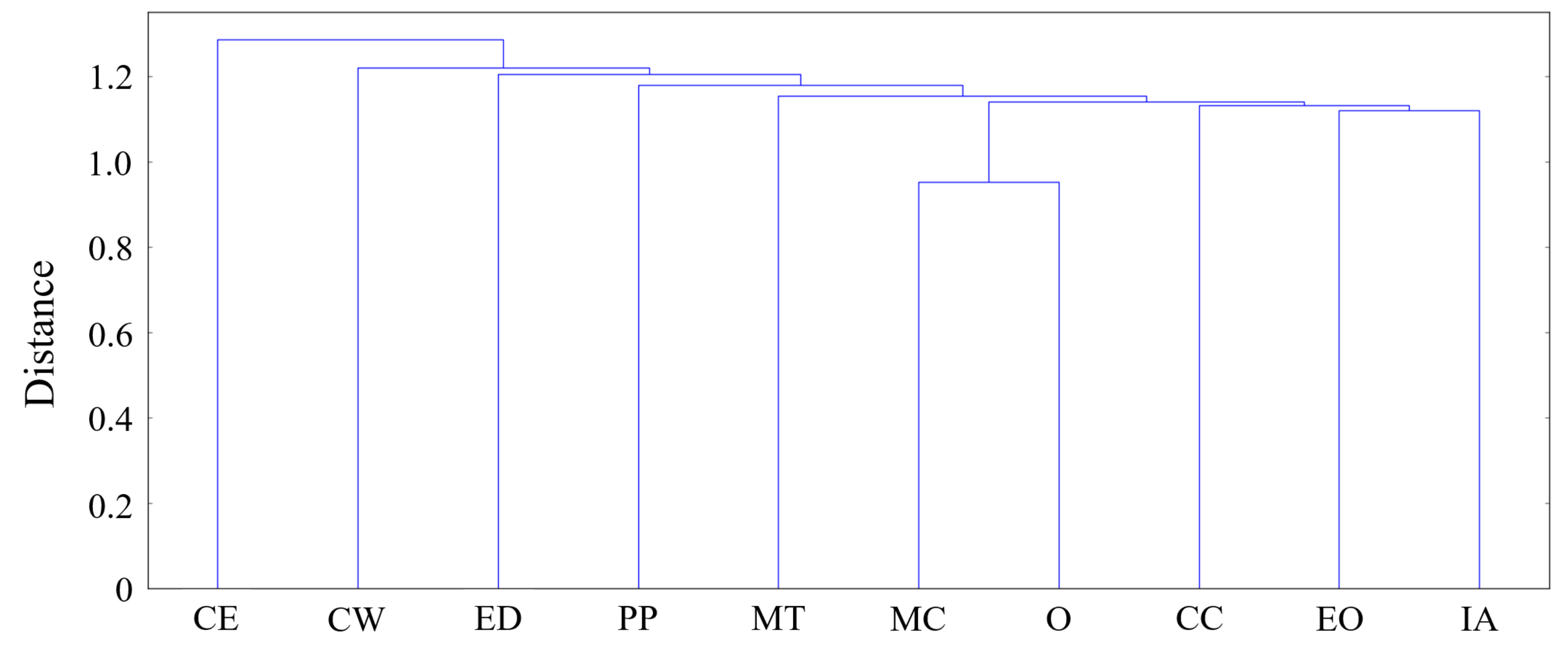}}\\
 \subfloat[]{\includegraphics[width=0.5\textwidth]{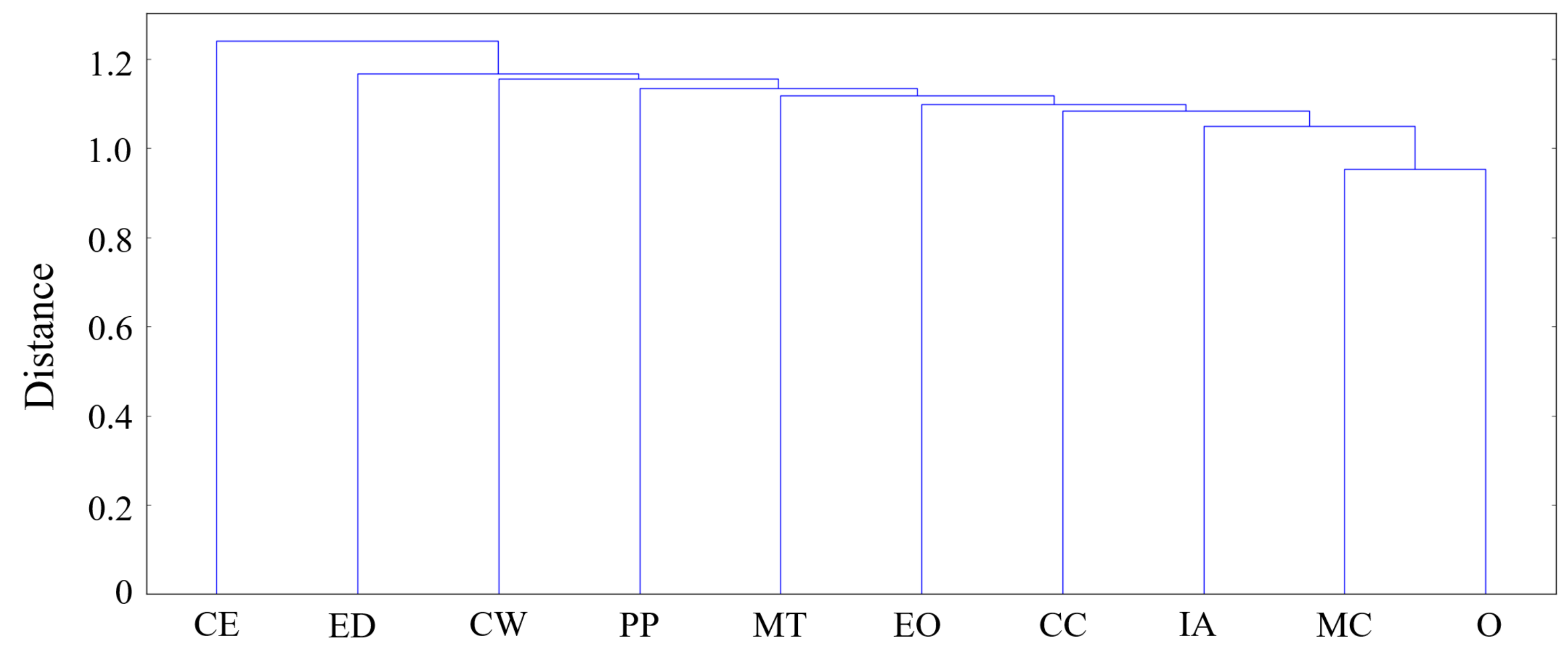}}
 \subfloat[]{\includegraphics[width=0.5\textwidth]{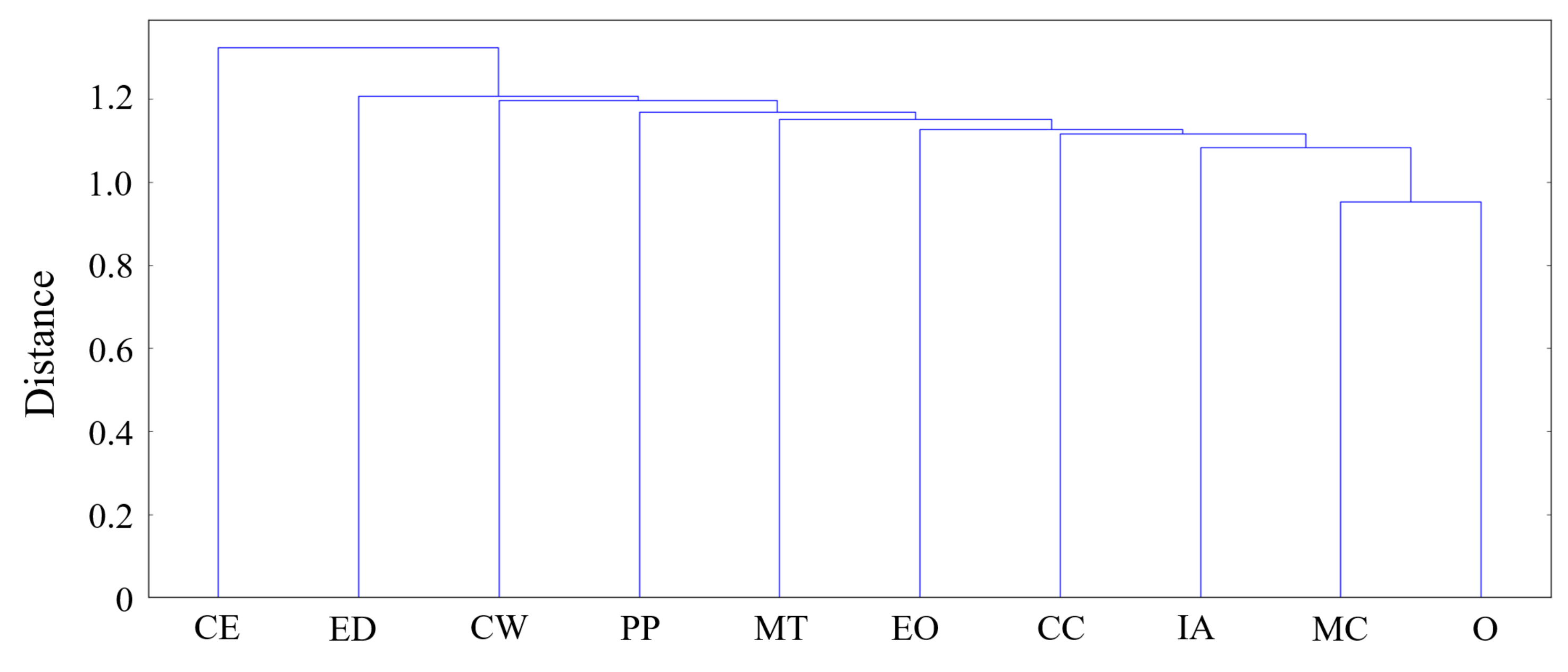}}
 \caption{Dendrograms for the SemEval-2010 Task 8 dataset: single linkage (a), complete linkage (b), average linkage (c) , and Ward linkage (d).}\label{fig:dendrograms_semantic}
\end{figure}

    The strategy for determining the optimal values of $\nu$ in this experiment was the same as those in the CIFAR-10 and UrbanSound 8K experiments. The test average utilities in set-valued classification of the two types of models are shown in Figure \ref{fig:semantic_gamma_utility}, which demonstrate the superiority of the evidential deep-learning classifiers. Figure \ref{fig:novelty_detection_semantic} indicates the acceptable capacity of novelty detection in the evidential deep-learning classifiers. Similar as the CIFAR-10 and UrbanSound 8K dataset, the acts generated from the complete-linkage dendrogram (Figure \ref{fig:dendrograms_semantic} and an inflection point whose CHI is 2.627 and a distance equals 1.107) works as well as the $2^\Omega$ acts if the classifier has a suitable $\gamma$.

\begin{table}[]
\centering
\caption{The three baseline stages used on SemEval-2010 Task 8.}\label{tab:stages_semantic}
\begin{tabular}{clclcl}
\hline
\multicolumn{2}{c|}{Stage 1 \cite{Zeng2014relation}}               & \multicolumn{2}{c|}{Stage 2}                     & \multicolumn{2}{c}{Stage 3}                     \\ \hline
\multicolumn{6}{c}{Pre-processing: word representation}                                                                                                                  \\ \hline
\multicolumn{6}{c}{Input: 50 $\times$ 1 $\times$ $t$, in which $t$ is the number of input sentences}                                                                     \\ \hline
\multicolumn{2}{c|}{\multirow{2}{*}{3 $\times$ 1 Conv. 200 $ReLU$}} & \multicolumn{2}{c|}{3 $\times$ 1 Conv. 200 $ReLU$} & \multicolumn{2}{c}{2 $\times$ 1 Conv. 200 $ReLU$} \\
\multicolumn{2}{c|}{}                                               & \multicolumn{2}{c|}{1 $\times$ 1 Conv. 200 $ReLU$} & \multicolumn{2}{c}{2 $\times$ 1 Conv. 200 $ReLU$} \\ \hline
\multicolumn{2}{c|}{\multirow{2}{*}{1 $\times$   1 Conv. 100 $tanh$}} & \multicolumn{2}{c|}{1 $\times$ 1 Conv. 200 $tanh$} & \multicolumn{2}{c}{1 $\times$ 1 Conv. 200 $tanh$} \\
\multicolumn{2}{c|}{}                                               & \multicolumn{2}{c|}{1 $\times$ 1 Conv. 100 $tanh$} & \multicolumn{2}{c}{1 $\times$ 1 Conv. 100 $tanh$} \\ \hline
\multicolumn{6}{c}{1 $\times$ 1 mean-pooling stride 1$\times$ 1}                                                                                                         \\ \hline
\end{tabular}
\end{table}

\begin{table}[]
 \centering
 \caption{Test average utilities in precise classification on SemEval-2010 Task 8.}\label{tab:average_utilities_semantic}
 \resizebox{\textwidth}{!}{
  \begin{tabular}{ccccccc}
   \hline
   \multirow{2}{*}{Models}                                                                    & \multicolumn{2}{c}{Stage 1 \cite{Zeng2014relation}} & \multicolumn{2}{c}{Stage 2}            & \multicolumn{2}{c}{Stage 3}            \\ \cline{2-7} 
   & Probabilistic classifier    & Evidential classifier    & Probabilistic classifier & Evidential classifier & Probabilistic classifier & Evidential classifier \\ \hline
   Utility                                                                                    & 0.8255            & 0.8347                   & 0.8351         & 0.8425                & 0.837          & 0.8436                \\
   \begin{tabular}[c]{@{}c@{}}\emph{p}-value\\ (McNemar's test)\end{tabular} & \multicolumn{2}{c}{0.0301}                   & \multicolumn{2}{c}{0.0415}             & \multicolumn{2}{c}{0.0430}             \\ \hline
 \end{tabular}}
\end{table}

\begin{figure}
\centering
\subfloat[\label{fig:semantic_nu_utility_model1}]{\includegraphics[width=0.5\textwidth]{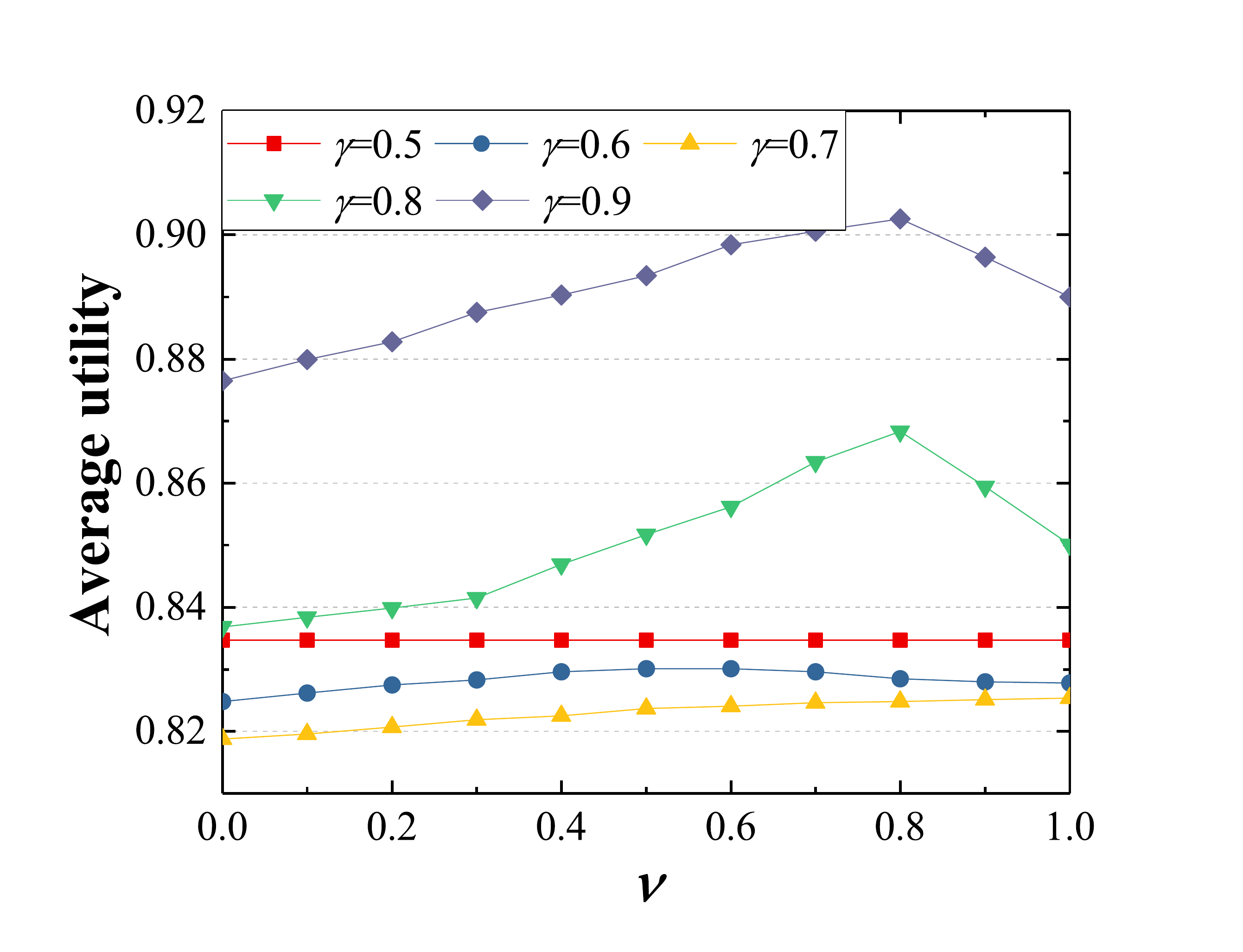}}
\subfloat[\label{fig:semantic_nu_utility_model2}]{\includegraphics[width=0.5\textwidth]{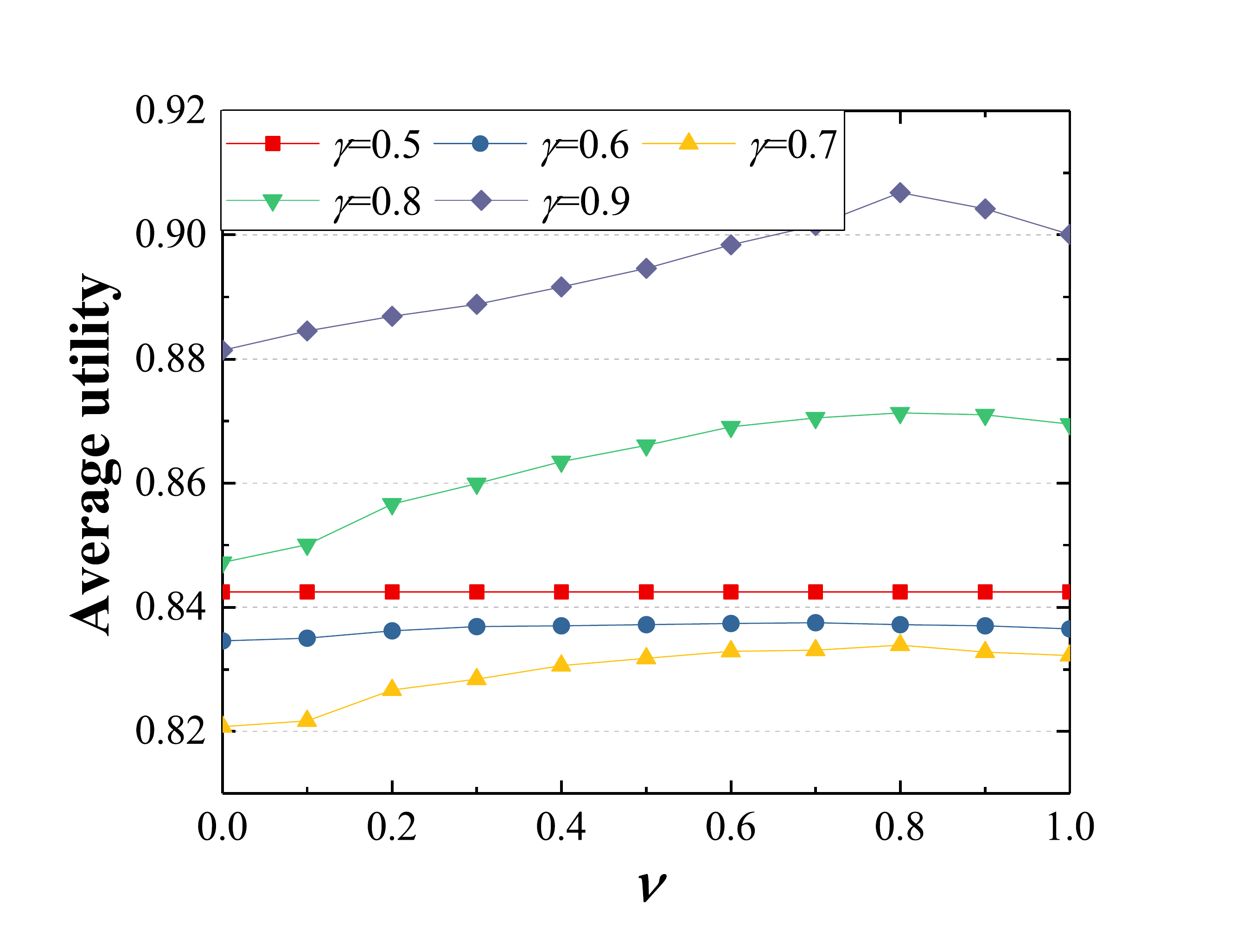}}\\
\subfloat[\label{fig:semantic_nu_utility_model3}]{\includegraphics[width=0.5\textwidth]{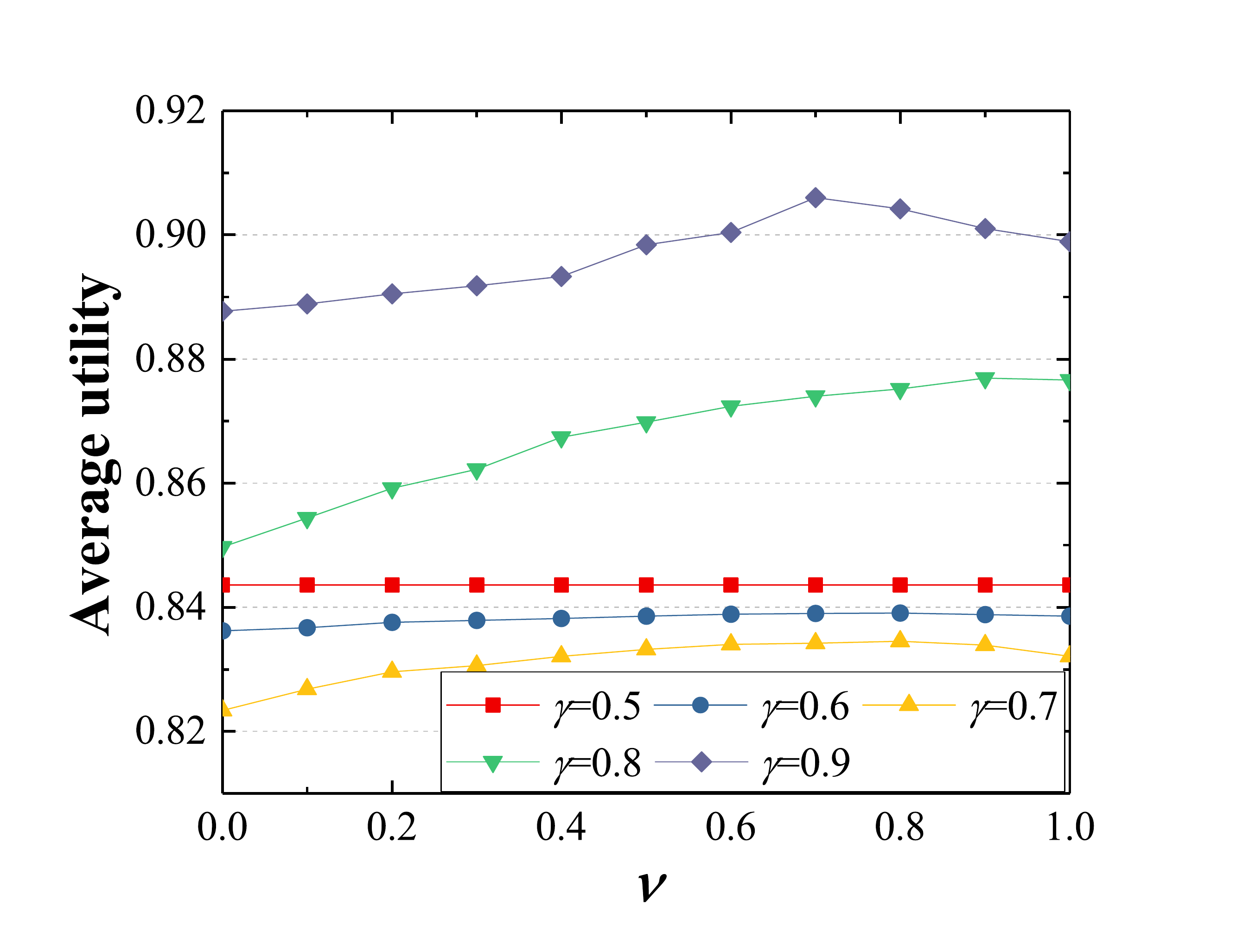}}
\caption{Curves in $\nu$-utility for the proposed classifiers on the SemEval-2010 Task 8 dataset: Stage 1 (a), Stage 2 (b), and Stage 3 (c).}\label{fig:semantic_nu_utility}
\end{figure}

\section{Conclusions}
\label{sec:conclusions}

In this paper, we have  presented a new neural network classifier based on deep CNN and DS theory for set-valued classification, called the evidential deep-learning classifier. This new classifier consists of several stages for feature representation, a DS layer to construct mass functions, and an expected utility layer to make set-valued assignments based on the mass functions. The classifier can be trained in an end-to-end way. Besides, we have proposed a strategy to select partial acts instead of considering all of them.

A major finding of this study is that the hybridization of deep CNNs and evidential neural networks by plugging  DS and expected utility layers at the output of a CNN makes it possible to improve the performance of deep CNN models by assigning  ambiguous patterns to multi-class sets. The proposed classifier is able to select a set of classes when the object representation does not allow us to select a single class unambiguously, which easily leads to incorrect classification in  probabilistic classifiers. This result provides a novel direction to improve the cautiousness of deep CNNs for object recognition. The use of DS and expected utility layers also improves precise classification performance. The hybridization also makes it possible to reject outliers together with ambiguous patterns when the tolerance degree of imprecise is between 0.7 and 0.9. Additionally, the strategy of selecting partial multi-class acts works as well as that of considering all $2^{|\Omega|}$ acts.

Future work will focus on three main aspects. First, we will extend the proposed classifier to pixel-wise segmentation, where each pixel in an image must be assigned to one of the subsets of $\Omega$. Secondly,  other advanced evidential combination rules, such as contextual-discounting evidential $K$-nearest neighbor \cite{denoeux19f} will be studied to improve the performance of the proposed classifier. Finally, we will consider modifications of the model   introduced in this paper to make it applicable to regression problems.

\section*{Acknowledgement}

This research was supported by a scholarship from the China Scholarship Council and by the Labex MS2T  (reference ANR-11-IDEX-0004-02).

\section*{References}
\label{sec:references}

\bibliography{mybibfile}
\end{document}